\documentclass[final,1p,times,twocolumn]{elsarticle}
\usepackage{graphicx,subcaption}
\usepackage{lineno,hyperref}
\modulolinenumbers[5]
\usepackage[section]{placeins}
\journal{Journal of Mechanical Systems and Signal Processing}
\usepackage{booktabs}
\usepackage{graphicx} 
\usepackage[figuresright]{rotating}
\usepackage{adjustbox}
\usepackage{bm} 
\usepackage{tensor} 
\usepackage{tikz}
\usepackage{colortbl}
\usepackage{tabulary}
\usepackage{etoolbox}
\usepackage{wrapfig}
\usepackage{ragged2e}
 \usepackage{cleveref}
\usepackage[utf8]{inputenc}
\usepackage{xcolor,soul}
\usepackage[english]{babel}
\usepackage{fancyhdr}
\usepackage{booktabs}
 \usepackage{nopageno}
\usepackage{caption}
\usepackage[font=normalsize]{caption} 
\addto\captionsenglish{}
\usepackage{nopageno}
\usepackage{wrapfig}
\usepackage{caption}
\usepackage{threeparttable}
\usepackage[ruled,vlined]{algorithm2e}
\usepackage[font=normalsize]{caption} 
\usepackage{lineno, blindtext}
\addto\captionsenglish{}
\SetKwInput{KwInput}{Input}  
\SetKwInput{KwOutput}{Output} 
\usepackage{placeins}
\usepackage{moreverb,url}
\usepackage{soul}
\usepackage[english]{babel}
\usepackage[T1]{fontenc}
\usepackage[utf8]{inputenc}
 \usepackage{threeparttable}
\usepackage[ruled,vlined]{algorithm2e}

\definecolor{C0}{rgb}{0, 0, 0}
\definecolor{C1}{rgb}{0, 0, 0}
\definecolor{C2}{rgb}{0, 0, 0}
\definecolor{C3}{rgb}{0, 0, 0}

\newcommand\BibTeX{{\rmfamily B\kern-.05em \textsc{i\kern-.025em b}\kern-.08em
T\kern-.1667em\lower.7ex\hbox{E}\kern-.125emX}}
\usepackage{xspace}
\newcommand*{\eg}{\emph{e.g.},\@\xspace}
\newcommand*{\ie}{\emph{i.e.},\@\xspace}

\usepackage{threeparttable}

\begin{document}
\begin{frontmatter}

\title{Zero-Shot Transfer Learning for Structural Health Monitoring using Generative Adversarial Networks and Spectral Mapping}



\author[1]{Mohammad Hesam Soleimani-Babakamali}
\ead{soleimanibabakamali@gmail,com}
\author[2]{Roksana Soleimani-Babakamali}
 \ead{Soleimanir78@univie.ac.at}
\author[3]{Kourosh Nasrollahzadeh}
\ead{nasrollahzadeh@kntu.ac.ir}
\author[4]{Onur Avci\corref{mycorrespondingauthor}}
\ead{onur.avci@mail.wvu.edu}
\author[5]{Serkan Kiranyaz}
\ead{mkiranyaz@qu.edu.qa}
\author[1]{Ertugrul Taciroglu}
\ead{etacir@g.ucla.edu}
\address[1]{Department of Civil and Environmental Engineering, University of California, Los Angeles, CA, USA}
\address[2]{Department of Computer Science, University of Vienna, Vienna, Austria}
\address[3]{Department of Civil Engineering, K. N. Toosi University of Technology, Tehran, Iran}
\address[4]{Department of Civil and Environmental Engineering, West Virginia University, WV, USA}
\address[5]{Department of Electrical Engineering, Qatar University, Qatar}

\cortext[mycorrespondingauthor]{Corresponding author, onur.avci@mail.wvu.edu}

\begin{abstract}
\noindent
Gathering properly labeled, adequately rich, and case-specific data for successfully training a purely data-driven or hybrid model for structural health monitoring (SHM) applications is a challenging task. We posit that a Transfer Learning (TL) method that utilizes available data in any relevant source domain and directly applies to the target domain through domain adaptation can provide substantial remedies to address this issue. Accordingly, we present a novel TL method that differentiates between the source's no-damage and damage cases and utilizes a domain adaptation (DA) technique. The DA module transfers the accumulated knowledge in contrasting no-damage and damage cases in the source domain to the target domain, given only the target's no-damage case. High-dimensional features allow employing signal processing domain knowledge to devise a generalizable DA approach. The Generative Adversarial Network (GAN) architecture is adopted for learning since its optimization process accommodates high-dimensional inputs in a zero-shot setting. At the same time, its training objective conforms seamlessly with the case of no-damage and damage data in SHM since its discriminator network differentiates between real (no damage) and fake (possibly unseen damage) data. An extensive set of experimental results demonstrates the method's success in transferring knowledge on differences between no-damage and damage cases across three strongly heterogeneous independent target structures. The area under the Receiver Operating Characteristics curves (Area Under the Curve - AUC) is used to evaluate the differentiation between no-damage and damage cases in the target domain, reaching values as high as 0.95. With no-damage and damage cases discerned from each other, zero-shot structural damage detection is carried out. The mean $F_1$ scores for all damages in the three independent datasets are 0.978, 0.992, and 0.975. The success of the proposed TL approach is expected to pave the way for further improvements in the accuracy and generalizability of data-driven SHM applications.

\end{abstract}

\begin{keyword}

\texttt Transfer Learning; SHM; Domain Adaptation; Zero-shot Learning; Generative Adversarial Networks; Structural Damage Detection.
\end{keyword}

\end{frontmatter}

\section{Introduction}
\label{sec:Intro}
\noindent
Dense wireless networks featuring inexpensive sensor units are now providing massive amounts of data ~\cite{jsan9030039}. Such sensor networks can be utilized for structural health monitoring (SHM) of civil infrastructure~\cite{MOHAMMADIGHAZI2017578,doi:10.1177/1475921719894186}—--more specifically, for detecting, localizing, and quantifying structural damage. Moreover, recent developments in big data science and computing hardware are rendering the analyses of massive data routinely accessible ~\cite{BIBRI2017449}. The missing piece of the puzzle for pervasive SHM at large scales is learning and transferring information across structures or structure populations. The inherent diversity of civil infrastructure (\eg bridges, buildings, tunnels, dams, and power plants) require versatile, robust, and generalizable approaches. 

SHM algorithms---operating on multi-modal sensor data such as accelerations, strains, temperature, etc.---can achieve pattern recognition and distinguish between no-damage and damage structural cases \cite{Worden2007}. Accordingly, SHM procedures, including Structural Damage Detection (SDD), often make use of Machine Learning (ML) techniques with various supervision levels (supervised, semi-supervised, or unsupervised). In classical methods---\eg parametric approaches that track modal features~\cite{wah2021regression,de2019automated,perez2016time,LANGONE201764}---various distance metrics across signals \cite{LEE2021112330,HOELL2016557}, as well as statistical or heuristic \textit{ad hoc} representative values of signals \cite{amezquita2019nonlinear,amezquita2015synchrosqueezed} are used for training ML or statistical models. Even though such features can be extracted globally from various systems, prior studies revealed that the sensitivity of those parameters to damage cases is not scalable across different systems~\cite{abdeljaber20181}.

With the emergence of deep neural networks and breakthroughs in their training---\eg ADAM optimizer~\cite{kingma2014adam}, rectified linear-unit activation layers~\cite{Relu}, and back-propagation \cite{rumelhart1986learning}---customized regression-based data-driven or model-based SHM approaches are constantly losing ground to deep neural networks \cite{A_Over}. Deep neural networks are able to extract damage-sensitive features automatically through representation learning \cite{rumelhart1986learning} from high-dimensional input data and simultaneously perform damage detection~\cite{abdeljaber20181, azimi2020structural,REDDY2019106823,https://doi.org/10.1002/stc.2824}. However, these techniques have certain critical limitations:  ($i$) they need labeled data for each new (unseen) structure, ($ii$) even if properly labeled data exists, one must train a new network for every structure, and even when this is possible, and ($iii$) such supervised methods could arguably contradict the ultimate goal of SHM, which is to detect unseen anomalies when/if they take place in the future. If these challenges can be addressed properly, the resulting methods would be generalizable and pave the way for the next generation of SHM tools that can be applied to entire populations of structures. 

There appear to be two major directions to address the need for labeled information for a new structure. The first direction involves using generative models, and the second involves developing Transfer Learning (TL) mechanisms. Generative models can generate SHM data with considerable variation \cite{soleimani2021system}. Generative Adversarial Networks (GANs) have recently been used for data augmentation in SHM applications~\cite{luleci2022generative} and to alleviate the data imbalance issues \cite{zareapoor2021oversampling,soleimani2021general}. Through data generation, Soleimani-Babakamali \textit{et al.} proposed an unsupervised SHM technique generating samples from the incoming data stream to address the need for comprehensive prior information~\cite{soleimani2021system,soleimani2022toward}. In comparison, TL can transfer the accumulated knowledge of discerning between no-damage and damage structural states across different systems (\ie ``source'' and ``target '' domains) through either model updating (\ie model transfer) or domain adaptation (\ie feature transfer). Fine-tuning is an instance of model transfer in which the trained source model parameters are updated via the target domain data and objectives (\eg classification). Several studies reported its application in SHM, most of which are image-based~\cite{https://doi.org/10.1111/mice.12363,gopalakrishnan2018crack,tang2022deep,rai2022transfer}. The image-based fine-tuning owes its success to the availability of massive image datasets, such as ImageNet~\cite{IMageN}, which takes advantage of pre-trained DL models on such datasets. Yet, labeled data must be available in the target domain to adapt the pre-trained networks through fine-tuning. Despite the usefulness of images for different damage types (\ie mechanical faults), managing such large datasets and unseen damages are the major complications of that method. 

In contrast, with the domain adaptation (DA) scheme to transfer the target features to the source domain, the network trained over the source domain applies to the target structure without further model training. Thus, regardless of the potential of generative models, ``zero-shot'' (\ie online damage detection without witnessing damage) DA-based TL is well-aligned with the requirements and constraints of large-scale SHM. In large-scale SHM, SHM tools are applied to populations of diverse infrastructure in a city, for instance, where gathering prior (no-damage and damage cases) data for every single entity is not practical. It can void the need for labeled data in a new (target) structure by using the knowledge gained from comprehensive source domain data while covering the second intricacy via a domain-knowledge-based DA method. Additionally, a TL method can contain generative models and benefit from their associated capabilities.

Transfer Learning for SHM has received increasing interest in the past two years, with several notable attempts appearing in the open literature. A recent TL study by Wang and Xia~\cite{W_A} explored a re-weighting method where no labeled data is required from the target space using adversarial domain adoption. In this approach, the TL takes place through re-weighting and discarding the source data cases that do not share considerable information with the target domain. Yet, there are two challenges with this approach. First, the target database should contain damage cases, even though it is unlabeled. Second, training a neural network with source and target domain data is required in one of the quadruple stages of that approach. 

Population-based TL, published in four parts, including knowledge transfer across homogeneous (\eg the population of wind turbines)~\cite{BULL2021107141}, and heterogeneous population of structures~\cite{GOSLIGA2021107144,GARDNER2021107142,TSIALIAMANIS2021107692}, is another notable study in the SHM-TL area. This method is capable of TL for the target dataset without using its labeled data. An abstract representation of structures is formed for heterogeneous systems to determine whether the information is transferable. Then communities of structures that can communicate information are formed. Later, the knowledge (\ie the detector trained on the source class) is transferred to identify anomalies in the unlabeled target class. This approach entails two constraints. First, it employs highly compressed representations of structural states; second, TL only occurs in systems within the same community. Summarizing the entire system to an abstract representation is an instance of dimensionality reduction (DR). DR and how it impacts the detection of unseen anomalies were studied in detail by Soleimani Babakamali~\cite{SOLEIMANIBABAKAMALI2023109910}. Results indicated that DR-induced loss of information could mask damages since, in zero-shot SHM, data dimensions through which future damages might manifest themselves can get removed.

To overcome the obstacles mentioned above, this study offers a novel TL approach for transferring the binary knowledge of no-damage/damage conditions across different structures for zero-shot damage detection. The proposed approach has the following characteristics: First, data features are selected as full-spectrum Fast Fourier Transform (FFT) amplitudes of input signals---\ie no damage-sensitive feature extraction is carried out, which helps generalizability. Second, the method transfers knowledge through domains without altering the neural network trained to differentiate between the source's no-damage and damage cases. Accordingly, the novel and significant contributions of this study can be summarized as follows:

\begin{itemize}

\item Utilizing full-spectrum FFT amplitudes as features allow the TL to take place across strongly heterogeneous, real-world structures---herein, the ``Z24 bridge,'' ``Yellow Frame,'' and ``Qatar University Grandstand Simulator.'' These structures are of different types and sizes and exhibit distinct damage types. Moreover, data collected from each structure were collected by different research groups.

\item The proposed TL method operates on the target no-damage case data only---that is, the ``baseline'' data that comes in as the SHM tool starts to work. This attribute enables large-scale SHM.

\item DA is performed by reordering and multiplying the spectral data of the target structure to have the same shape as the spectral data of the source structure. This simple arithmetic domain adaptation transforms the target data so that the model trained to differentiate between the source's no-damage and damage cases applies to the target domain.

\end{itemize}


\section{Datasets}
\label{sec:datasets}
\noindent
In this section, three benchmark datasets that are used to evaluate the proposed TL framework are introduced. All of them contain vibration data, including ($i$) a concrete box girder bridge (Z24 bridge), ($ii$) a scaled steel frame ambient excitation (Yellow Frame), and ($iii$) a scaled stadium model excited in a laboratory setting (Qatar University Grandstand Simulator).

\subsection{The Yellow Frame}
\label{Section:YF}
\begin{figure}[t!]
\centering
    \begin{subfigure}[b]{0.45\linewidth} 
        \centering
\includegraphics[width=0.7\linewidth]{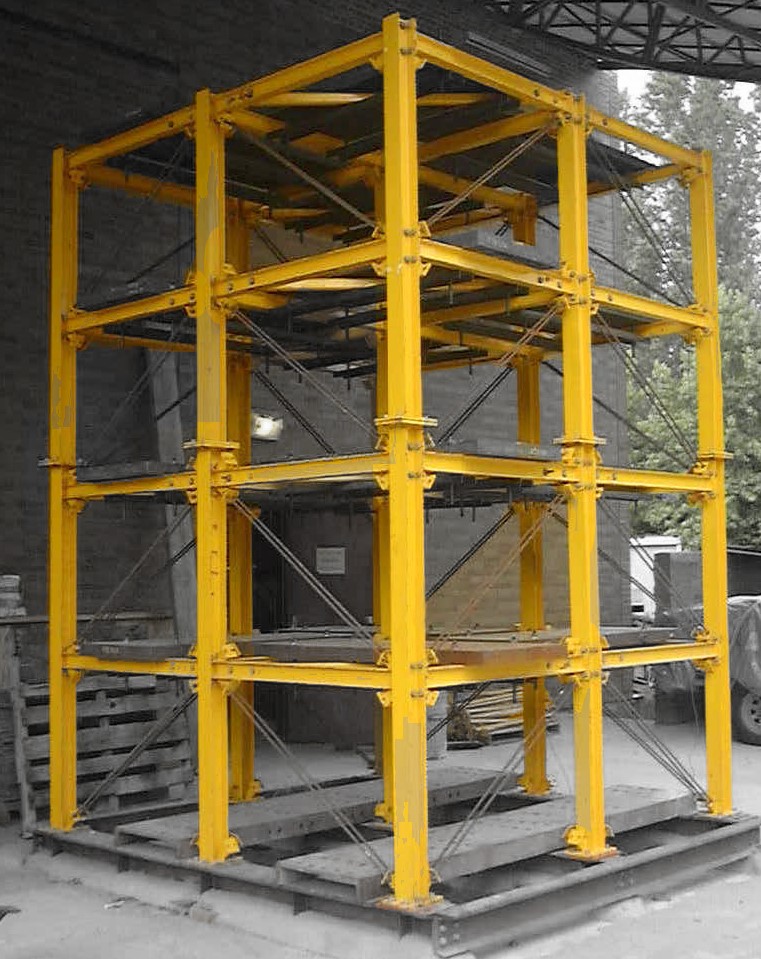}
        \caption {}
        \label{fig:Y1S}
    \end{subfigure}
    \begin{subfigure}[b]{0.45\linewidth} 
        \centering
\includegraphics[width=0.92\linewidth]{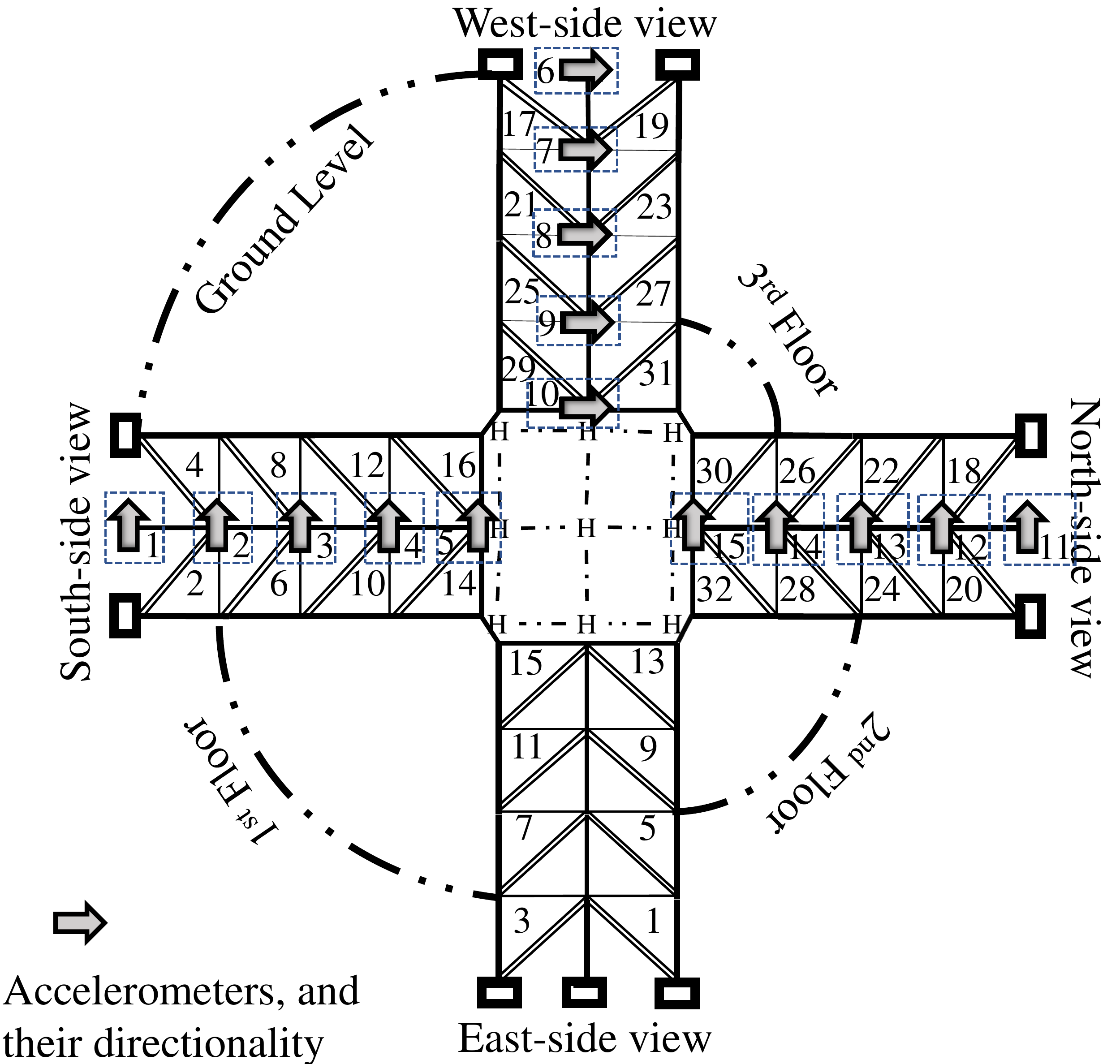}
        \caption {}
        \label{fig:Y2S}
    \end{subfigure}
\caption {Yellow Frame: (a) structure (b) brace and sensor configuration.}
\label{Fig:YF_Str}
\end{figure}

\noindent
The benchmark ``Yellow Frame'' has been the source for various control and SHM datasets. This one-third scaled, three-story steel frame located at the University of British Columbia is equipped with modular structural elements, including masses, braces, beams, and columns, to synthesize various structural states ~\cite{mendler2019yellow} (Fig.~\ref{fig:Y1S}). The SHM-based Yellow Frame dataset studied herein considers ``brace removal'' as damage. The configuration of sensors is shown in Fig.~\ref{fig:Y2S}. Accordingly, data cases are explained in Table ~\ref{Table:Yellow_Frame_Data}. This table includes the fundamental frequency (\ie first-mode frequency) of data cases obtained via the ARTeMIS software~\cite{Art} with the Enhanced Frequency Domain Decomposition method~\cite{brincker2015introduction}. This dataset has a sampling rate of 1000 Hz.

\begin{table}[h!]
\begin{threeparttable}
  \caption{Yellow Frame dataset summary; brace labels are according to Fig.~\ref{fig:Y2S}.}
  \label{Table:Yellow_Frame_Data}
  \centering
  \begin{tabular*}{\linewidth}{@{\extracolsep{\fill}} lccccc}
    \toprule
\parbox{ 0.8cm}{Damage class}&\parbox{3.5cm}{\centering Removed brace ID} &  \parbox{0.8cm}{\centering F.F (Hz)\tnote{*}}  & \parbox{0.8cm}{\centering Damage class}&\parbox{4cm}{\centering Removed brace ID} &  \parbox{0.8cm}{\centering F.F (Hz)}\\
    \midrule
DC0&	None&	\parbox{0.8cm}{\centering 7.62}&	DC11&\parbox{4cm}{\centering DC10 + (17, 19, 25, 27)(I)\tnote{**}}&	\parbox{ 0.8cm}{\centering 6.77}\\

DC1&	2,4(II)\tnote{***}&	\parbox{ 0.8cm}{\centering 5.79}&	DC12&\parbox{4cm}{\centering DC7 + (1, 3, 17, 18)(I)}&	\parbox{ 0.8cm}{\centering 6.40} \\

DC2&	DC1 + (18, 20)(II)&	\parbox{ 0.8cm}{\centering 4.76}&	DC13&	(10, 12)(II)&	\parbox{ 0.8cm}{\centering 6.41}\\

DC3&	\parbox{3.5cm}{\centering  DC2+ (1, 3, 17, 19)(II)}& 	\parbox{ 0.8cm}{\centering 4.88}&DC14	&\parbox{4cm}{\centering DC13 + 21(II),  23(I)}&	\parbox{ 0.8cm}{\centering 6.52}\\

DC4&	DC1 + (17, 19)(II)&	\parbox{ 0.8cm}{\centering 5.38}&	DC15	&(21, 23)(II)&	\parbox{ 0.8cm}{\centering 6.87}\\

DC5&	DC1 + (18, 20)(I)&	\parbox{ 0.8cm}{\centering 5.40}&	DC16&	\parbox{4cm}{\centering (7-8, 21, 22)(I)}&\parbox{ 0.8cm}{\centering 6.98}\\

DC6	&2(II)	&\parbox{ 0.8cm}{\centering 7.19}&	DC17	&\parbox{4cm}{\centering (5, 6, 7, 8, 21, 24)(I)}&	\parbox{ 0.8cm}{\centering 6.58}\\

DC7&	(2 , 4)(I)&	\parbox{ 0.8cm}{\centering 6.85}&	DC18	&\parbox{4cm}{\centering DC17 + (7, 8, 21, 22)(I)}	&\parbox{ 0.8cm}{\centering 6.04}\\

DC8&	(25, 27)(I)&	\parbox{ 0.8cm}{\centering 7.29}&	DC19	&\parbox{4cm}{\centering DC18 + (5, 6,  23, 24)(I)}&	\parbox{ 0.8cm}{\centering 4.92}\\

DC9	&\parbox{3.5cm}{\centering (29, 31, 8, 6)(I)}&	\parbox{ 0.8cm}{\centering 6.90}&DC20&\parbox{4cm}{\centering	(6, 8)(II), (21, 22, 23, 24)(I)}&	\parbox{ 0.8cm}{\centering 5.40}\\

DC10	&\parbox{3.5cm}{\centering (21, 23 ,29 , 31)(I)}&	\parbox{ 0.8cm}{\centering 7.18}&&&\\
\bottomrule
\end{tabular*}
\begin{tablenotes}\footnotesize
\item\item[*] Fundamental mode frequency
\item[**] One brace is removed
\item[***] Both braces are removed
\end{tablenotes}
\end{threeparttable}
\end{table}

\subsection{The Z24 Bridge}
\begin{figure}[h!]
    \centering
    \begin{subfigure}[b]{1\linewidth}
        \centering
         \includegraphics[width=1\linewidth]{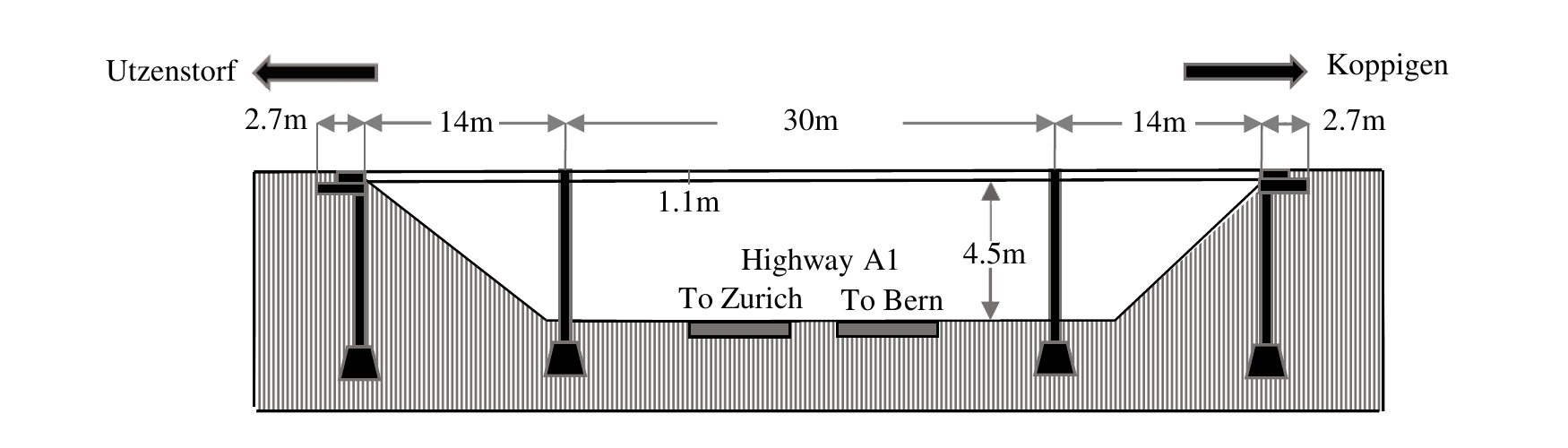}
    \caption{Z24 geometry~\cite{reynders2009continuous}.}
    \label{Fig:Z24_Structure}
    \end{subfigure}
    \begin{subfigure}[b]{1\linewidth}
        \centering
         \includegraphics[width=1\linewidth]{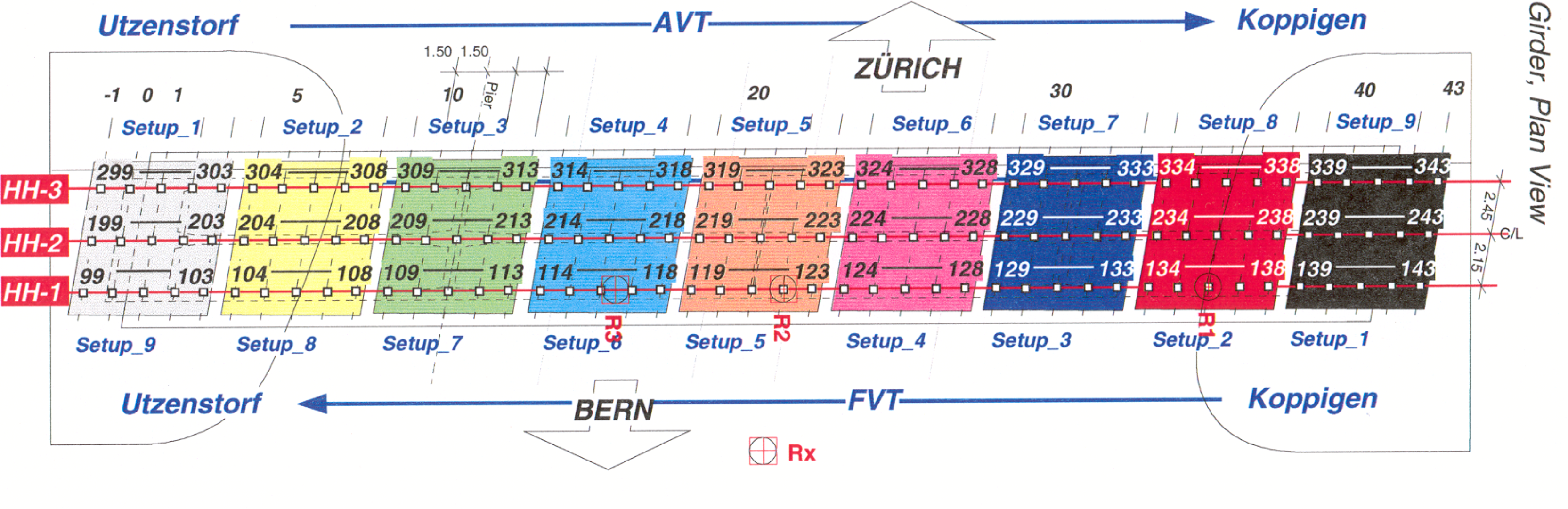}
    \caption{Z24 sensor setups (adopted from~\cite{reynders2009continuous}).}
        \label{Fig:Z24_Setup}
    \end{subfigure}
    \caption{Z24 bridge.}
        \label{Fig:Z24}
    \end{figure}

\noindent
Before its demolition, Z24 Bridge overpassed the A1 highway between Bern and Zürich in Switzerland (Fig.~\ref{Fig:Z24_Structure}). Two benchmark datasets from the Z24 Bridge exist, one from long-term observations captured over a year, and the other covers a series of damage states during the last month before its demolition. Long-term monitoring was primarily utilized for zero-shot anomaly detection (\ie no-damage and damage cases) in the literature~\cite{wah2019removal, wah2021regression, sarmadi2021early}. These prior studies used comprehensive long-term data to build a baseline under various environmental and operational  conditions~\cite{doi:10.1177/1475921713502836}. That baseline is used for anomaly detection on the last portion of damage case data. The short-term dataset includes 17 data cases from various severity of seven distinct damage cases. This study utilizes the short-term dataset, with one class from each of the seven damage cases. Z24 dataset summary is reported in Table~\ref{Z24-D}. This dataset has a sampling rate of 100 Hz.

Z24 instrumentation includes several setups (Fig.~\ref{Fig:Z24_Setup}). Each setup includes multiple sensors and data channels (\ie vertical, horizontal, and transverse channels). We filtered out faulty data channels to ensure damages are being identified from the structure's vibrations rather than faults in the sensors' readings (\eg channels with trending, dead, or square wave signals). After inspecting all data channels in the eight data cases above (\ie one no-damage and seven damage cases), 15 channels are selected from various sensor setups across the bridge (Fig.~\ref{Fig:Z24_Setup}), reported in Table~\ref{Table:Z24_Setups}.

\renewcommand{\arraystretch}{1}
\begin{table}[h!]
  \caption{Z24 dataset summary.}
  \label{Z24-D}
  \centering
  \begin{tabular*}{\linewidth}{@{\extracolsep{\fill}} lcc}
    \toprule
\parbox{2cm}{Data class}&\parbox{5cm}{\centering Description~\cite{reynders2014vibration}} \\
    \midrule
DC0&First reference measurement (Reference class)\\
DC1&Lowering of the pier, 80 mm\\
DC2&Lifting of the pier, the tilt of the foundation \\
DC3&Spalling of concrete at soffit, $24 m^2$ \\
DC4&Landslide of 1 m at abutment \\
DC5&Failure of the concrete hinge \\
DC6&Failure of 4 anchor heads\\
DC7&Rupture of 6 out of 16 tendons \\
\bottomrule
\end{tabular*}
\end{table}

\renewcommand{\arraystretch}{1}
\begin{table}[h!]
  \caption{Selected channels from Z24 benchmark~\cite{reynders2009continuous}.}
  \label{Table:Z24_Setups}
  \centering
  \begin{tabular*}{\linewidth}{@{\extracolsep{\fill}} l ccccc}
    \toprule
\parbox{1.5cm}{Setup}&Channel&Setup&&\parbox{1.5cm}{Channel}&\\
\midrule
Setups 1, 7&299V&Setup 4 &100V&101V&103V\\
Setup 2&199V&Setup 5 &99V&101V&299V\\
Setups 3, 9&300V&Setup 6 &101V&299V&300V\\
Setup 8&101V&&&  &
\\

\bottomrule
\end{tabular*}
\end{table}
   
\subsection{The Qatar University Grandstand Simulator}
\begin{figure}[t!]
    \centering
     \includegraphics[width=0.6\linewidth]{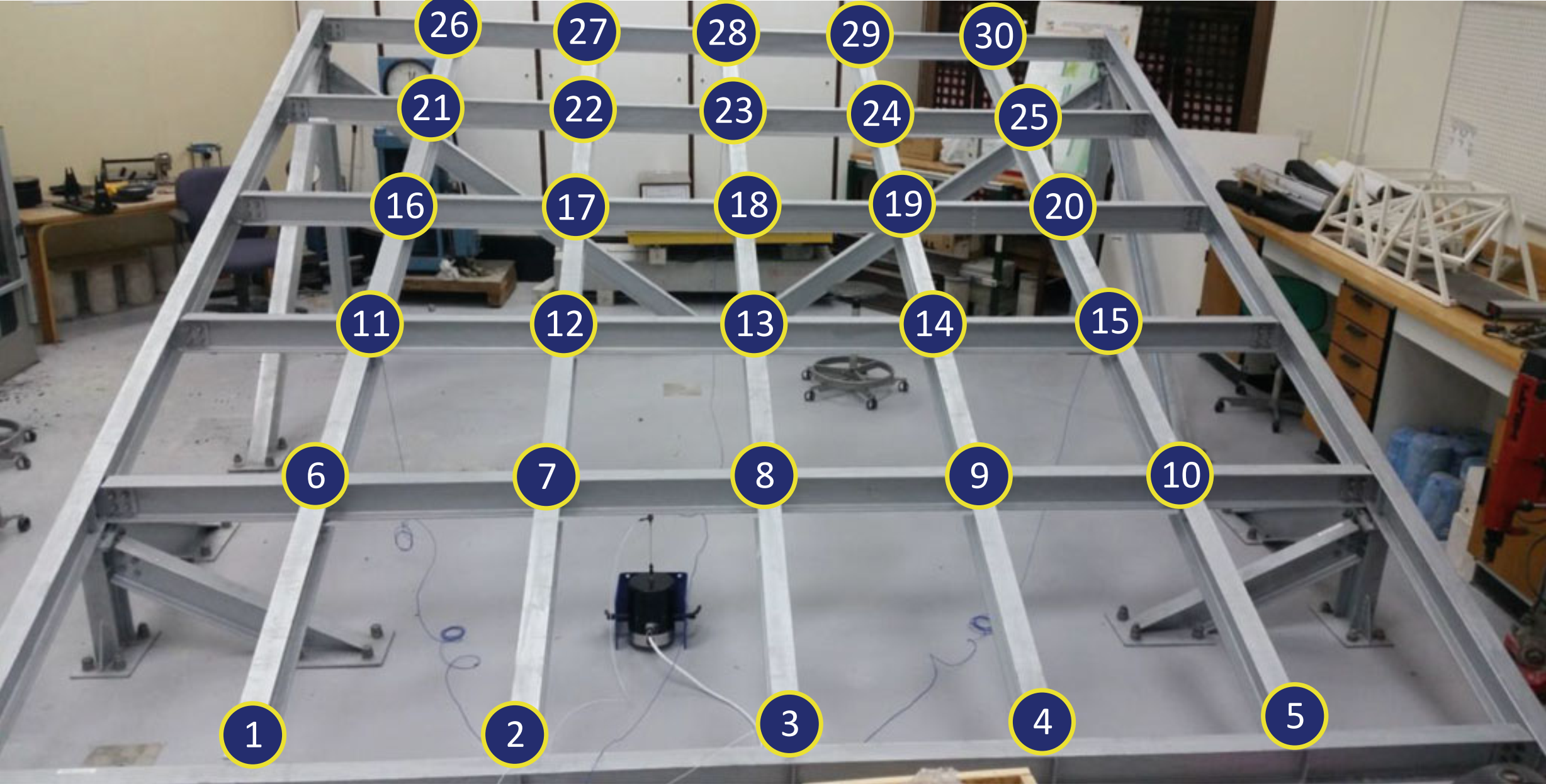}
    \caption{QUGS structure~\cite{QUGS_D}.}
            \label{Fig:QUGS_Structure}
    \end{figure}

\noindent
Qatar University Grandstand Simulator (QUGS) dataset aims at benchmarking stadium models for SHM and damage detection purposes. The structure is an inclined steel frame representing a classical stadium seating layout in grid form (Fig.~\ref{Fig:QUGS_Structure}). QUGS instrumentation and details are made available by Avci \textit{et al.}~\cite{avci2018wireless}. This dataset contains one no-damage (DC0) and thirty damage cases (DC1-DC30) induced by bolt-loosening at the beam to girder connections~\cite{QUGS_D,abdeljaber2016dynamic}. At each joint, one permanent accelerometer is dedicated to collect data (Fig.~\ref{Fig:QUGS_Structure}) with a sampling frequency of 1000 Hz.

\begin{figure}[t!]
    \centering
     \includegraphics[width=0.55\linewidth]{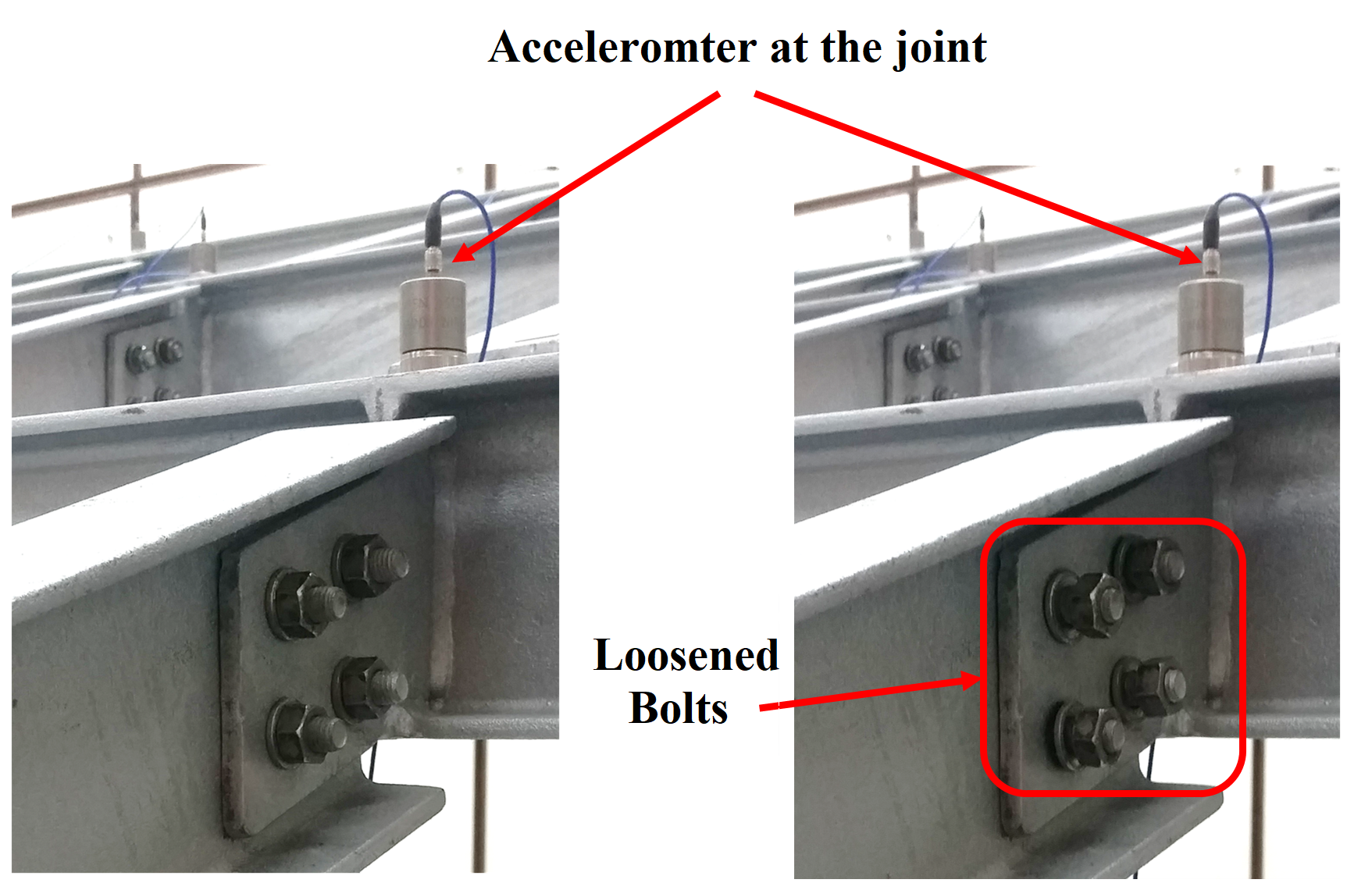}
              \vspace{-0.3cm}
    \caption{QUGS joint detailing~\cite{abdeljaber2017real}.}
            \label{Q-D}
    \end{figure}

\section{Methodology}
\begin{figure}[h!]
\centering
\includegraphics[width=1\linewidth]{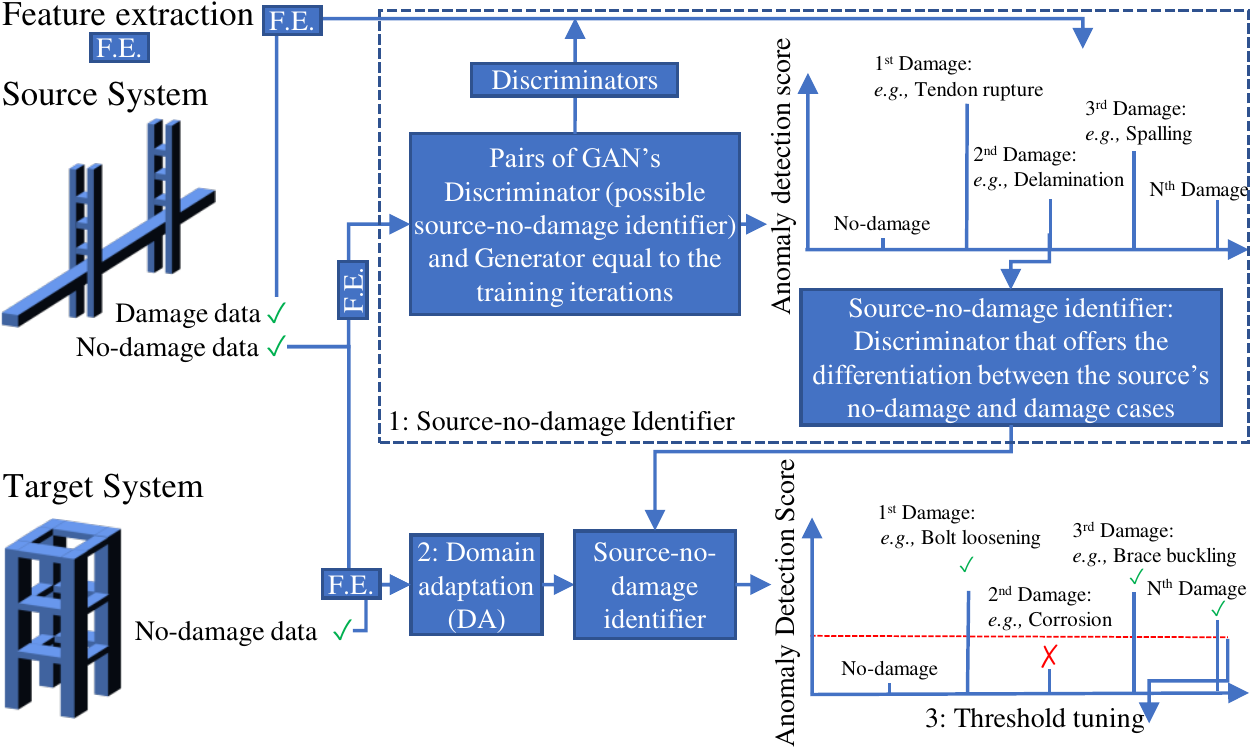}
        \caption {Overview of the proposed TL method.}
        \label{Fig:Overview}
      \end{figure}

\noindent
The overview and application of the proposed TL technique are shown in Fig.~\ref{Fig:Overview}. It consists of three modules, including ($i$) a source-no-damage identifier that differentiate between the source domain's no-damage and damage cases, ($ii$) a DA module that transforms the target domain data features obtained through the feature extraction routine (Section~\ref{Sec:Feature}), so that the source-no-damage identifier applies to the target domain, and ($iii$) a threshold tuning method for zero-shot structural damage detection (SDD). The majority of the article describes and assesses the first two modules. A threshold tuning method is introduced in Section~\ref{sec:perf}, and zero-shot SDD is carried out in the results section. 

\subsection{Feature Extraction}
\label{Sec:Feature}
\noindent
The source-no-damage identifier is designed to work with the full-spectrum FFT amplitudes of input signals without performing any damage-sensitive feature selection. There are two reasons for this: ($i$) to avoid the sensitivity loss to unseen damages~\cite{SOLEIMANIBABAKAMALI2023109910}, and ($ii$) to accommodate knowledge transfer across various systems since the full-spectrum captures a broadband representation of systems' dynamic characteristics. Each structure's sensor (vibration) data is considered a distinct data channel where the number of data channels varies among the structures. Performing the data acquisition from the $N$ data channels of a particular structure, each data channel's feature extraction is carried out as follows: First, each channel data is segmented into the analysis windows of length $W$. We perform $W$-points Fast Fourier Transform (FFT), and the first $W/2$ power spectral coefficients are divided over their mean value and are used for analysis. GANs, which contain the source-no-damage identifiers, have shown impressive generative capabilities over bounded data---\eg image data (between 0-255 at each color-channel)~\cite{zhang2019self} and text data (between -1 and 1 via Transformers~\cite{xu2018diversity}). Considering the vibration data, the application of GAN and how bounding GAN inputs impact its generative power (\ie the better the data generation, the better discriminator distinguishes between real and fake data) was studied by Soleimani- Babakamali \textit{et al.}~\cite{soleimani2021general}. The bounding value of 10---\ie rounding normalized amplitudes greater than ten to ten---for features to vary between 0 to 10, was suggested in that study. We apply the same standard rounding normalized FFT amplitudes over ten to ten. The feature vector is labeled as $F_i$ for the channel $i$. The concatenation of all $N$-channel features constitutes the final feature vector for each analysis window, $F$ ($F \in  \mathbb{R}^{(N* \frac{W}{2} \times 1)}$). $F$ is naturally high-dimensional feature vector considering typical $W$ and $N$ values in SHM problems. For instance, each $F$ has 15000 dimensions for $N$ = 15 and $W = 2000$.

\subsection{source-no-damage Identifier Module}
\noindent
 Generative Adversarial Networks'~\cite{goodfellow2014generative} optimization process aligns perfectly with the SHM; the GAN models are optimized by training a Discriminator ($\mathcal{D}$) network to distinguish between ``real,'' \ie inputs of $\mathcal{D}$, and ``fake'' data generated by the Generator ($\mathcal{G}$) network. When the real data is sampled from the no-damage case of the source structure, $\mathcal{D}$ will learn to differentiate the no-damage case from any other data---including the damage case data. Here the basic assumption is that the no-damage case data patterns are sufficiently different from the damage case. Therefore, $\mathcal{D}$ can detect any data of the source structure damage case with proper training. Using the damage case data as the validation set, during the training, the $\mathcal{D}$ model, which gives the highest detection accuracy over the damage case of the source structure, can then be selected as the source-no-damage identifier network. It is worth mentioning that in online SHM, the current state of structures equals the no-damage state, as sought damages occur in the future.

\begin{figure}[t!]
\centering
\includegraphics[width=0.9\linewidth]{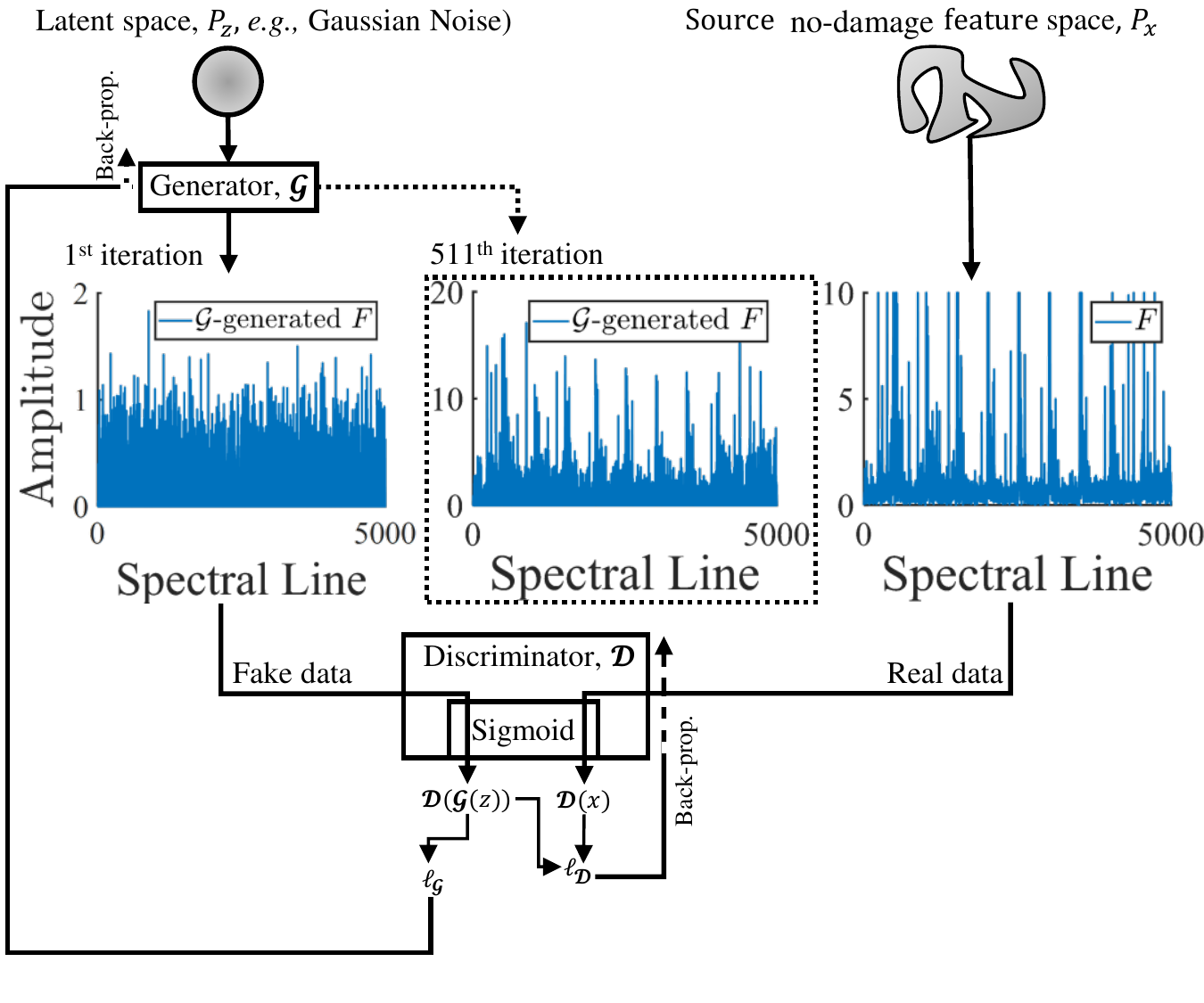}
\caption {GAN's training schema; $\ell_\mathcal{D}$ and $\ell_\mathcal{G}$ denote training losses for $\mathcal{D}$ and $\mathcal{G}$, respectively. Sample GAN's generation at iteration 511 shows the progress of $\mathcal{G}$ in generating more real-resembling fake data as the training continues.}
\label{Fig:GAN_Training_Scheme}
\end{figure}

In the GAN training process, $\mathcal{G}$ samples from a ``latent space'', denoted as $P_{z}$, generating fake data. A schematic of GAN's training with features $F$ is shown in Fig.~\ref{Fig:GAN_Training_Scheme}. The GAN's training is shown on that image with the Yellow Frame data features with $W=1000$ and $N=10$, leading to 5000 spectral lines in $F$ (Section~\ref{Sec:Feature}). In that image, $\ell_\mathcal{D}$ and $\ell_\mathcal{G}$ are loss functions that optimize $\mathcal{G}$, and $\mathcal{D}$, respectively. $\ell_\mathcal{G}$ penalizes $\mathcal{G}$ if $\mathcal{G}$-generated fake data are correctly identified by $\mathcal{D}$ as fake; otherwise, $\ell_\mathcal{D}$ penalizes $\mathcal{D}$. The impact of $\ell_\mathcal{G}$ shows itself with $\mathcal{G}$ implicitly understanding the training data distribution on the course of training, which led to generating more real-resembling data. As the training goes on with more realistic generations, $\mathcal{D}$ better understands the patterns in the no-damage case to differentiate it from dissimilar data. Those loss functions resemble a zero-sum game, in which the gain of one participant (\eg $\mathcal{D}$) is the loss of the other (\ie $\mathcal{G}$). The losses are expected to reach an equilibrium state, following the Nash equilibrium~\cite{nash1950equilibrium}. There exist various loss formulations for GAN; The following formulae are selected from the seminal GAN paper~\cite{goodfellow2014generative} with Sigmoid activation on $\mathcal{D}$'s output layer:

\begin{equation}
\label{Eq:D_Loss}
\ell_\mathcal{D} = -\mathbb{E}_{x \sim P_{x}}\left[\log \mathcal{D}(x)\right] - \mathbb{E}_{z \sim P_{z}}\left[\log \left(1-\mathcal{D}(\mathcal{G}(z))\right)\right]
 \end{equation}
\begin{equation}
 \label{Eq:G_Loss}
\ell_\mathcal{G}  =  -\mathbb{E}_{z \sim P_{z}}\left[\log\mathcal{D}\left(\mathcal{G}(z)\right)\right],
 \end{equation}
where $x$ is a data instance from the training set, herein the no-damage case, and $z$ is the random latent data drawn by $\mathcal{G}$ from the latent space $P_{z}$.

\subsubsection{GAN Architecture}
\begin{figure}[t!]
\begin{subfigure}[b]{1\linewidth}
\centering
\includegraphics[width=0.85\linewidth]{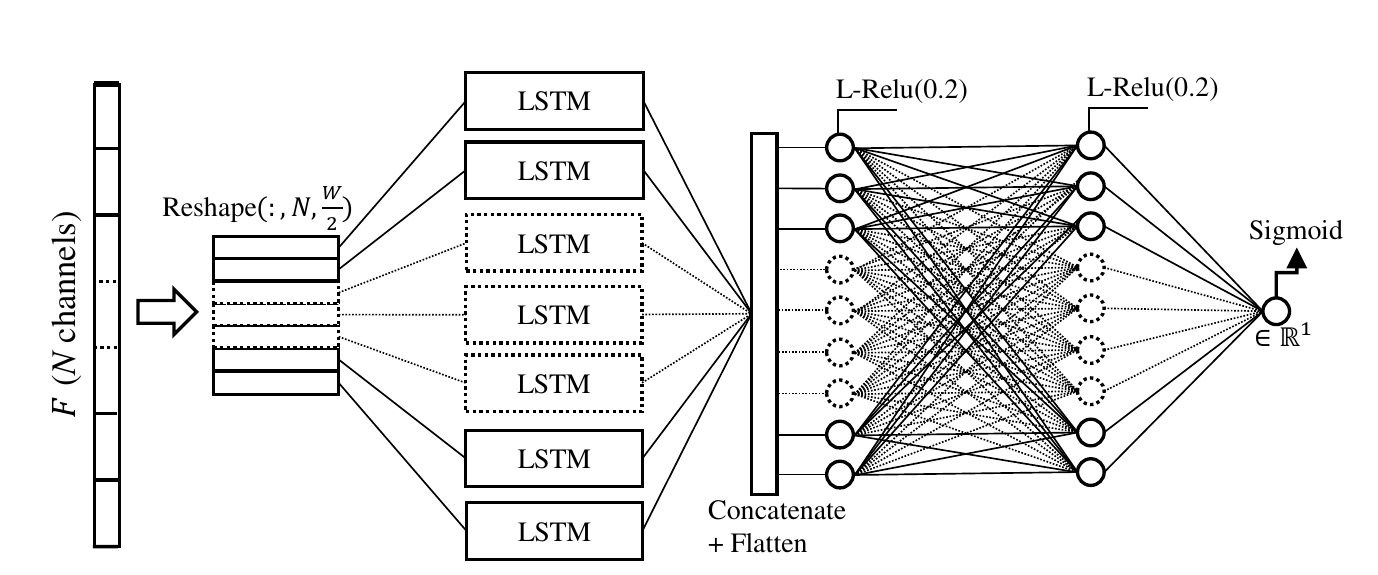}
        \caption {LSTM-based $\mathcal{D}$ architecture.}
        \label{Fig:LSTM_Net}
\end{subfigure}

\begin{subfigure}[b]{1\linewidth}
\centering
\includegraphics[width=0.66\linewidth]{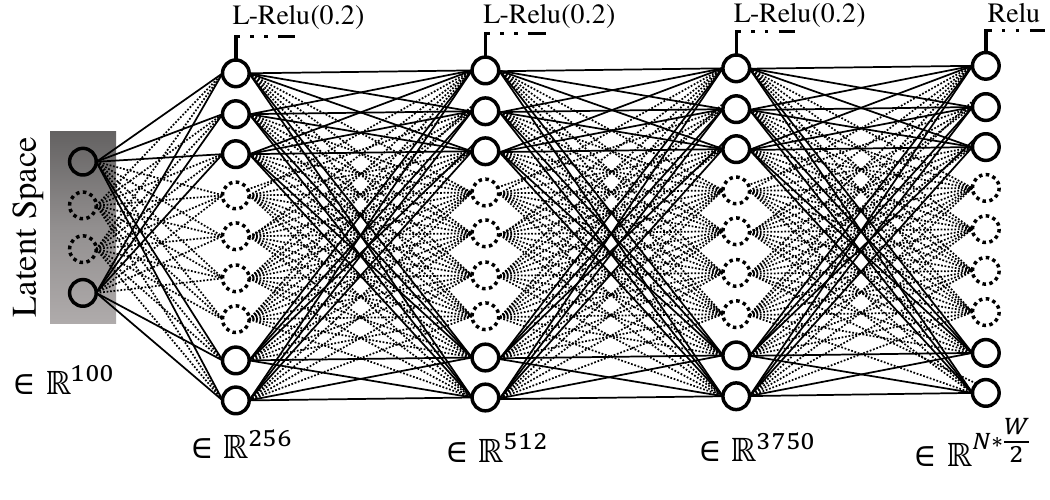}
        \caption {MLP-based $\mathcal{G}$ architecture.}
        \label{Fig:MLP_Net}
\end{subfigure}

\caption {Proposed GAN architecture.}
\label{Fig:GAN_Net}
\end{figure}

\noindent
The Long Short-Term Memory (LSTM), a recurrent neural network, was reported to decipher FFT-amplitude-based features efficiently and with high generalizability in SHM queries ~\cite{soleimani2021general,soleimani2022toward}. Therefore, the LSTM-based detector model is used in this study, as shown in Fig.~\ref{Fig:LSTM_Net}, consisting of one layer of LSTM at each branch. The channel-based architecture scales $\mathcal{D}$ to structures with different numbers of sensors. The Generator $\mathcal{G}$ architecture is outlined in Fig.~\ref{Fig:MLP_Net} as a Multi-Layer Perceptron (MLP) network. $\mathcal{G}$ samples from $P_{z}$ with 100 dimensions, followed by layers, up-scaling the latent sample to have the same dimensions as $F$ ($\in \mathbb{R}^{ \frac{W}{2}*N\times 1}$). In case $W/2*N\leq3750$, which occurs experimenting with different $W$ and $N$ values, $\mathcal{G}$ will have the first three layers with an output layer size of $W/2*N$.

With the Sigmoid activation function, the output neuron hypothetically outputs 0 for the most probable fake data and 1 for the most probable real data. From that output, we formulate an anomaly detection score, $S$, as:

\begin{equation}
\label{EQ:S_GAN}
    s_x=-\textrm{log}(\mathcal{D}(x)),
\end{equation}

where $s_x$ denotes an instance of $S$ for an arbitrary signal $x$. The range of $S$ is between 0 for the most probable real data to $+\infty$ for the most probable fake data.

\subsubsection{GAN Training}
\label{Sec:Source_Tuning}
\begin{figure}[t!]
\centering
\includegraphics[width=0.7\linewidth]{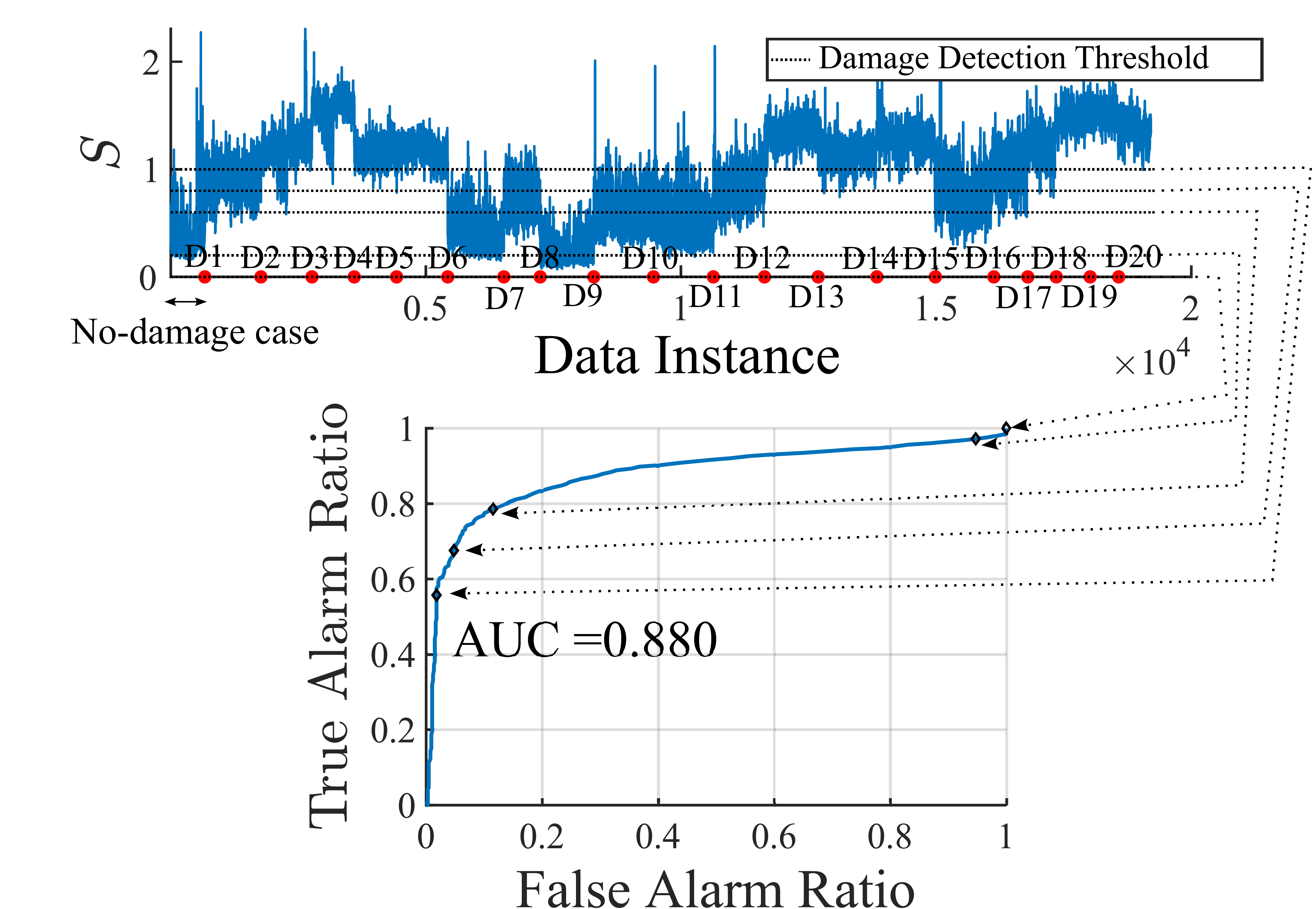}
        \caption {Constructing the ROC curve from the detection scores; a Yellow frame example.}
        \label{Fig:Score_To_Roc}
\end{figure}

\noindent
For training, we choose the ADAM optimizer~\cite{kingma2014adam} with a suggested learning rate of $1e-4$ and the first and second moment's decay rate of $0.9$ and $0.999$ to train GANs. GANs are mainly used for their generative capability, in which the equilibrium state between $\ell_{\mathcal{D}}$ and $\ell_{\mathcal{G}}$ signals the end of training. In the case of SHM, though, the trained $\mathcal{D}$ that offers the best discrimination between the source's no-damage and damage cases is sought. The main assumption of TL is that both no-damage and damage cases are known for the \textit{source domain}. During training various $\mathcal{D}$-$\mathcal{G}$ models will emerge per iteration (Fig.~\ref{Fig:GAN_Training_Scheme}). Different GAN training loss patterns from which $\mathcal{D}$ is drawn---\eg equilibrium or overfitted $\mathcal{D}$---might offer different sensitivity to damages~\cite{soleimani2022toward}. We train GAN until saturation and select the $\mathcal{D}$ that offers the best performance according to the area under the Receiver Operating Characteristics (ROC) curves (Area Under the Curve - AUC)---\ie highest true alarms and lowest false alarms over the source data. In ROC, the damage detection threshold is set at different values. The true alarm ratio is the Recall rate of the damage cases, the ratio of damage case data identified by the detector with $S$ over the threshold. The false alarm ratio is 1 – Recall of the no-damage case. Since the curve is constructed with all possible values of damage detection threshold $T$, the area under the curve (\ie AUC) offers a threshold-bias-free estimation of how well $\mathcal{D}$ contrasts those cases. $\textrm{AUC}=1$ indicates the perfect distinction between no-damage and damage cases' detection scores. In contrast, an $\textrm{AUC}=0.5$ indicates the random guess (\ie values below 0.5 are worse than a random guess: the model understood the data incorrectly). An instance of building the ROC curve from the detection scores, $S$, is shown in Fig.~\ref{Fig:Score_To_Roc} for a GAN trained on the Yellow Frame no-damage case with $W=1000$, and $N=10$.

\begin{figure}[t!]
\centering
\includegraphics[width=0.8\linewidth]{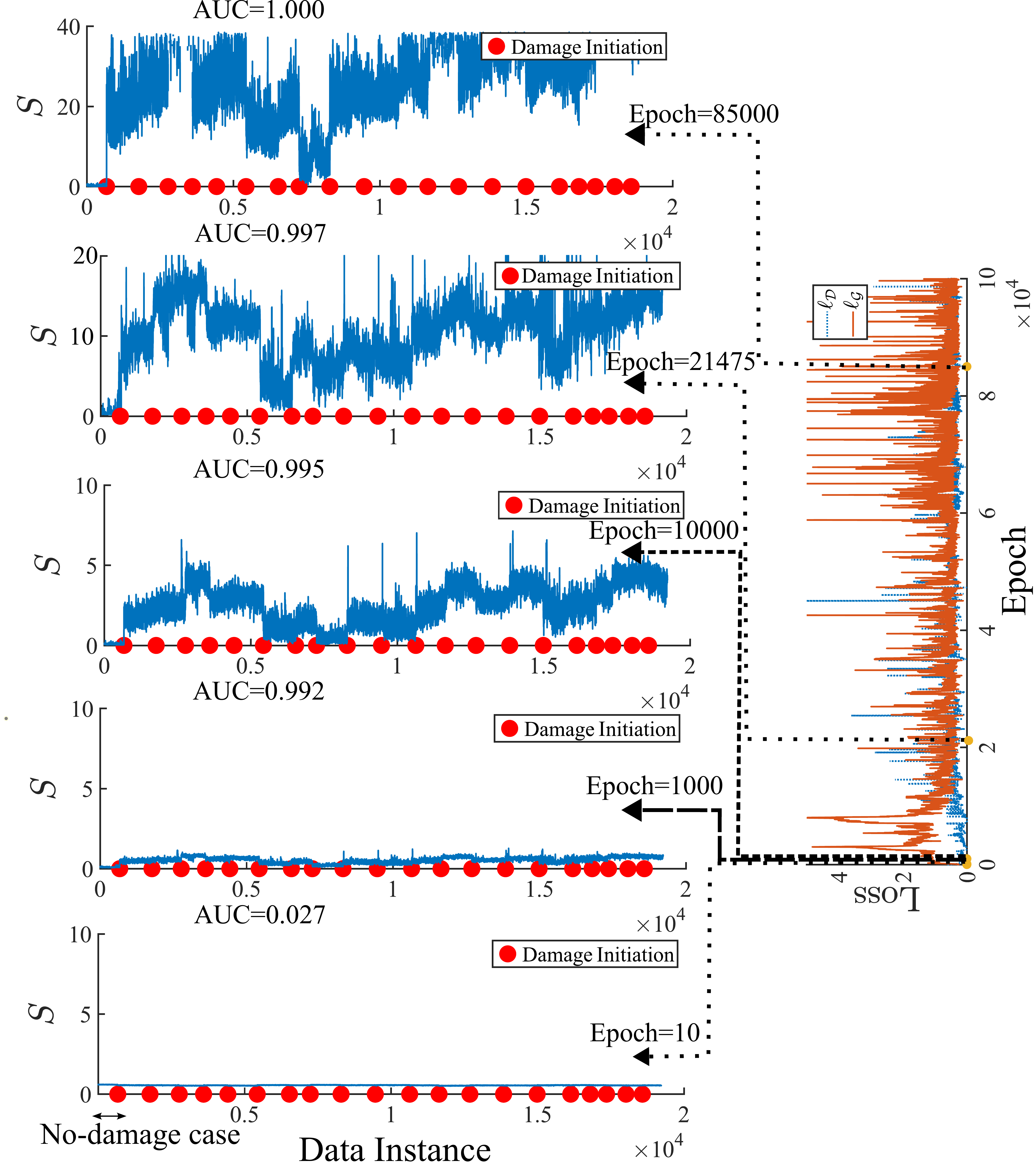}
        \caption {Evolution of $\mathcal{D}$'s differentiation between no-damage and damage cases during GAN's training; Yellow Frame dataset with $W=1000$ and $N=10$.}
        \label{Fig:Yellow_Step}
\end{figure}

The GAN training process, and the loss functions (Eqs.~\ref{Eq:D_Loss} and~\ref{Eq:G_Loss}), yield the $S$ of the no-damage case toward zero and the damage cases toward higher $S$ scores. A sample GAN's training result with 100000 epochs is shown in Fig.~\ref{Fig:Yellow_Step} for the Yellow Frame structure with $W=1000$ and $N=10$. A close inspection of the figure will reveal the fact that $\mathcal{D}$ reaches high AUC values discriminating no-damage from damage cases fairly fast, around the epoch 100. PyTorch~\cite{NEURIPS2019_9015} is used to carry out this study; thus, an $S$ of 40 (\ie $10^{-40}$) is the smallest PyTorch logarithmic Float precision. Consequently, as a rule of thumb, the source-no-damage identifier is selected as follows; the detector with the highest AUC without $S$ scores equals or above 40 (\ie float precision). Following the above criteria, Epoch 21475 is chosen for the Yellow Frames source-no-damage identifier. This source-no-damage identifier is labeled as YS10\_1000 (\ie Yellow Frame source with $N=10$ and $W=1000$); other source-no-damage identifiers are labeled accordingly throughout this study. 


\subsection{Domain Adaptation}
\label{Sec:Feature_Transfer}
\noindent
Domain-knowledge-based DA are more generalizable than their DL-based counterparts; the reason behind the current surge of physics-guided Machine Learning, specifically for mechanical problems~\cite{raymond2021applying,AMININIAKI2021113959}. In this study, and with the FFT-based features $F$, we set the frequency power spectrum as the grounds of the domain-knowledge-based DA method. Instead of direct manipulation of the target $F$, the digital signal processing (DSP) domain knowledge is used herein. The signal's power spectrum offers the relative magnitudes of frequency components that make up the signal upon combination. To estimate the power spectrum, the signal is windowed to smaller and sometimes overlapping frames (\eg Welche's method~\cite{welch1967use}). Herein, the source and target signals are framed with windows of size $W$ (Section~\ref{Sec:Feature}). The mean-square of normalized features of the source domain is taken as an estimation of the source domain power spectrum---``source spectrum'' (SS). Accordingly, the target power spectrum--- ''target spectrum'' (ST)---is created utilizing the target domain no-damage case data. A spectral line mapping, $\mathcal{M}$, is formed by transforming the target spectrum to the source spectrum. This transformation serves the core DA operation needed to use the source-no-damage identifier over the target vibration data. 

\begin{figure}[t!]
\begin{subfigure}[b]{0.49\linewidth}
\centering
 \includegraphics[width=0.8\linewidth]{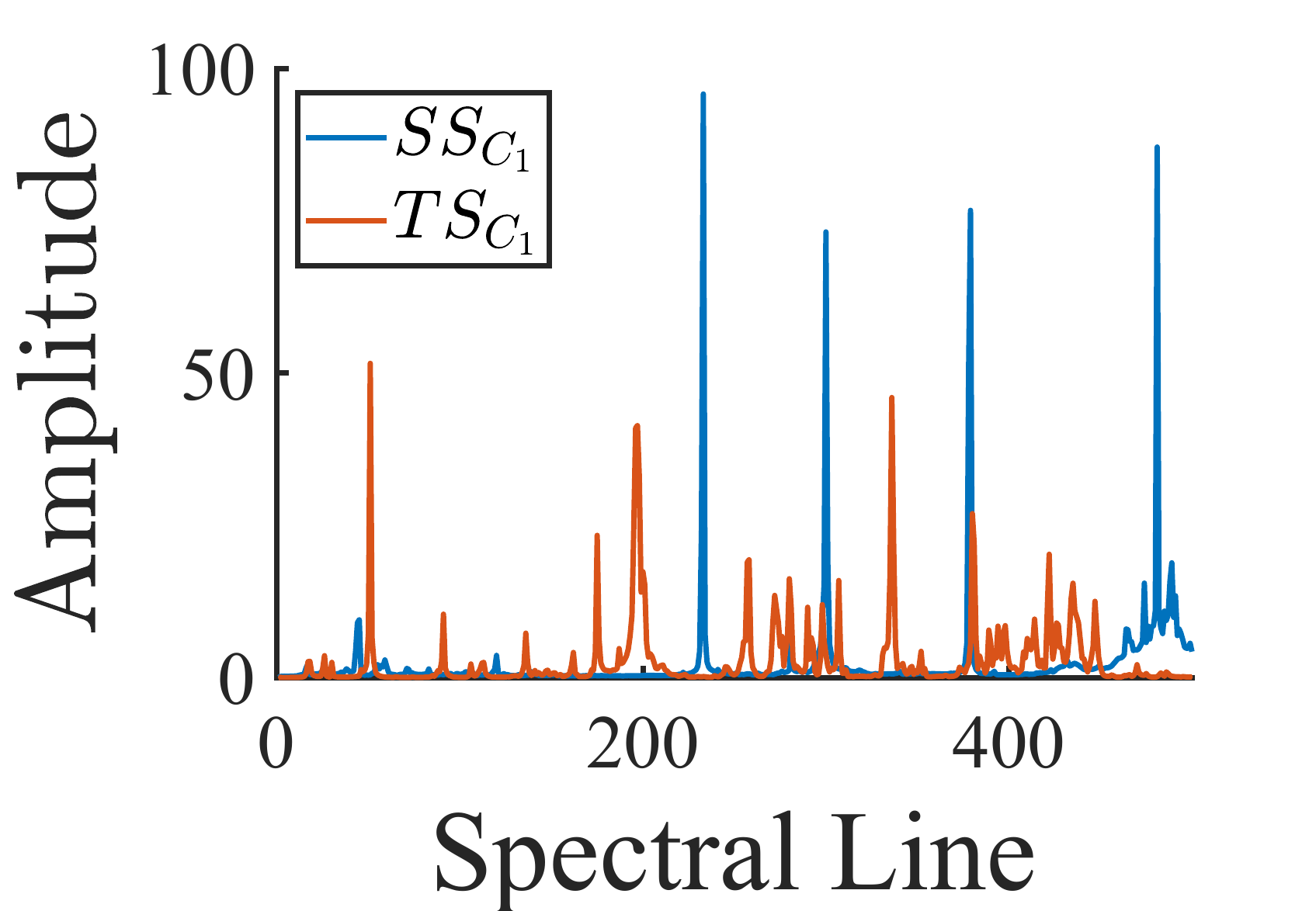}
        \caption {Source spectrum: Yellow Frame first channel, $W=1000$.}
        \label{Fig:Yellow_PS}   
\end{subfigure}
\begin{subfigure}[b]{0.49\linewidth} 
\centering
 \includegraphics[width=0.8\linewidth]{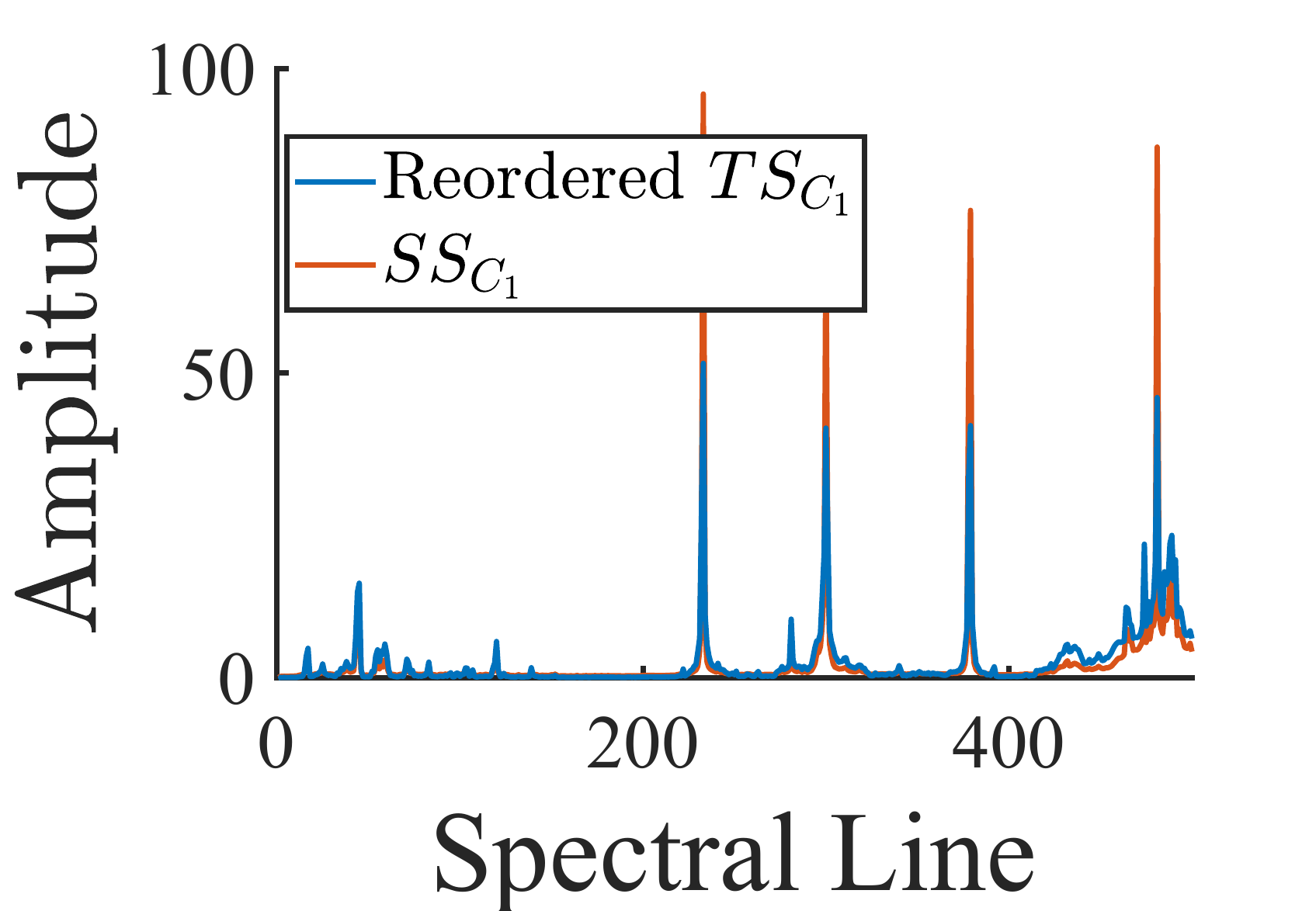}
        \caption {Target spectrum: QUGS first channel, $W=1000$.}
        \label{Fig:Yellow_Order}   
\end{subfigure}

\begin{subfigure}[b]{0.49\linewidth} 
\centering
     \includegraphics[width=0.8\linewidth]{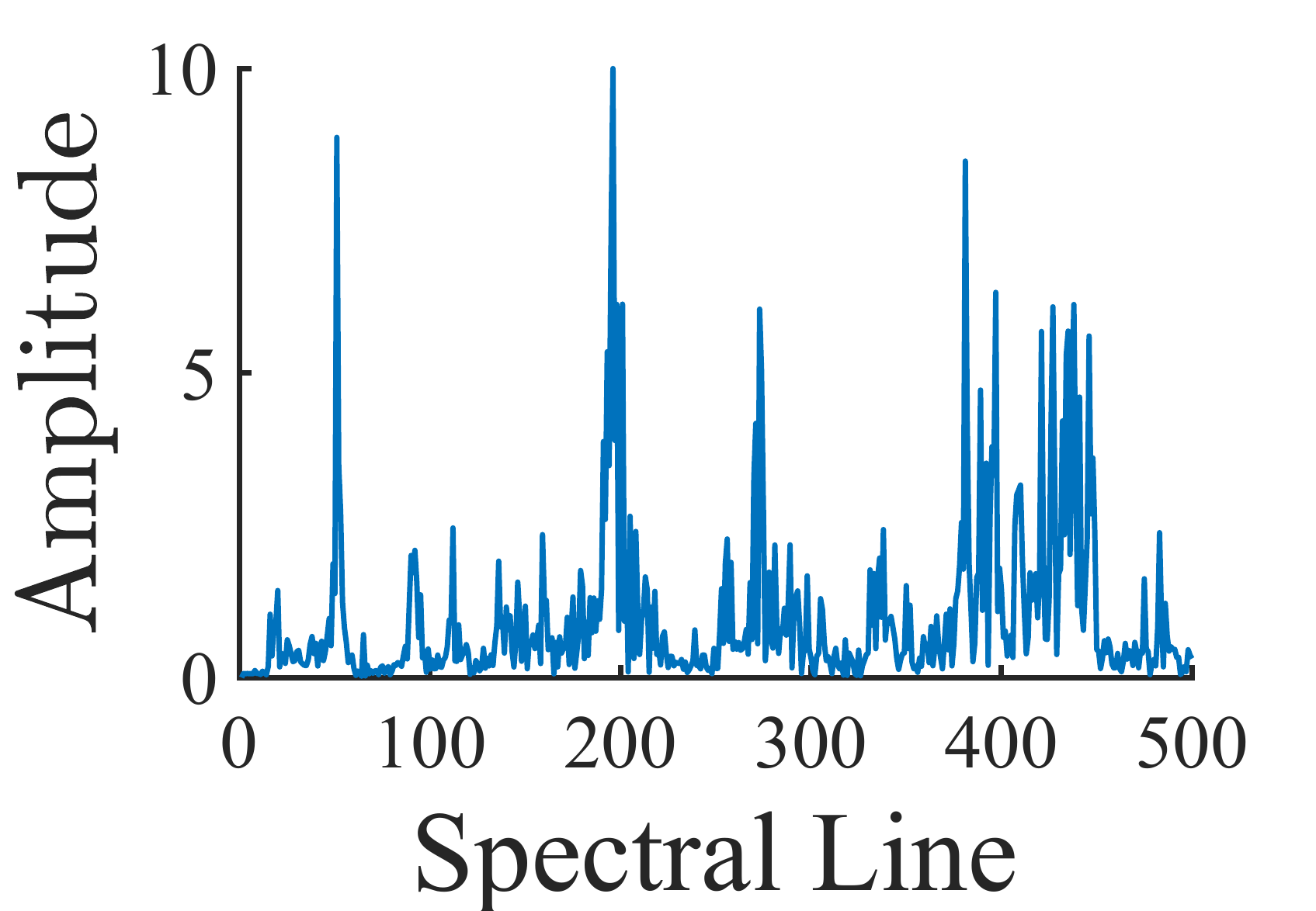}
    
        \caption {A sample $F$ from the QUGS first channel; $W=1000$ $N=10$.}
        \label{Fig:Yellow_S_F}   
\end{subfigure}
\begin{subfigure}[b]{0.49\linewidth} 
\centering
     \includegraphics[width=0.8\linewidth]{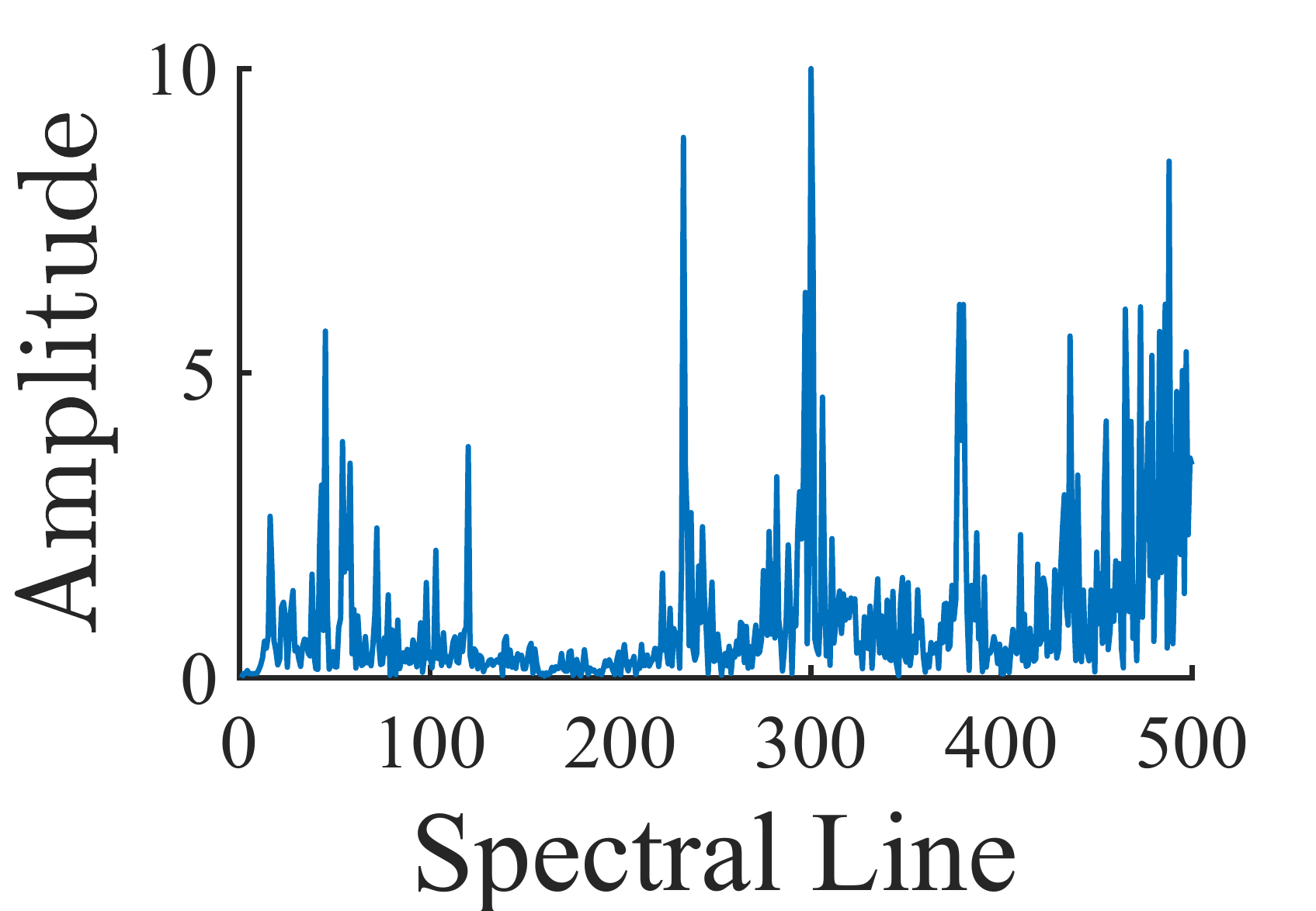}
        \caption {$F^{*}=\mathcal{M}(F)$ with $F$ shown in Fig.\ref{Fig:Yellow_S_F}.\\\phantom{}}
        \label{Fig:Yellow_S_FR}   
\end{subfigure}
\caption{Instance of DA application.}
\label{Fig:d1}
\end{figure}

A sample source spectrum for Yellow Frame with $W=1000$ and $N=10$ is shown for its first data channel, $SS_{C1}$, in Fig.~\ref{Fig:Yellow_PS}. A target spectrum for the QUGS's first data channel with $W=1000$ and $N=10$, $TS_{C1}$, is shown in Fig.~\ref{Fig:Yellow_PS}. The two spectra are different, depicting strongly heterogeneous systems with different vibrational content. In that regard, the following channel-wise mapping is proposed. First, the spectral line \textit{positions} of both source and target domain no-damage cases (\ie from 1 to $W/2$) are sorted according to their amplitudes (\ie argsort), named $Arg_S$ and $Arg_T$, respectively. The sorting order (\ie ascending or descending) should be the same for both domains. The argsort operator defines the mapping (\ie reordering of the target frequency spectrum), in which target spectral lines at positions $Arg_{T_{i}}$ are moved to the positions $Arg_{S{i}}$. The procedure is recapped in Algorithm~\ref{Algo:swap}. $\mathcal{M}$ is pictorially shown with target QUGS and source Yellow Frame with $W=1000$ for the first data channel in Fig.~\ref{Fig:Yellow_Order}. As shown in the figures, with reordering, source and target power spectra align but with different amplitudes. Therefore, a post-reordering scaling multiplier, $C_{S-T}$ (Algorithm~\ref{Algo:FS}), is applied to them. The correction of amplitudes pursuing the reordering is necessary since spectral lines' relative amplitudes demonstrate their contribution to the structure's vibrational content. With reordering, target spectral lines are ordered the same as the source domain regarding their relative amplitudes. The argsort and scaling multiplier values describe the mapping $\mathcal{M}$. $\mathcal{M}$ is obtained with the source and target domain's no-damage cases and is applied to any incoming data $F$, transforming it into $F^{*}$ (Algorithm~\ref{Algo:FS}). In brief, if the target (new) data has no damages, the mapping is expected to create a spectral line similar to the no-damage case in the source domain, and thus, the detector can easily classify it as the real (no damage); otherwise, as fake (damage case). A sample $F$ and $F^{*}$ pairs are shown in Figs.~\ref{Fig:Yellow_S_F}, and~\ref{Fig:Yellow_S_FR}.

\begin{algorithm}[H]
 \caption{DA: obtaining channel $i$'s spectral line mapping, $\mathcal{M}_i$}
 \label{Algo:swap}
\SetAlgoLined
\KwInput{Channel i's source spectrum $SS_{C_i}$, Channel i's target spectrum  $TS_{C_i}$}
\KwOutput{Mapping $\mathcal{M}_i$}
$Arg_S\gets argsort(SS_{C_i})$\\
$Arg_T\gets argsort(TS_{C_i})$\\
$C_{ST} \gets \sqrt{\frac{SS_{C{i}}(Arg_{S})}{TS_{C_{i}}(Arg_{T})}}$\\
$\mathcal{M}_i\gets Arg_{S}, Arg_{T}, C_{ST}$ \\
\end{algorithm}

\begin{algorithm}[H]
 \caption{DA: transforming incoming data features.}
 \label{Algo:FS}
\SetAlgoLined
\KwInput{Channel $i$'s mapping $\mathcal{M}_i$, Channel $i$'s incoming data features $F_i$}
\KwOutput{$F^{*}_i$}
$Arg_{S}, Arg_{T}, C_{ST}\gets \mathcal{M}_i $\\
${F_{i}}^{*}(Arg_{S})=F(Arg_{T})*C_{ST}$
\end{algorithm}

\section{Experimental Setup}
\label{sec:perf}
\noindent
This study decouples the differentiation between no-damage and damage cases from threshold tuning since devising a tuning standard is a source of bias. That bias is the reason for using control charts, like ROC curves, to obtain a bias-free estimation of how well a detector model performs. We argue that damage detection will become a plain procedure if the no-damage and damage cases are well-separated. Given the reasoning above, the performance assessment of the proposed TL method is divided into three parts. First, the performance of the source-no-damage identifier on the source domain. Given the GAN and signals' full-spectrum FFT amplitudes, the source-no-damage identifier is expected to distinguish between no-damage and damage cases in the source domain. The second assessment is regarding the devised DA method: whether the target domain features can be mapped to the source domain in such a way that the source-no-damage identifier discerns between the target's no-damage and unseen damage cases. Third, a threshold tuning method is introduced to perform and evaluate such zero-shot detection accuracy of the proposed approach. Details of those analyses are given in this section, and results are provided in the Results section.

\begin{figure}[t!]
\centering

  \includegraphics[width=0.9\linewidth]{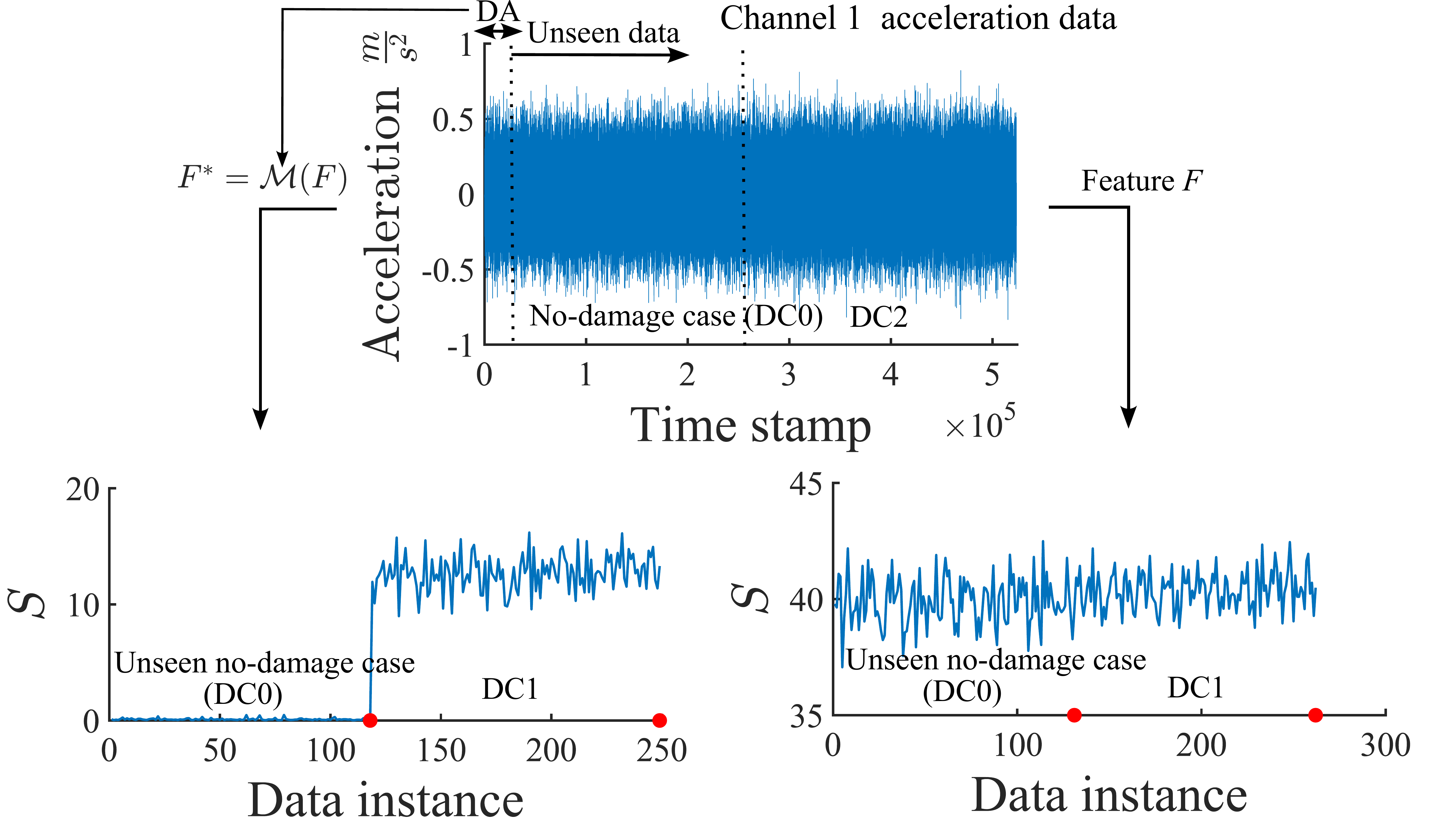}
\caption{$S$ given by the source-no-damage identifier with and without DA; $W=2000$, $N=15$.}
\label{Fig:Perf_Eval} 
\end{figure}

\begin{figure}[t!]
\centering

  \includegraphics[width=0.48\linewidth]{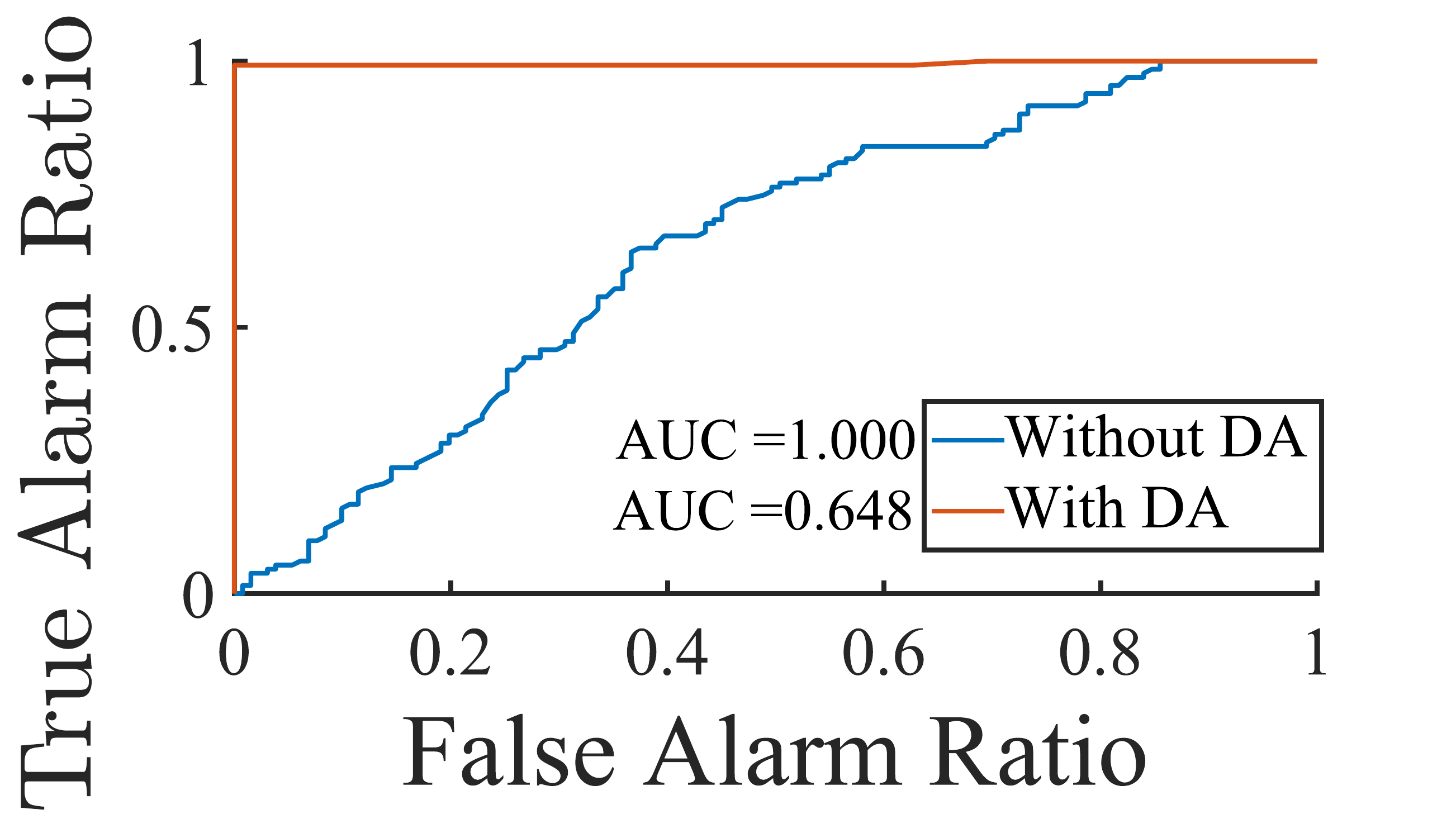}
\caption{Evaluating DA through ROC curves given Fig.~\ref{Fig:Perf_Eval}.}
\label{Fig:ROC_Perf_Eval} 
\end{figure}

Regarding the source-no-damage identifier performance on the source domain, the first evaluation will assess whether $\mathcal{D}$ can reach a high AUC score over the source data. Then to evaluate the DA method ($i$) the proposed spectral mapping $\mathcal{M}$ is accomplished over the target's no-damage case, and ($ii$) $\mathcal{M}$ transforms the (unseen) target data $F$, to $F^*$ (Section~\ref{Sec:Feature_Transfer}). Different combinations of no-damage/damage cases for each target dataset are formed for evaluation. For instance, for QUGS, there are 30 damage cases (DC1-DC30); each represents the damage in one of the 30 joints. Thus, 30 no-damage/damage scenarios are formed to test the proposed approach. We use 10\% of the available no-damage case data to perform the DA, and the rest is used for testing. Suppose the DC1 damage case is sought to be identified within target QUGS through the source: Yellow Frame; the process is shown in Fig.~\ref{Fig:Perf_Eval}. DA is performed by taking the first 10\% of each channel's data (the first channel is shown as an example) to define each channel's $\mathcal{M}$. The formed mapping transforms incoming data features $F$ into $F^*$. Inputting $F^*$ to the source-no-damage identifier results in $S$ shown in Fig.~\ref{Fig:Perf_Eval}'s lower left image. Without DA, inputting target $F$ to the source-no-damage identifier produces $S$ on the lower right image. Obviously, without DA, the source-no-damage detector $\mathcal{D}$ cannot distinguish between the no-damage and damage cases' target data (Fig.~\ref{Fig:ROC_Perf_Eval})--- \ie observes both as novelties regarding its source domain---outputting significantly high $S$ for them.

The threshold tuning method is outlined in Algorithm~\ref{alg:2}. Previously, 10\% of the no-damage case data were used for DA. Since dataset sizes are known for evaluating the accuracy, the next 40\% data from the no-damage case is used to tune the anomaly detection threshold (Algorithm~\ref{alg:2}). First, given $\mathcal{M}$. obtained through DA over the 10\% of no-damage case data, features $F$ are transformed to $F^*$ for all channels (Alg~\ref{Algo:FS}). Then, 40\% of the target no-damage case data scores are used for tuning. The Normal distribution is fit to that chunk's detection scores. Mean, $\mu$, and standard deviation, std, of the distribution are estimated, and $T$ is assigned as $\mu+3\textrm{std}$. Yet, abiding by the GAN's training paradigm, $T$ takes the maximum of $\mu+3\textrm{std}$ and $-log10(0.5)=0.3$, which is the $S$ of GAN's sigmoid activation for the uncertain case of fake and real data. Finally, precision, recall, and F1 scores evaluate the target damage detection performance.

\begin{algorithm}[H]
 \caption{Threshold tuning for zero-shot damage detection.}
 \label{alg:2}
\SetAlgoLined
\KwInput{$\mathcal{M}$, $F$ from target's no-damage case.}
\KwOutput{Damage detection threshold $T$}
$F^*=\mathcal{M}(F)$\\
mean, std =Normfit($-\log 10(\mathcal{D}(F^{*})))$  {Fitting a Gaussian distribution to the $S$ of $F^*$}\\
$T=max(\mu+3*\textrm{std}, -log10(0.5))$\\
\end{algorithm}

$N$ and $W$ are the only TL parameters; accordingly, TL scenarios reported in Table~\ref{Table:TL_Scenarios} are designed to perform the above evaluations across the three datasets. As outlined in this table, the target and source domain data should be derived with the same $W$ and $N$; parameters which are readily adoptable by users. Regarding that table, a total of 57 (\ie Yellow Frame: 20, QUGS: 30, and Z24: 7) damage cases exist. Thus, 57 combinations of no-damage/damage cases are available for evaluation. Given three $W$ and $N$ values and two source structures for each target structure, $57*3*3*3=1026$ damage identification/detection instance exists with hundreds of data points (\eg Fig.~\ref{Fig:Perf_Eval}). The Z24 dataset has the lowest number of data samples among the others, given its shorter data acquisition time and sampling frequency of 100 Hz (1000 Hz for the other datasets). Accordingly, 10\% of the Z24 no-damage case is allocated to perform the DA consisting of only 3 data points given $W=2000$. Accordingly, 50\% of the no-damage case data is given to the DA algorithm. Algorithm~\ref{alg:2} remains the same regarding the SDD, allocating the other 40\% of the Z24 no-damage case for threshold tuning. The remaining 10\% no-damage and damage cases remain for testing true and false alarms. Concerning the different sampling frequencies, we argue that the mapping is possible. The reason is that the relative position of spectral lines and not their actual frequency is the input of the $\mathcal{D}$ (\ie feature extraction in Section~\ref{Sec:Feature}). Z24 with a sampling frequency of 100 HZ or the Yellow frame with a sampling frequency of 1000 Hz, both span the same dimensions in $F$.

\begin{sidewaystable}[t!]
\begin{threeparttable}

  \caption{TL scenarios between Z24, Yellow Frame, and QUGS datasets.}
  \label{Table:TL_Scenarios}
  \centering
  \begin{tabular*}{1\linewidth}{@{\extracolsep{\fill}} lcccccccc}
    \toprule
\parbox{ 0.5cm}{Transfer label} & Source&Target &\parbox{0.5cm}{\centering $W$} &\parbox{3cm}{\centering Source sensors}&\parbox{3cm}{\centering Target  sensors}\\
    \midrule

Y2Q15\_2000&Yellow F. & QUGS& 2000 &15: All channels in Fig.~\ref{Fig:YF_Str}&15: Odd channels in Fig.~\ref{Fig:QUGS_Structure}\\
Y2Q15\_1000&Yellow F. & QUGS& 1000& 15: All channels in Fig.~\ref{Fig:YF_Str}&15: Odd channels in Fig.~\ref{Fig:QUGS_Structure}\\
Y2Q15\_500&Yellow F. & QUGS& 500	& 15: All channels in Fig.~\ref{Fig:YF_Str}&15: Odd channels in Fig.~\ref{Fig:QUGS_Structure}\\

Y2Q10\_2000&Yellow F. & QUGS& 2000& 10: Channels 1,3,5,6,8,9,10,12,13,15 (Fig.~\ref{Fig:YF_Str})&10: Channels 1,3,5,9,13,17,21,25,27,29 (Fig.~\ref{Fig:QUGS_Structure})\\
Y2Q10\_1000&Yellow F. & QUGS& 1000& 10: Channels 1,3,5,6,8,9,10,12,13,15 (Fig.~\ref{Fig:YF_Str})&10: Channels 1,3,5,9,13,17,21,25,27,29 (Fig.~\ref{Fig:QUGS_Structure})\\
Y2Q10\_500&Yellow F. & QUGS& 500	& 10: Channels 1,3,5,6,8,9,10,12,13,15 (Fig.~\ref{Fig:YF_Str})&10: Channels 1,3,5,9,13,17,21,25,27,29 (Fig.~\ref{Fig:QUGS_Structure})\\

Y2Q6\_2000&Yellow F. & QUGS&2000& 6: Channels 1,3,6,10,12,14 (Fig.~\ref{Fig:YF_Str})&6: 1,5,13,17,25,29 (Fig.~\ref{Fig:QUGS_Structure})\\
Y2Q6\_1000&Yellow F. & QUGS&1000& 6: Channels 1,3,6,10,12,14 (Fig.~\ref{Fig:YF_Str})&6: 1,5,13,17,25,29 (Fig.~\ref{Fig:QUGS_Structure})\\
Y2Q6\_500&Yellow F. & QUGS&500	& 6: Channels 1,3,6,10,12,14 (Fig.~\ref{Fig:YF_Str})&6: 1,5,13,17,25,29 (Fig.~\ref{Fig:QUGS_Structure})\\

Y2Z15\_2000&Yellow F. & Z24& 2000& 15: All channels in Fig.~\ref{Fig:YF_Str}&15: All channels in Table~\ref{Table:Z24_Setups}\\
Y2Z15\_1000&Yellow F. & Z24& 1000& 15: All channels in Fig.~\ref{Fig:YF_Str}&15: All channels in Table~\ref{Table:Z24_Setups}\\
Y2Z15\_500&Yellow F. & Z24& 500	& 15: All channels in Fig.~\ref{Fig:YF_Str}&15: All channels in Table~\ref{Table:Z24_Setups}\\

Y2Z10\_2000&Yellow F. & Z24& 2000& 10: Channels 1,3,5,6,8,9,10,12,13,15 (Fig.~\ref{Fig:YF_Str})&10: Setups 4,5, 6-channel 1 + the rest (Table~\ref{Table:Z24_Setups})\\
Y2Z10\_1000&Yellow F. & Z24& 1000& 10: Channels 1,3,5,6,8,9,10,12,13,15 (Fig.~\ref{Fig:YF_Str})&10: Setups 4,5, 6-channel 1 + the rest (Table~\ref{Table:Z24_Setups})\\
Y2Z10\_500&Yellow F. &Z24& 500& 10: Channels 1,3,5,6,8,9,10,12,13,15 (Fig.~\ref{Fig:YF_Str})&10: Setups 4,5, 6-channel 1 + the rest (Table~\ref{Table:Z24_Setups})\\

Y2Z6\_2000&Yellow F. & Z24& 2000& 6: Channels 1,3,6,10,12,14 (Fig.~\ref{Fig:YF_Str})&6: Setups 3,4,6,7-channel 1, setup 5-channels 1, 2 (Table~\ref{Table:Z24_Setups})\\
Y2Z6\_1000&Yellow F. & Z24& 1000&6: Channels 1,3,6,10,12,14 (Fig.~\ref{Fig:YF_Str})&6: Setups 3,4,6,7-channel 1, setup 5-channels 1, 2 (Table~\ref{Table:Z24_Setups})\\
Y2Z6\_500&Yellow F. & Z24& 500& 6: Channels 1,3,6,10,12,14 (Fig.~\ref{Fig:YF_Str})&6: Setups 3,4,6,7-channel 1, setup 5-channels 1, 2 (Table~\ref{Table:Z24_Setups})\\

Q2Z15\_2000&QUGS & Z24& 2000&15: 15: Odd channels in Fig.~\ref{Fig:QUGS_Structure}&15: All channels in Table~\ref{Table:Z24_Setups}\\
Q2Z15\_1000&QUGS& Z24& 1000&15: 15: Odd channels in Fig.~\ref{Fig:QUGS_Structure}&15: All channels in Table~\ref{Table:Z24_Setups}\\
Q2Z15\_500&QUGS & Z24& 500&15: 15: Odd channels in Fig.~\ref{Fig:QUGS_Structure}&15: All channels in Table~\ref{Table:Z24_Setups}\\

Q2Z10\_2000&QUGS & Z24& 2000& 10: Channels 1,3,5,9,13,17,21,25,27,29 (Fig.~\ref{Fig:QUGS_Structure})&10: Setups 4,5, 6-channel 1 + the rest (Table~\ref{Table:Z24_Setups})\\
Q2Z10\_1000&QUGS & Z24& 1000& 10: Channels 1,3,5,9,13,17,21,25,27,29 (Fig.~\ref{Fig:QUGS_Structure})&10: Setups 4,5, 6-channel 1 + the rest (Table~\ref{Table:Z24_Setups})\\
Q2Z10\_500&QUGS &Z24& 500& 10: Channels 1,3,5,9,13,17,21,25,27,29 (Fig.~\ref{Fig:QUGS_Structure})&10: Setups 4,5, 6-channel 1 + the rest (Table~\ref{Table:Z24_Setups})\\
Q2Z6\_2000&QUGS & Z24& 2000&6: 1,5,13,17,25,29 (Fig.~\ref{Fig:QUGS_Structure})&6: Setups 3,4,6,7-channel 1, setup 5-channels 1, 2 (Table~\ref{Table:Z24_Setups})\\
Q2Z6\_1000&QUGS& Z24& 1000&6: 1,5,13,17,25,29 (Fig.~\ref{Fig:QUGS_Structure})&6: Setups 3,4,6,7-channel 1, setup 5-channels 1, 2 (Table~\ref{Table:Z24_Setups})\\
Q2Z6\_500&QUGS & Z24& 500&6: 1,5,13,17,25,29 (Fig.~\ref{Fig:QUGS_Structure})&6: Setups 3,4,6,7-channel 1, setup 5-channels 1, 2 (Table~\ref{Table:Z24_Setups})\\

\bottomrule
\end{tabular*}
\begin{tablenotes}\footnotesize
\item[*] TL scenarios are reciprocal---\eg Y2Q15\_2000 means that Q2YS15\_2000 is also investigated.
\end{tablenotes}
\end{threeparttable}
\end{sidewaystable}

\section{Experimental Results}
\label{sec:res}
\noindent Three sets of evaluations are reported in this section: ($i$) the source-no-damage intensifier's performance over the source data, ($ii$) DA assessment on the target domain, and ($iii$) the zero-shot damage detection performance. The initial step is training the $\mathcal{D}$ models each for one of the TL scenarios (Table~\ref{Table:TL_Scenarios}). A mean AUC of 0.998 is obtained across all datasets (Section~\ref{Sec:Source_Tuning}). Accordingly, the first element of the TL model is deemed practical where each $\mathcal{D}$ model can differentiate between the source's no-damage and damage cases given various $W$ and $N$ values. The rest of this section concerns DA evaluation and performing the zero-shot damage detection.


\subsection{Target: The Yellow Frame}
\begin{figure}[h!]
\centering
\begin{subfigure}{0.49\linewidth} 
\centering
  \includegraphics[width=0.7\linewidth]{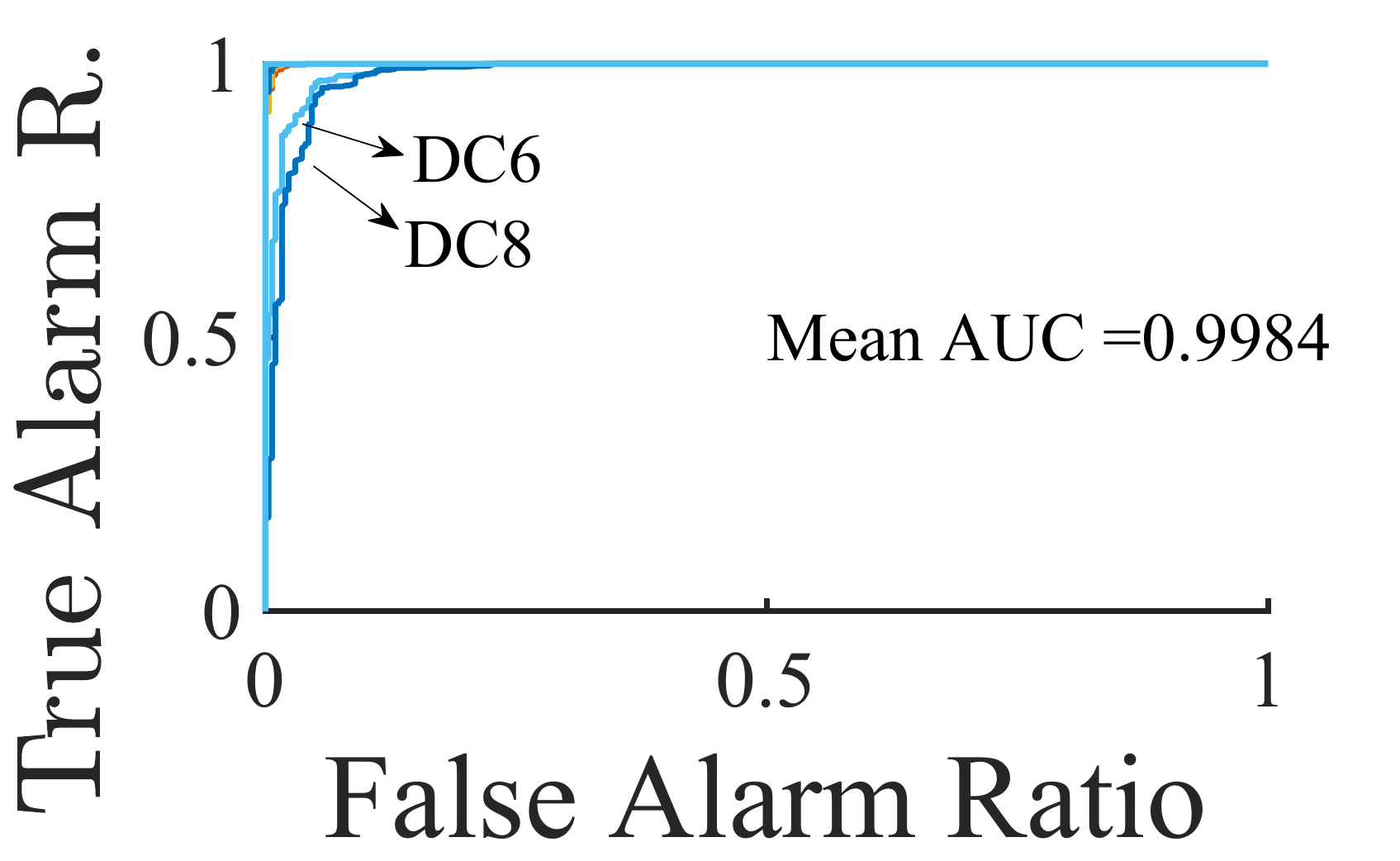}
        \caption {Q2YS15\_2000; ROC of 20 damages.}
        \label{Fig:Y_ROC_1}  
\end{subfigure}
\begin{subfigure}{0.49\linewidth} 
\centering
  \includegraphics[width=0.7\linewidth]{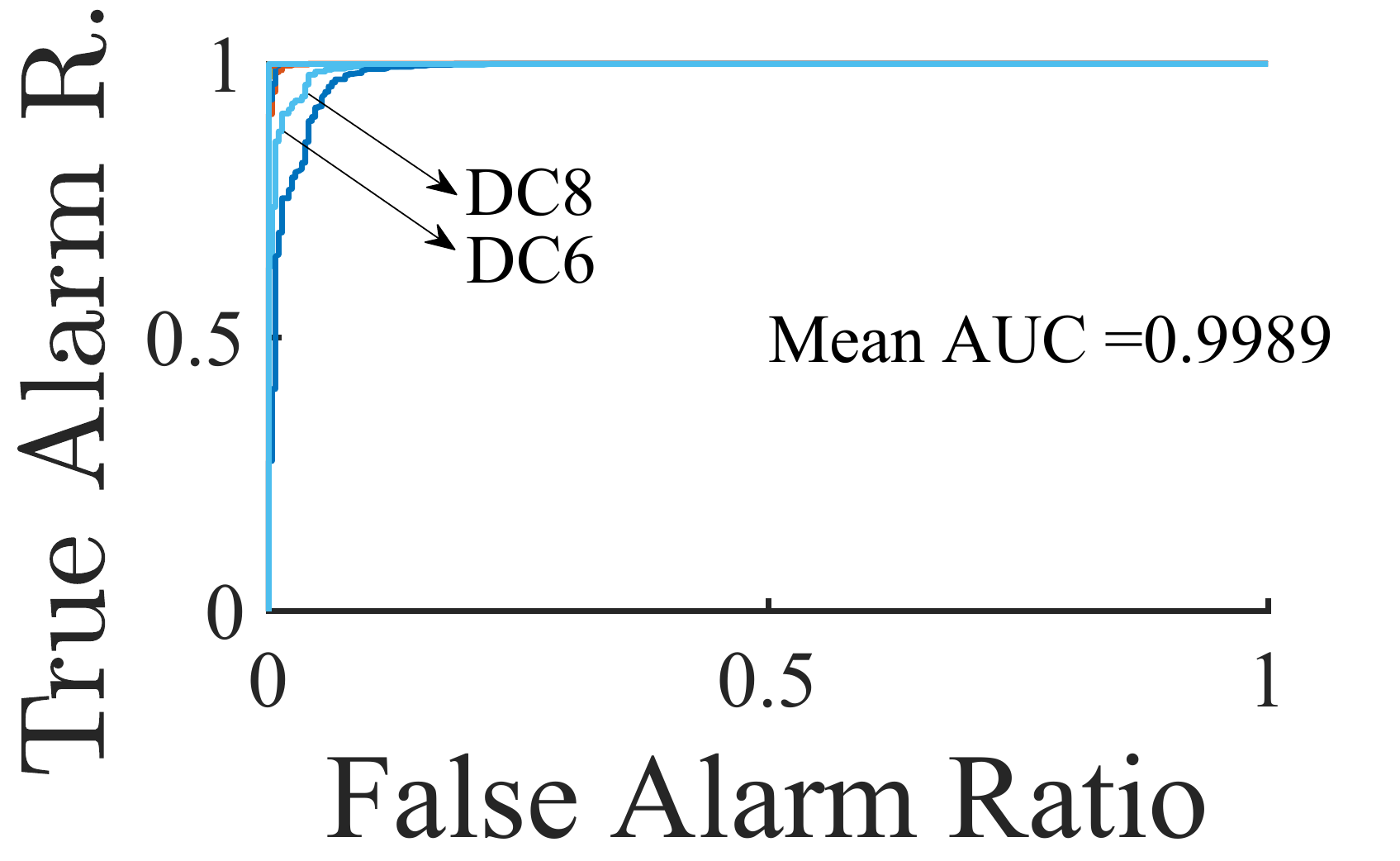}
        \caption {Z2YS6\_2000; ROC of 20 damages.}
        \label{Y_ROC_2}  
\end{subfigure}

\begin{subfigure}{0.49\linewidth} 
\centering
  \includegraphics[width=0.7\linewidth]{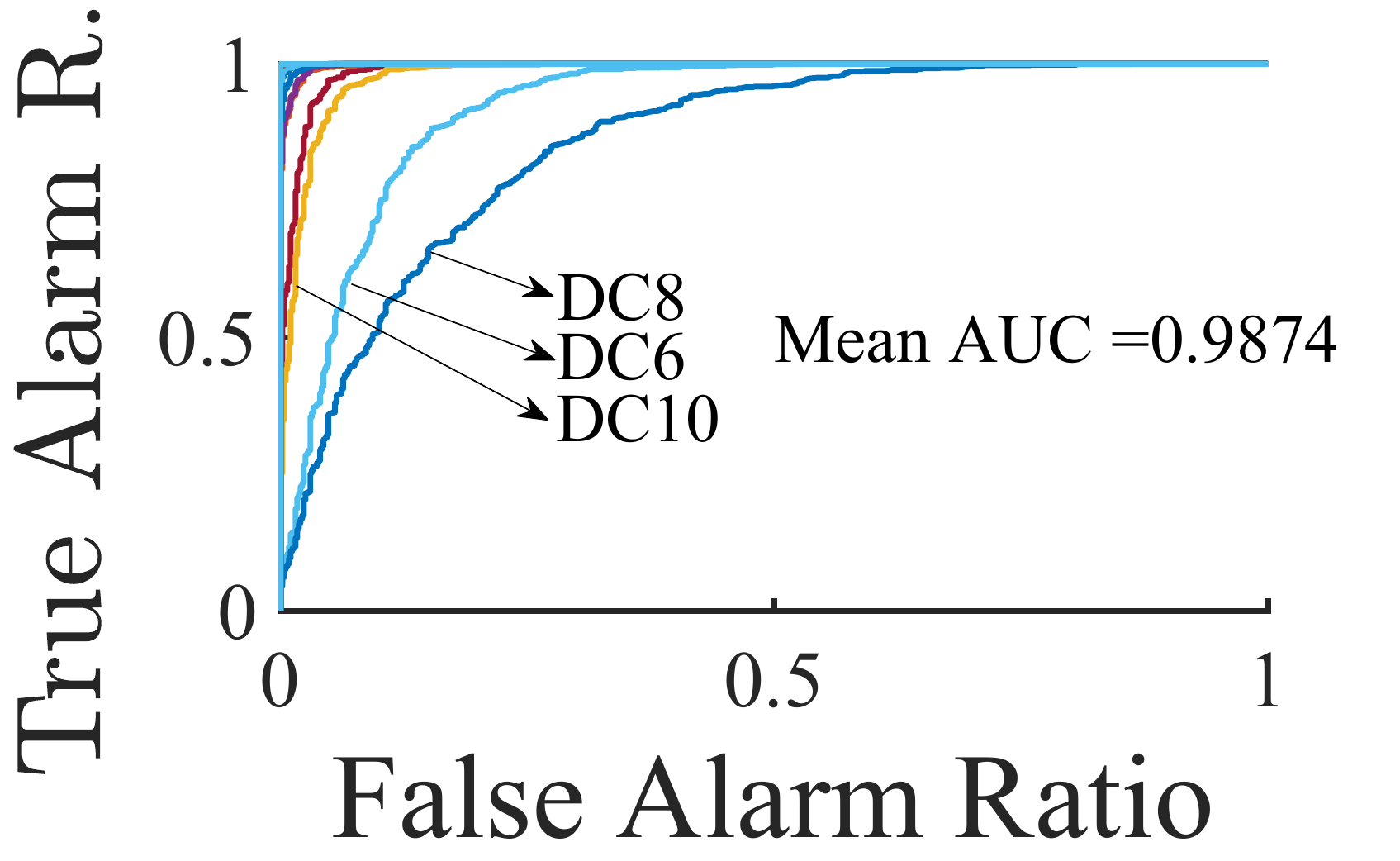}
        \caption {Z2YS10\_1000; ROC of 20 damages.}
        \label{Y_ROC_3}  
\end{subfigure}
\begin{subfigure}{0.49\linewidth} 
\centering
  \includegraphics[width=0.7\linewidth]{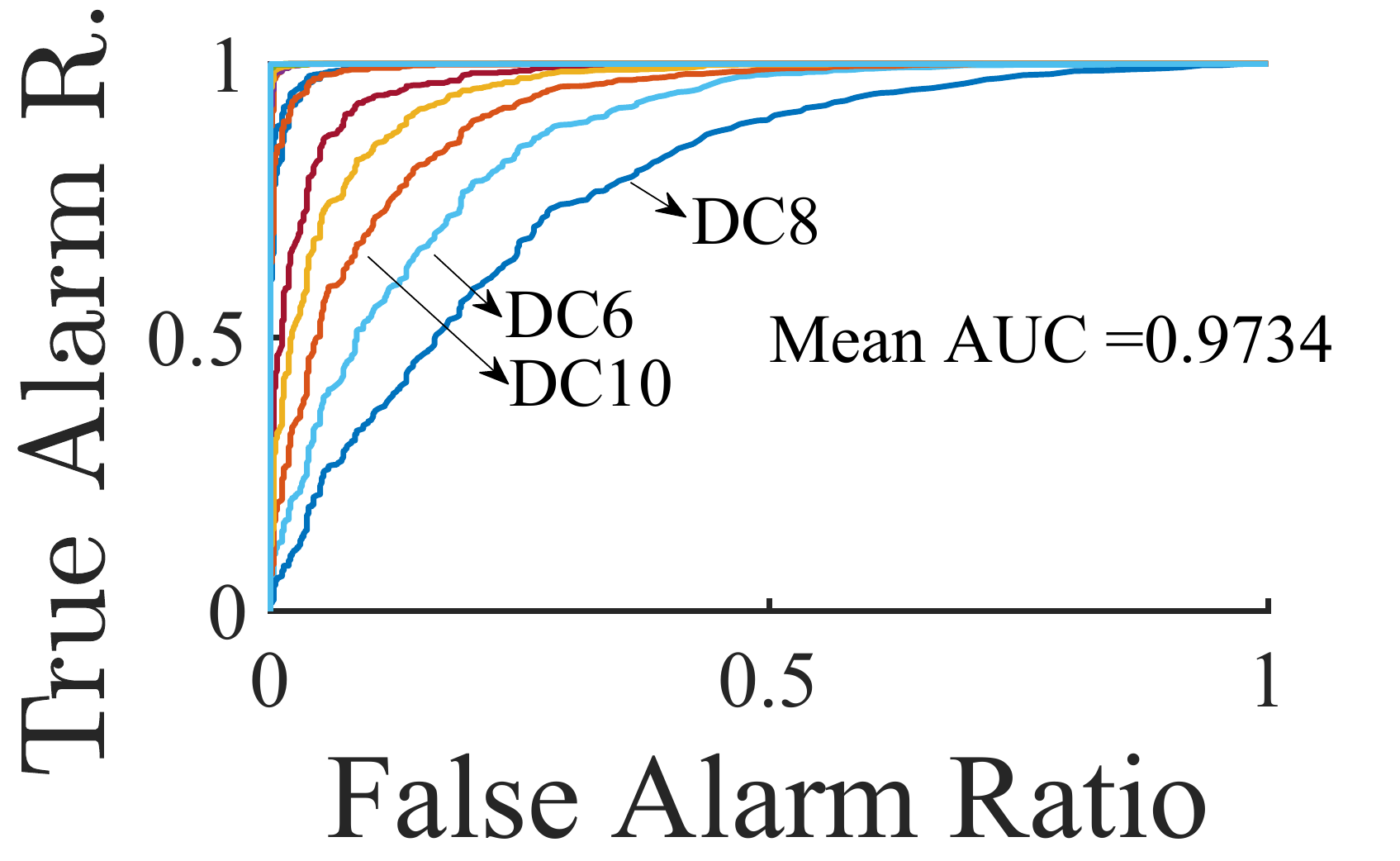}
        \caption {Q2YS6\_1000; ROC of 20 damages.}
        \label{Y_ROC_4}  
\end{subfigure}
\begin{subfigure}{0.49\linewidth} 
\centering
  \includegraphics[width=0.7\linewidth]{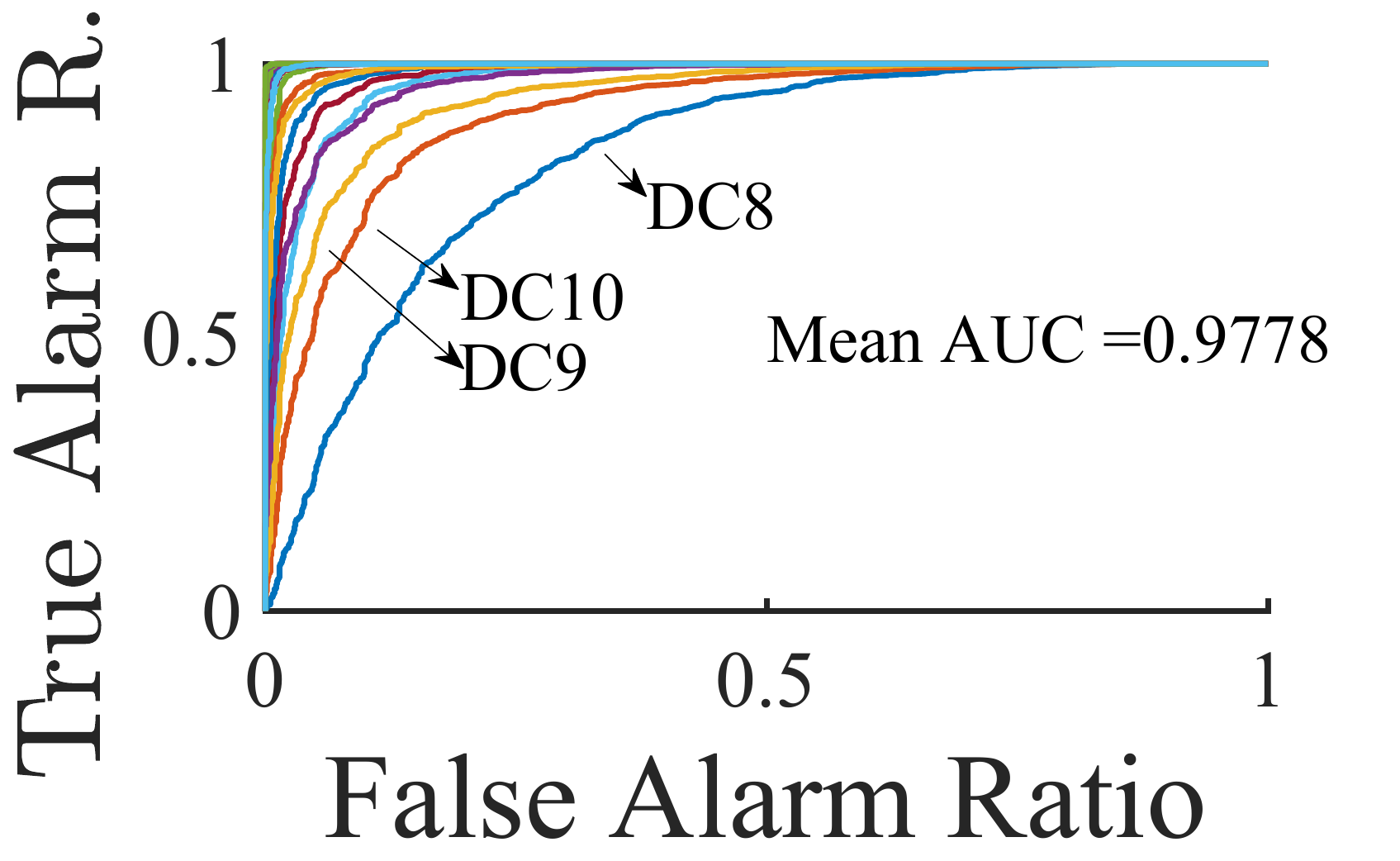}
        \caption {Q2YS15\_500; ROC of 20 damages.}
        \label{Y_ROC_5}  
\end{subfigure}
\begin{subfigure}{0.49\linewidth} 
\centering
  \includegraphics[width=0.7\linewidth]{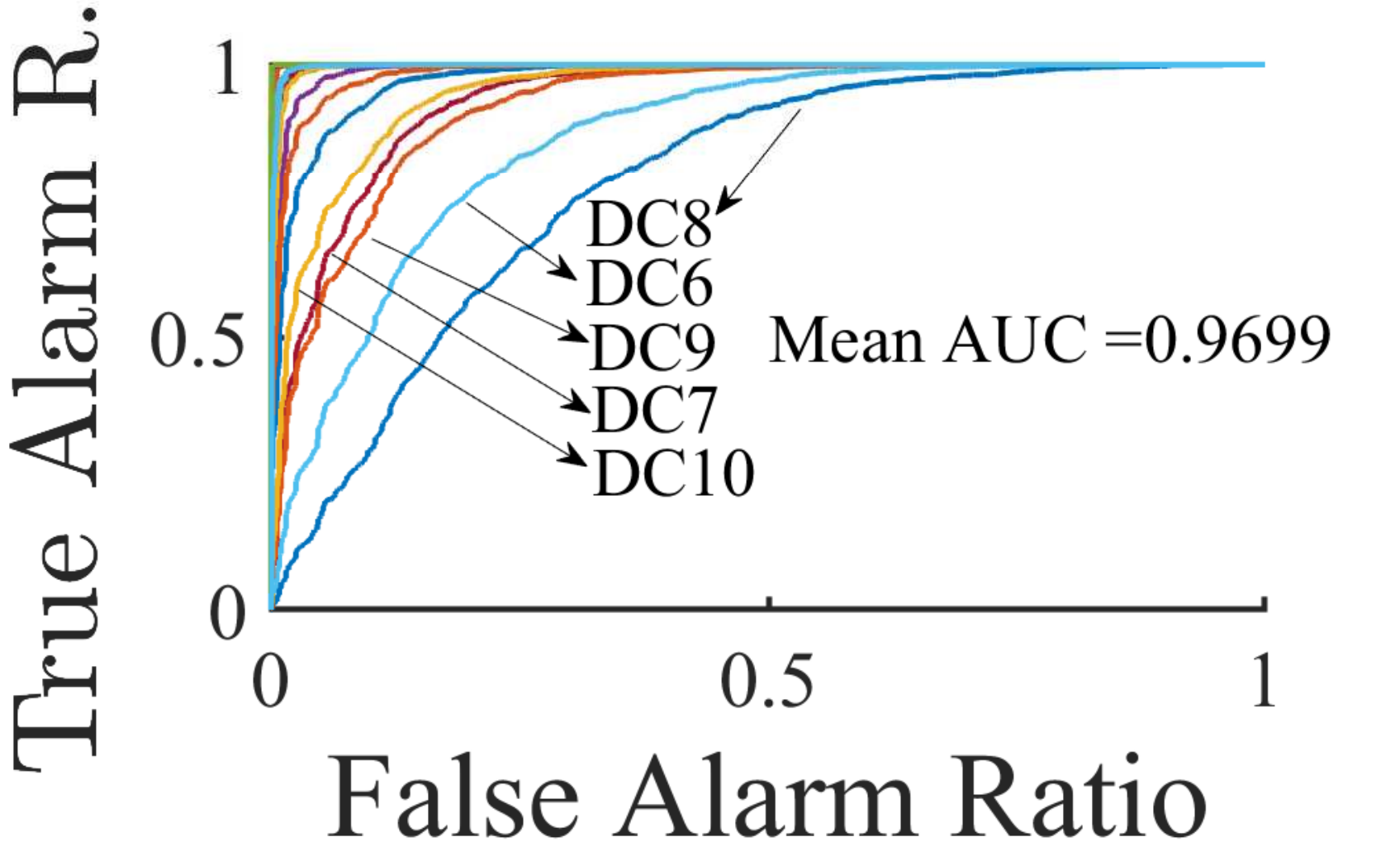}
        \caption {Z2YS6\_500; ROC of 20 damages.}
        \label{Y_ROC_6}  
\end{subfigure}
\caption{TL performance; Yellow Frame as the target domain.}
\label{Y_ROC_A}
\end{figure}

\noindent
The $\mathcal{D}$ models trained on QUGS and Z24 are evaluated over the Yellow Frame as the target per Section~\ref{Sec:Feature_Transfer}. ROC curves for 6 of the TL scenarios outlined in Table~\ref{Table:TL_Scenarios} are shown in Fig.~\ref{Y_ROC_A}. Each figure embodies 20 ROC curves---\ie 20 damage cases in Yellow Frame--- with the mean AUC values reported. With $W=2000$ and $N=15$ (Fig.~\ref{Fig:Y_ROC_1}), almost all ROC curves are ideal, with a mean AUC of 0.9984. DC6 and DC8 show a lower AUC than the others. Regarding parameters $W$ and $N$, lowering those values reduces not all but several damage cases' AUCs. With the lowest $W$ and $N$, DC6 and DC8 AUCs drop to values as low as 0.7 with QUGS as the source. Z24 shows better overall performance on the Yellow Frame. Results suggest that the TL performs better with more sensors (\ie higher $N$), which is trivial as more sensors increase the chance of capturing damage-related data. Yet, the notable observation is that with only six out of the initial 15 sensors ($W=2000$, Fig.~\ref{Y_ROC_2}), 18 out of 20 damages are distinguished from the no-damage case almost certainly (\ie the AUC is near 1) with Z24 as the source structure.

\begin{figure}[h!]
\centering
\begin{subfigure}{0.48\linewidth} 
\centering
  \includegraphics[width=1\linewidth]{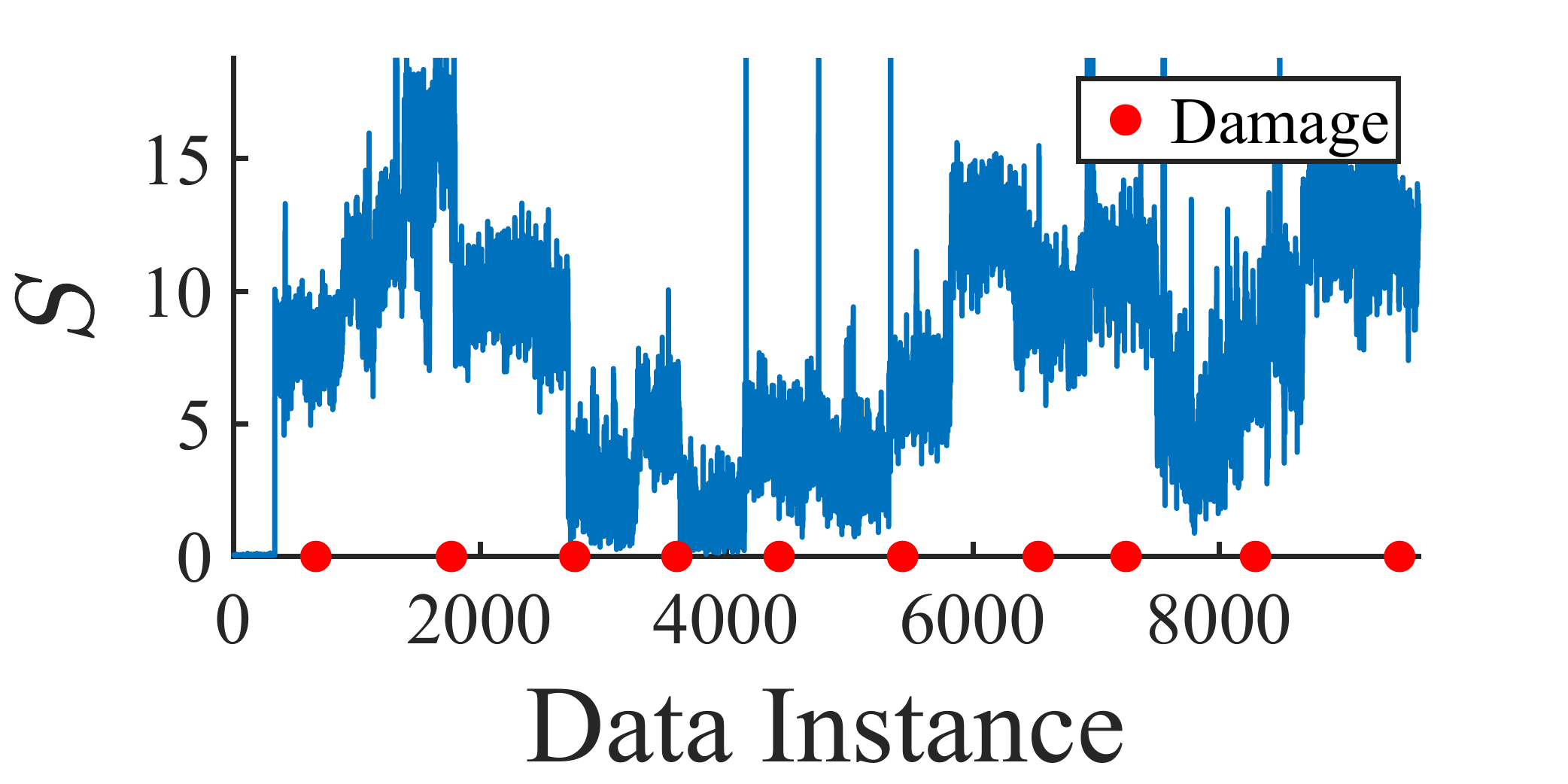}
        \caption {YS15\_2000 performance on Yellow Frame (no TL).}
        \label{Fig:Yellow_Self_1}  
\end{subfigure}
\begin{subfigure}{0.48\linewidth} 
\centering
  \includegraphics[width=1\linewidth]{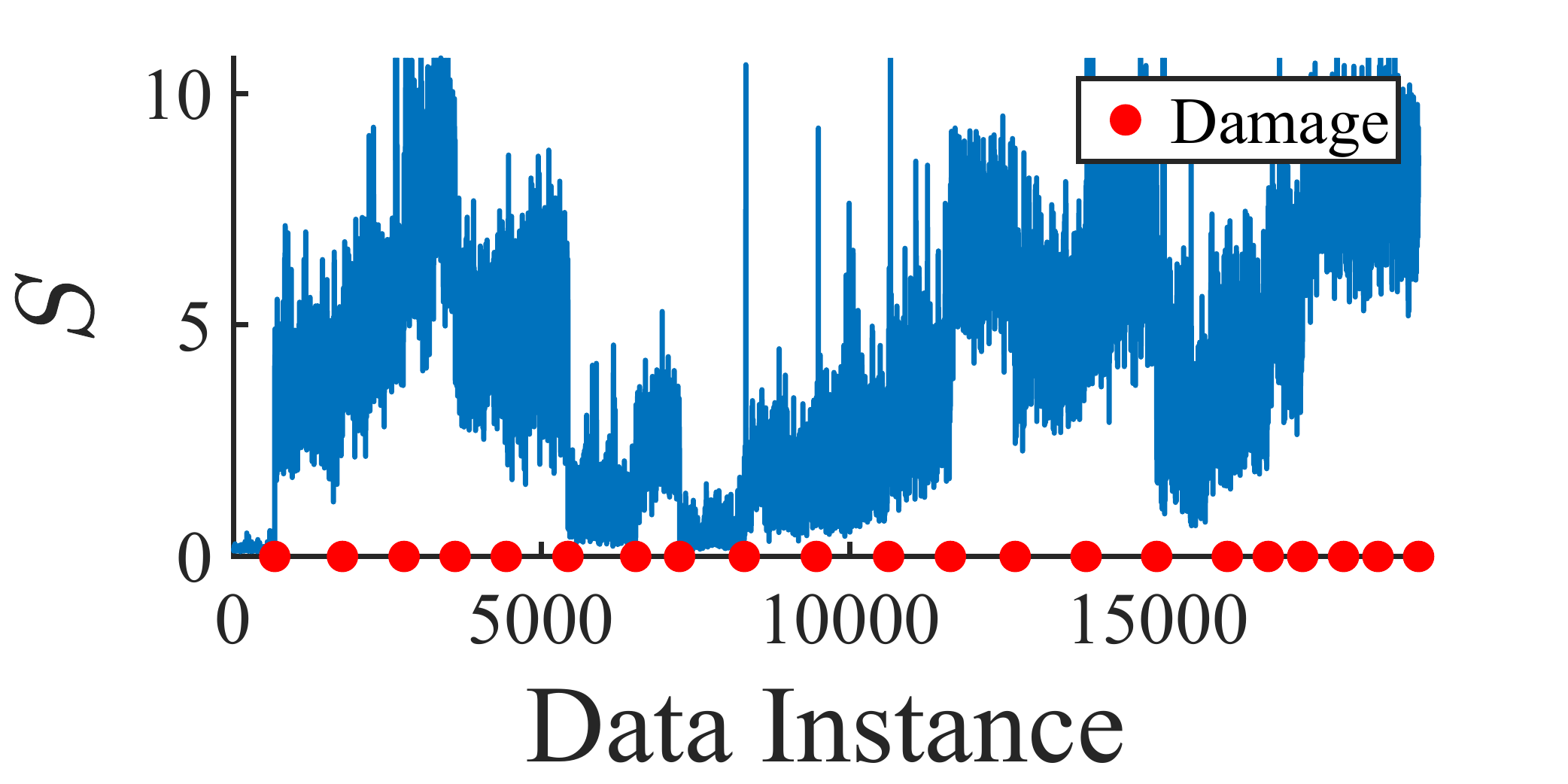}
        \caption {YS10\_1000 performance on Yellow Frame (no TL).}
        \label{Fig:Yellow_Self_2}  
\end{subfigure}
\begin{subfigure}{0.48\linewidth} 
\centering
  \includegraphics[width=1\linewidth]{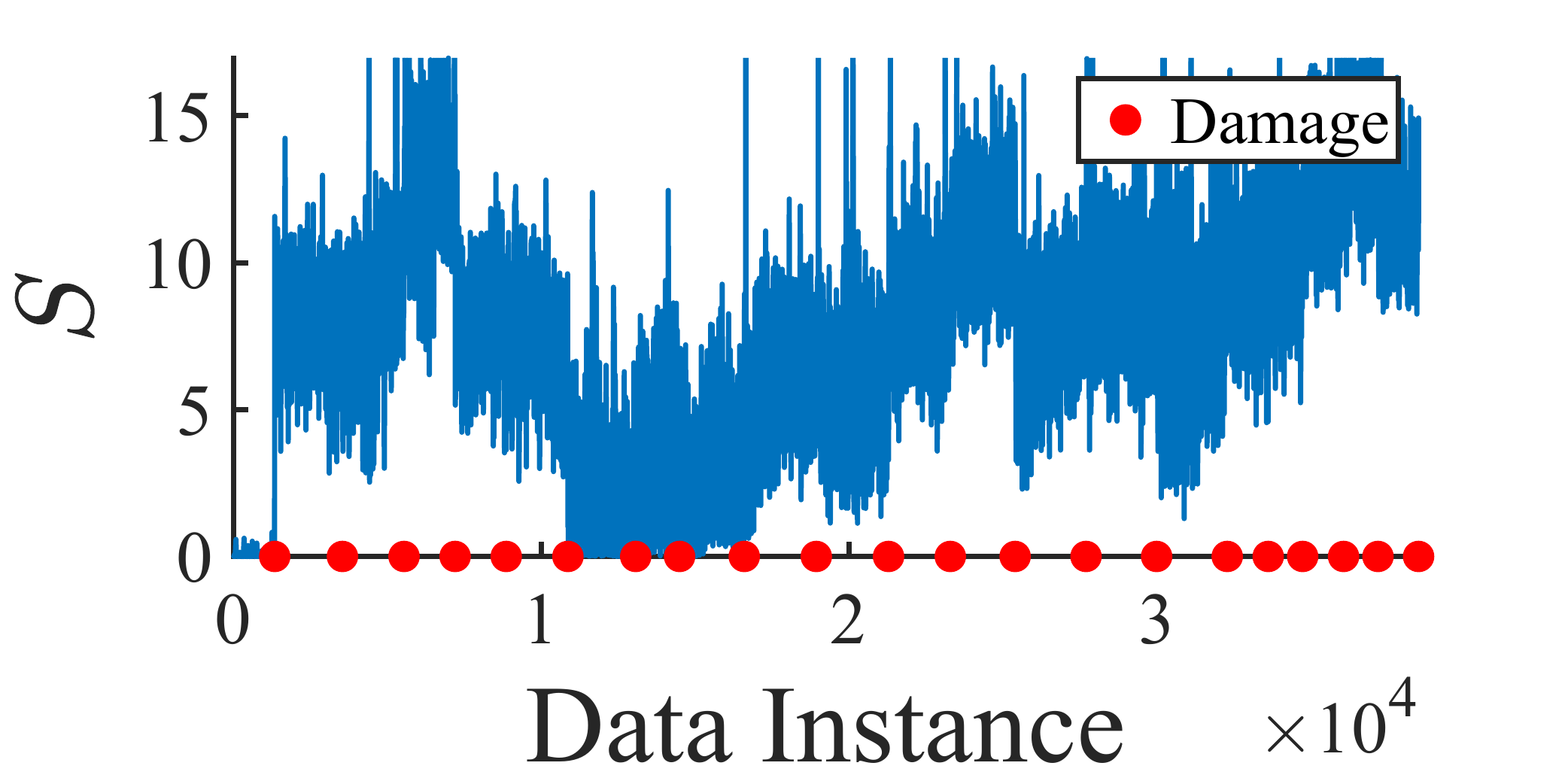}
        \caption {YS6\_500 performance on Yellow Frame (no TL).}
        \label{Fig:Yellow_Self_3}  
\end{subfigure}
\begin{subfigure}{0.48\linewidth} 
\centering
  \includegraphics[width=1\linewidth]{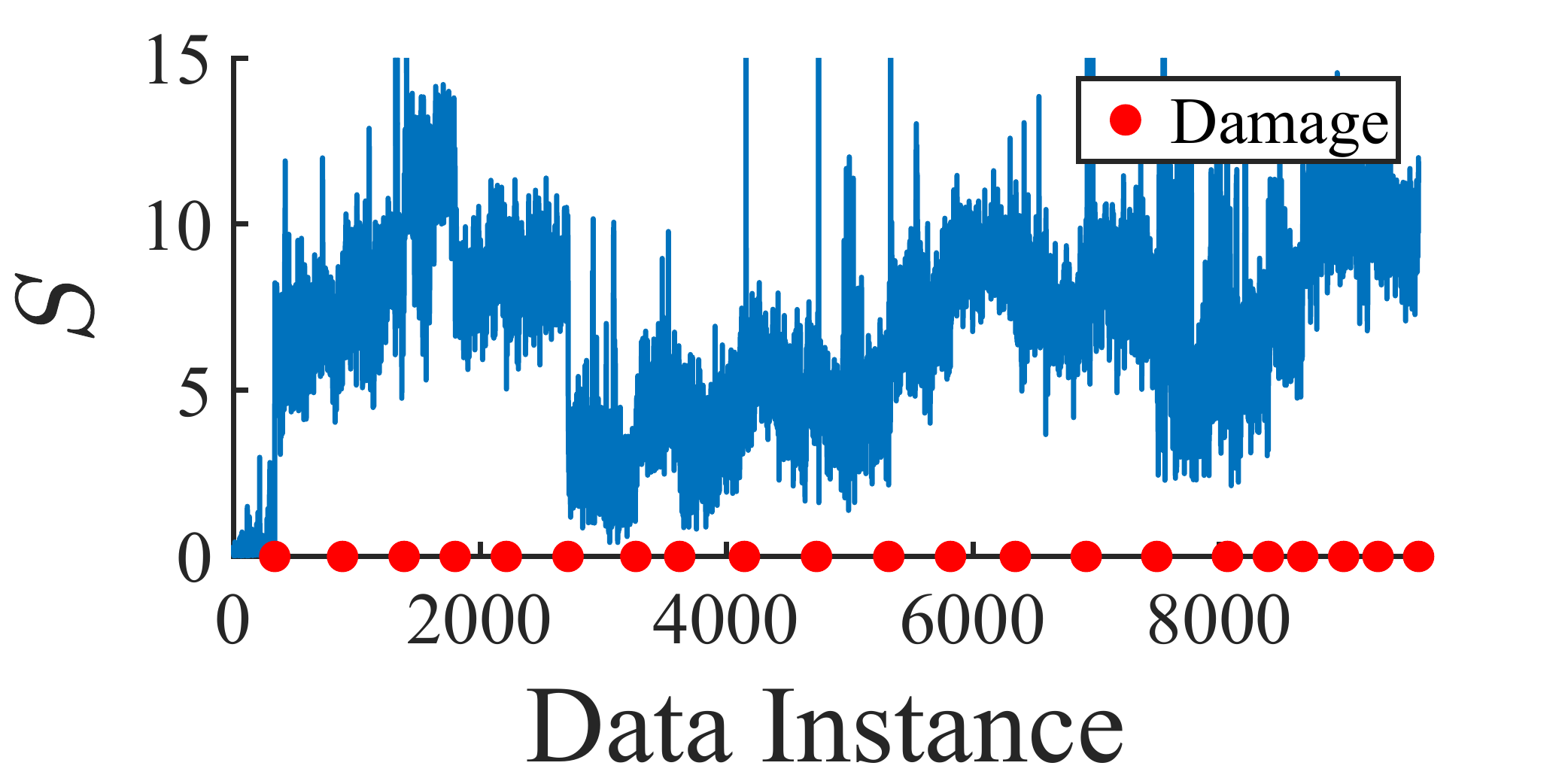}
        \caption {Q2YS10\_1000 performance on Yellow Frame with all no-damage case data given for the DA.}
        \label{Fig:Yellow_Self_4}  
\end{subfigure}

\caption{Impact of $W$ and $N$ on source-no-damage identifiers; Yellow Frame.}
\label{Fig:Yellow_Self}
\end{figure}

\begin{figure}[h!]
\centering
\begin{subfigure}{1\linewidth} 
\centering
  \includegraphics[width=1\linewidth]{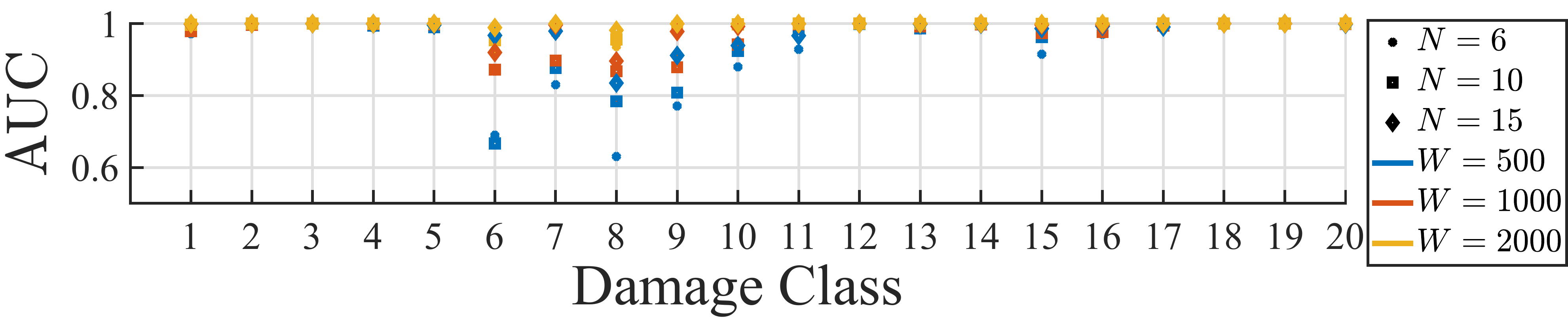}
        \caption {Source QUGS.}
        \label{Fig:Y_R_1}  
\end{subfigure}
\begin{subfigure}{1\linewidth} 
\centering
  \includegraphics[width=1\linewidth]{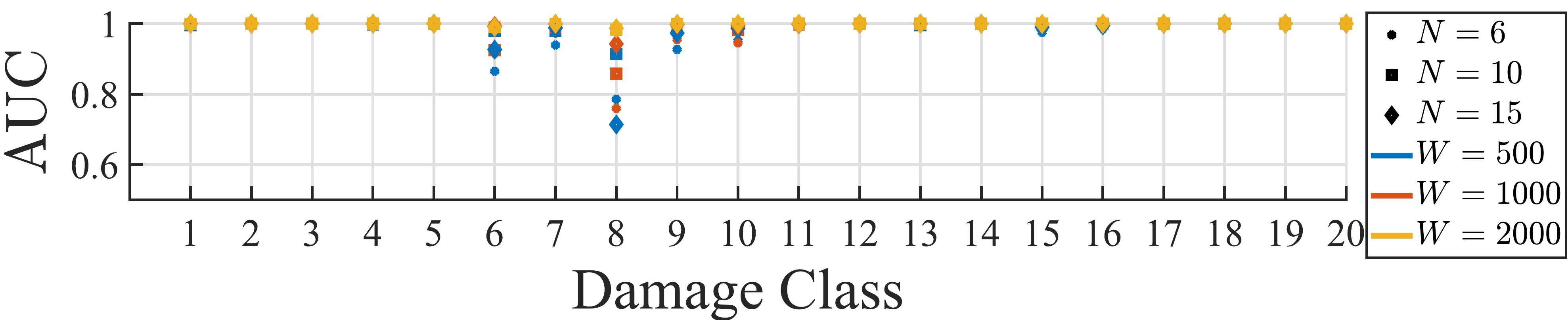}
        \caption {Source Z24.}
        \label{Fig:Y_R_2}  
\end{subfigure}
\caption{TL performance; AUC values for all Yellow Frame TL scenarios.}
\label{Fig:Y_R_A}
\end{figure}

Regarding the impact of $W$ on the TL's performance, a lower $W$ tends to lower the AUC, which is expected. With a lower $W$, the spectral resolution of both FFTs and power spectrums drops. For instance, impacts of lowering $N$ and $W$ are shown for several $W$ and $N$ cases in Fig.~\ref{Fig:Yellow_Self} for the Yellow Frame $\mathcal{D}$ model tested on itself (\ie no TL). A lower $W$ in a constant $N$ deteriorates the detection scores, increasing false alarms and reducing AUC in return. DC8 and DC6 have the closest $S$ (Fig.~\ref{Fig:Yellow_Self}) and the closest fundamental frequency to the no-damage case (Table~\ref{Table:Yellow_Frame_Data}) and are the cases that undergo the most AUC reduction with lower $W$ and $N$ (Fig.~\ref{Fig:Yellow_Self_1} to Fig.~\ref{Fig:Yellow_Self_3}). Yet, given the whole spectrum of AUC values for all TL cases of Yellow Frame (Figs.~\ref{Fig:Y_R_A}), and with full sensors, the TL method thus shows acceptable robustness to $W$. TL-based anomaly detection scores for the Yellow Frame dataset are shown in Fig.~\ref{Fig:Yellow_Self_4}. It depicts how well DA transferred the differentiation power of the QUGS detector between no-damage and damage cases data to the target Yellow Frame, with DA only applied over 10\% of its no-damage data.

\subsection{Target: The QUGS}
\begin{figure}[h!]
\centering
\begin{subfigure}{0.49\linewidth} 
\centering
  \includegraphics[width=0.7\linewidth]{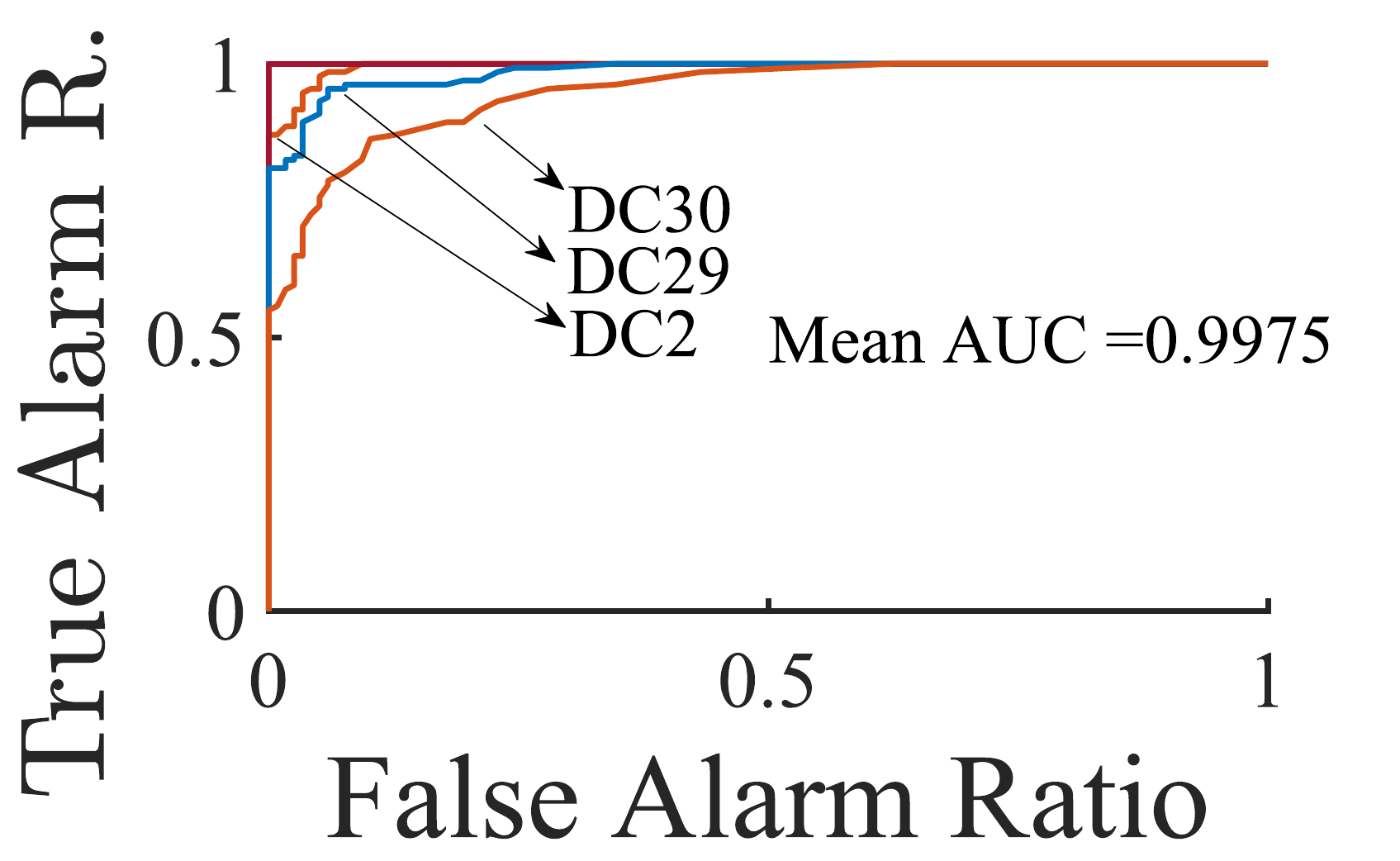}
        \caption {Y2QS15\_2000; ROC of 30 damages.}
        \label{Fig:Q_ROC_1}  
\end{subfigure}
\begin{subfigure}{0.49\linewidth} 
\centering
  \includegraphics[width=0.7\linewidth]{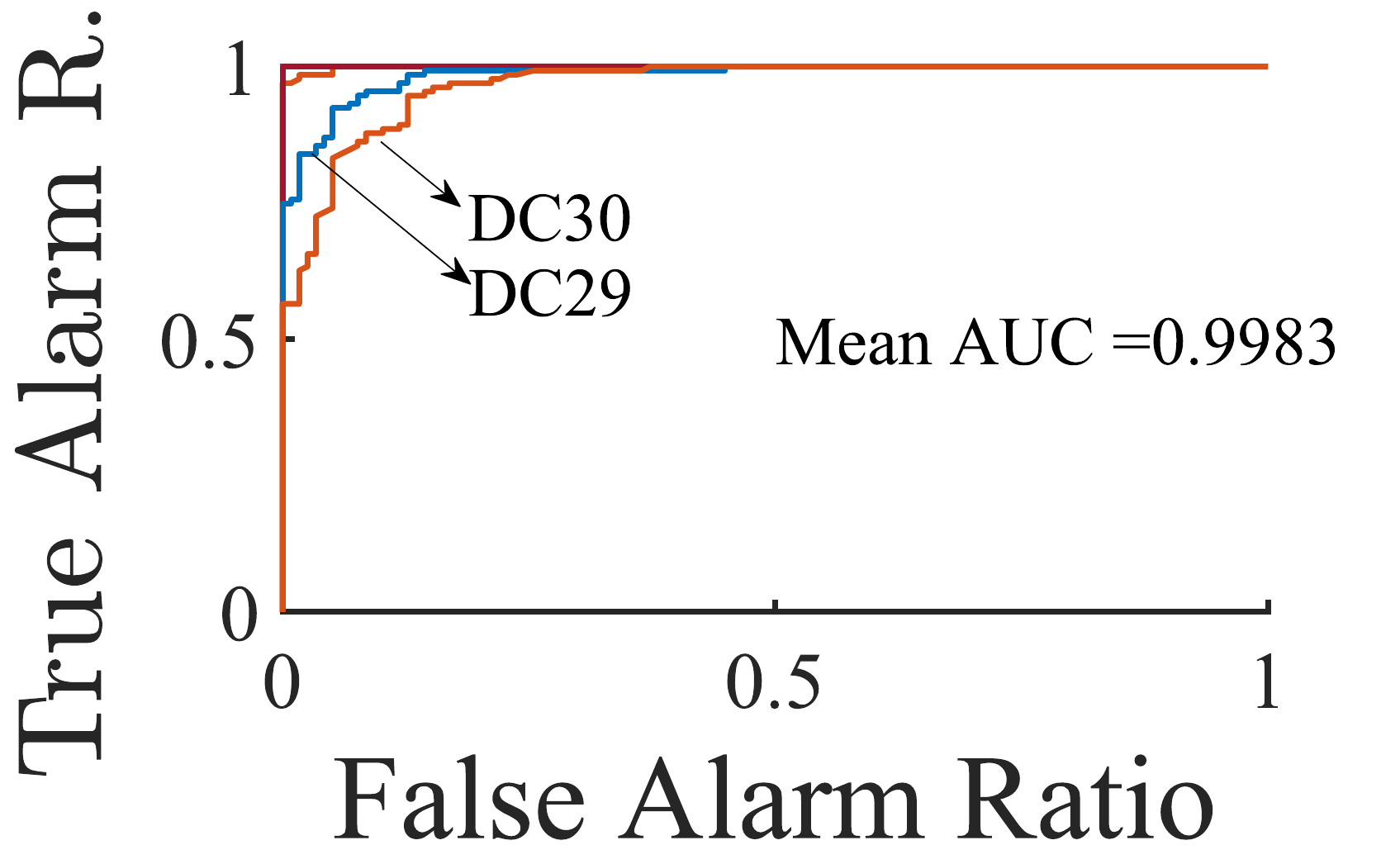}
        \caption {Z2QS6\_2000; ROC of 20 damages.}
        \label{Fig:Q_ROC_2}  
\end{subfigure}

\begin{subfigure}{0.49\linewidth} 
\centering
  \includegraphics[width=0.7\linewidth]{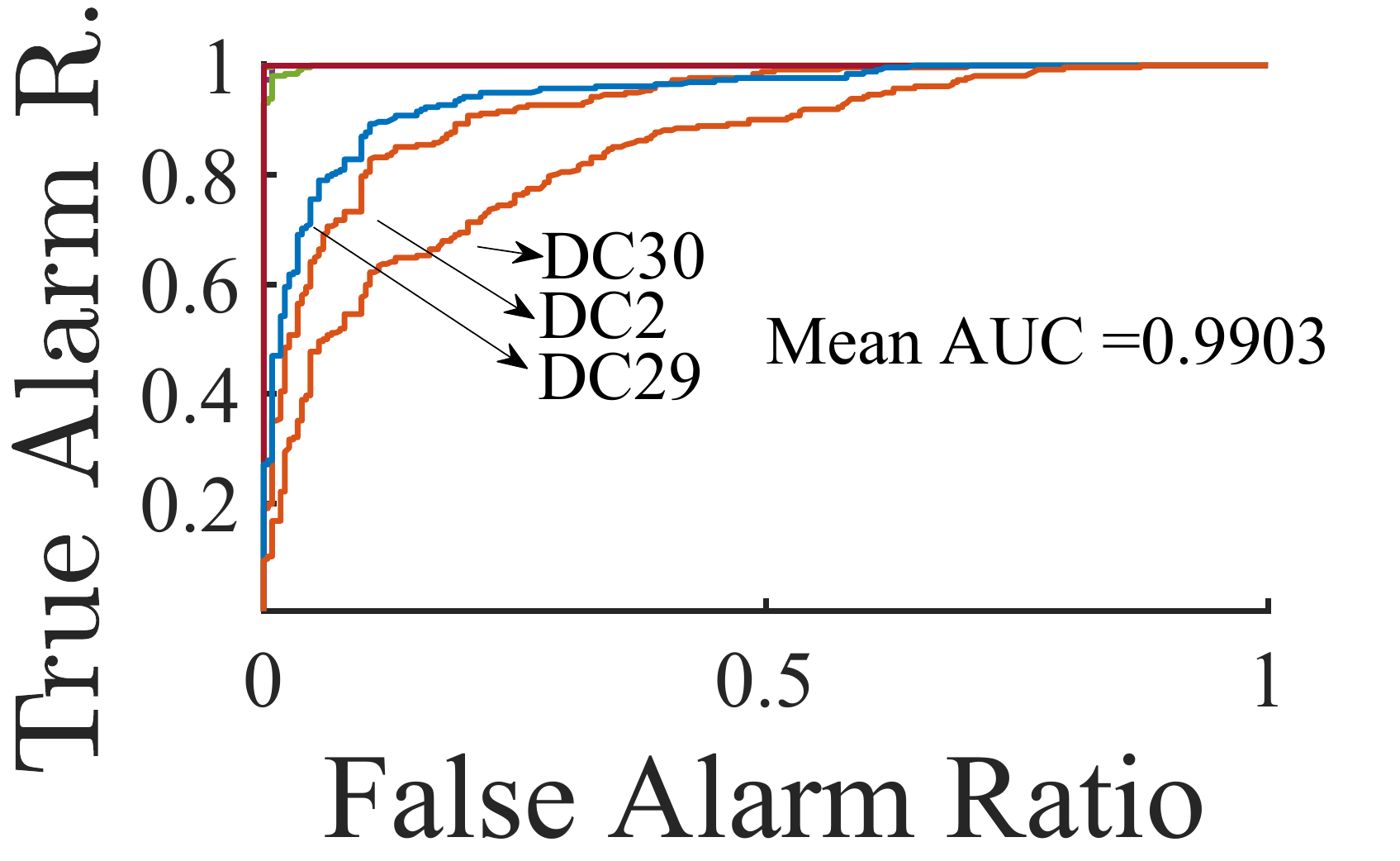}
        \caption {Z2QS10\_1000; ROC of 20 damages.}
        \label{Fig:Q_ROC_3}  
\end{subfigure}
\begin{subfigure}{0.49\linewidth} 
\centering
  \includegraphics[width=0.7\linewidth]{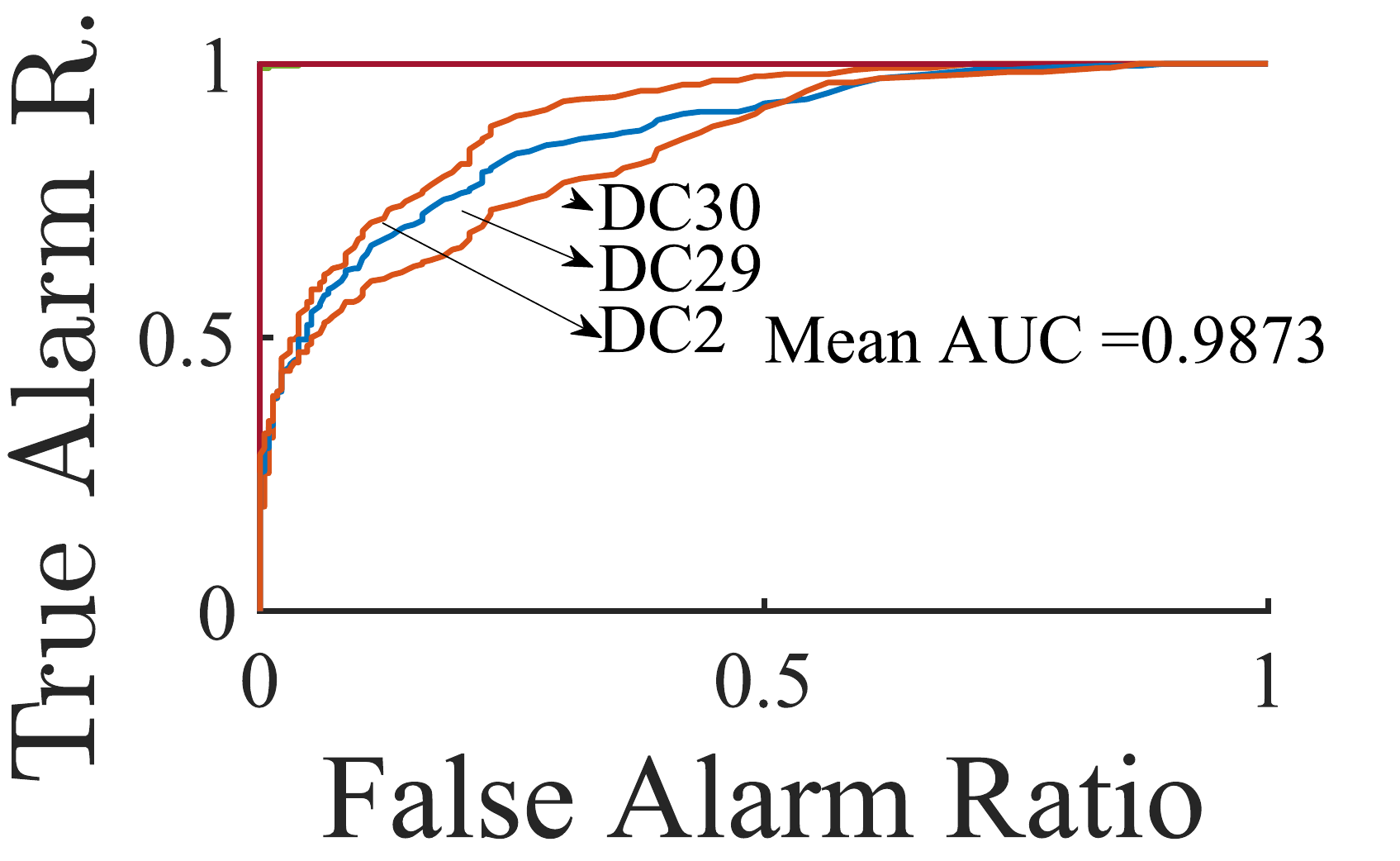}
        \caption {Y2QS6\_1000; ROC of 20 damages.}
        \label{Fig:Q_ROC_4}  
\end{subfigure}
\begin{subfigure}{0.49\linewidth} 
\centering
  \includegraphics[width=0.7\linewidth]{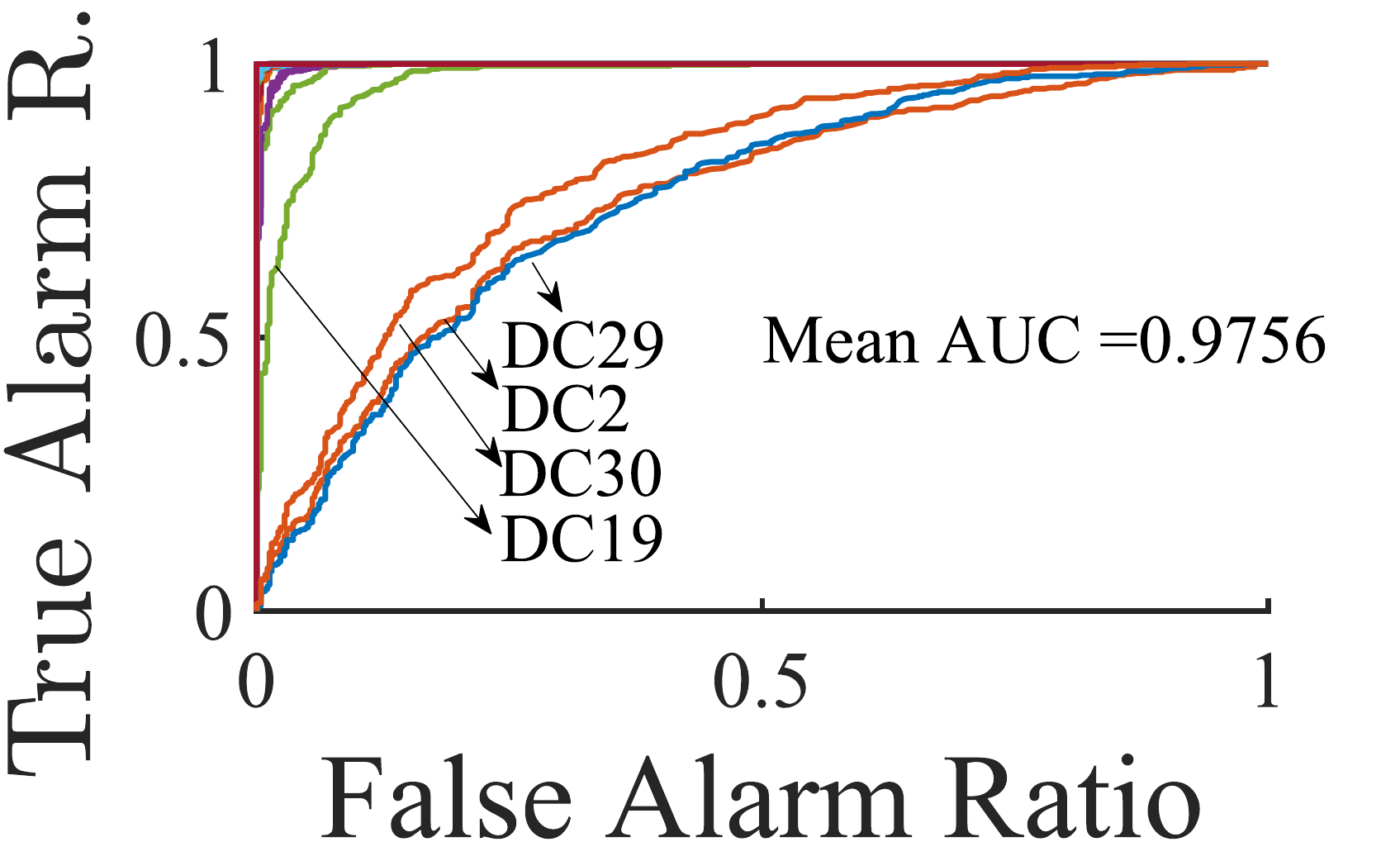}
        \caption {Y2QS15\_500; ROC of 20 damages.}
        \label{Fig:Q_ROC_5}  
\end{subfigure}
\begin{subfigure}{0.49\linewidth} 
\centering
  \includegraphics[width=0.7\linewidth]{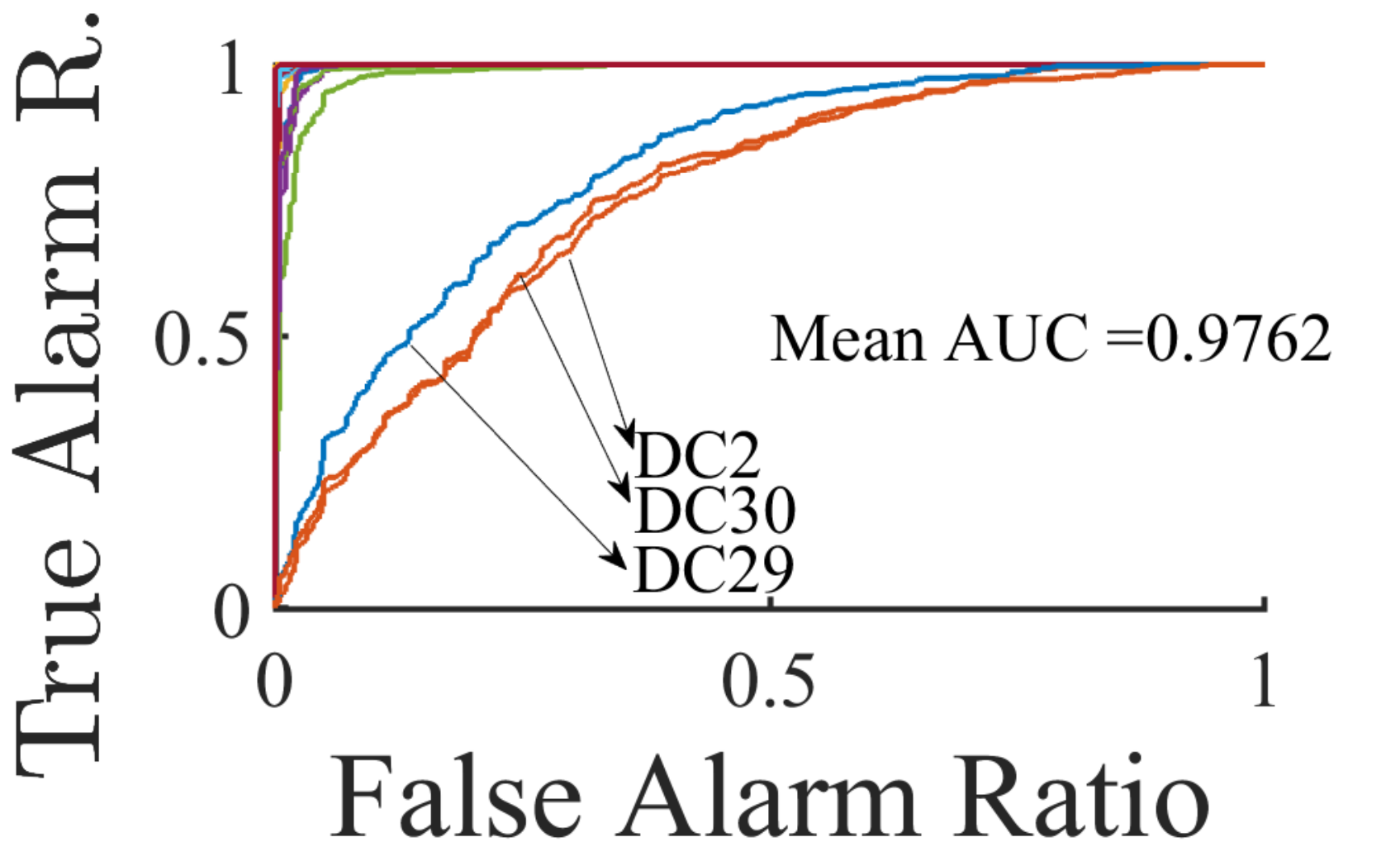}
        \caption {Z2QS6\_500; ROC of 20 damages.}
        \label{Fig:Q_ROC_6}  
\end{subfigure}
\caption{TL performance; QUGS as the target domain.}
\label{Fig:Q_ROC_A}
\end{figure}

\begin{figure}[h!]
\centering
\begin{subfigure}{0.48\linewidth} 
\centering
  \includegraphics[width=1\linewidth]{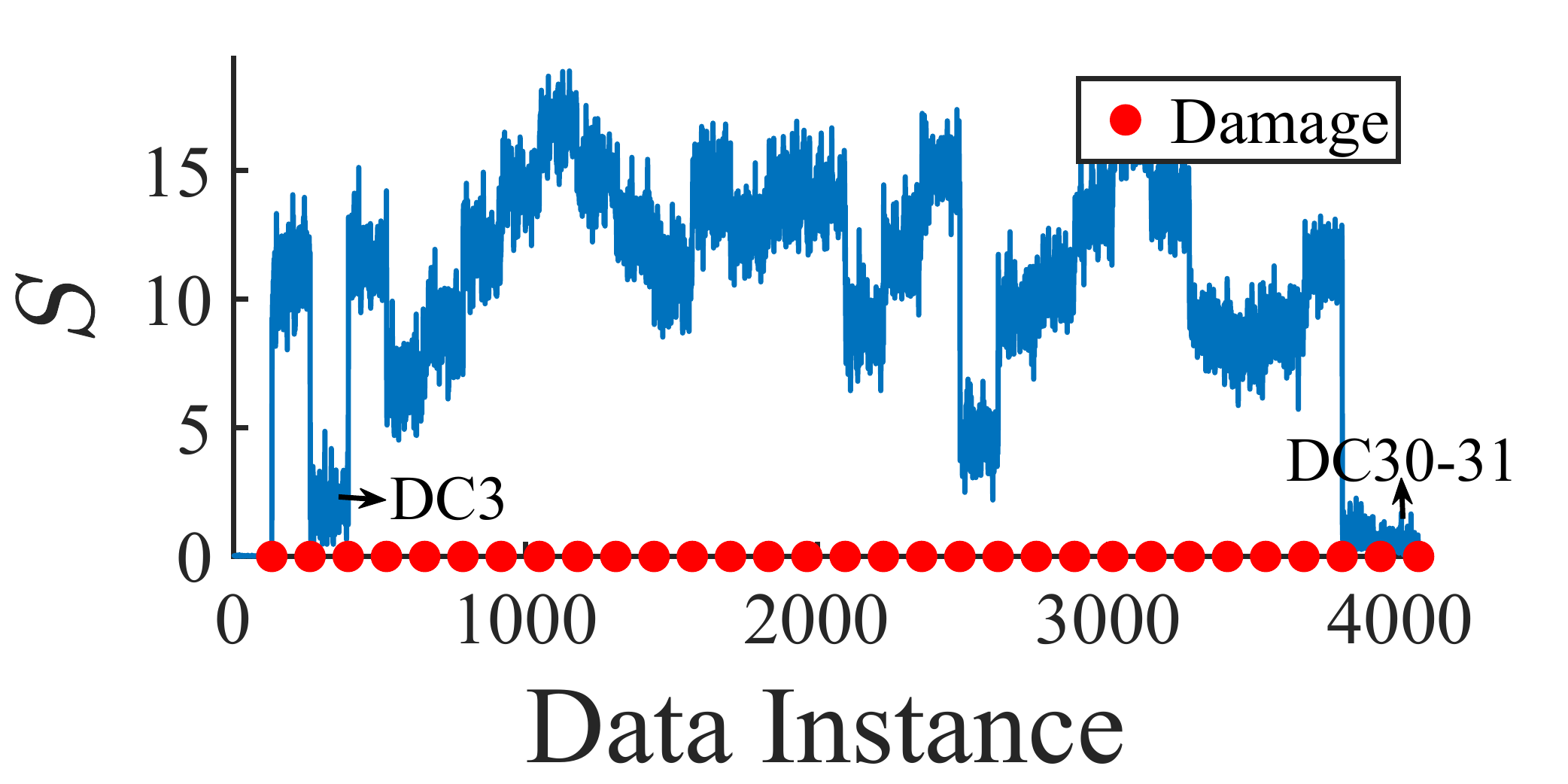}
        \caption {QS15\_2000 performance on QUGS  (no TL).\\\phantom{}}
        \label{Fig:QUGS_Self_1}  
\end{subfigure}
\begin{subfigure}{0.48\linewidth} 
\centering
  \includegraphics[width=1\linewidth]{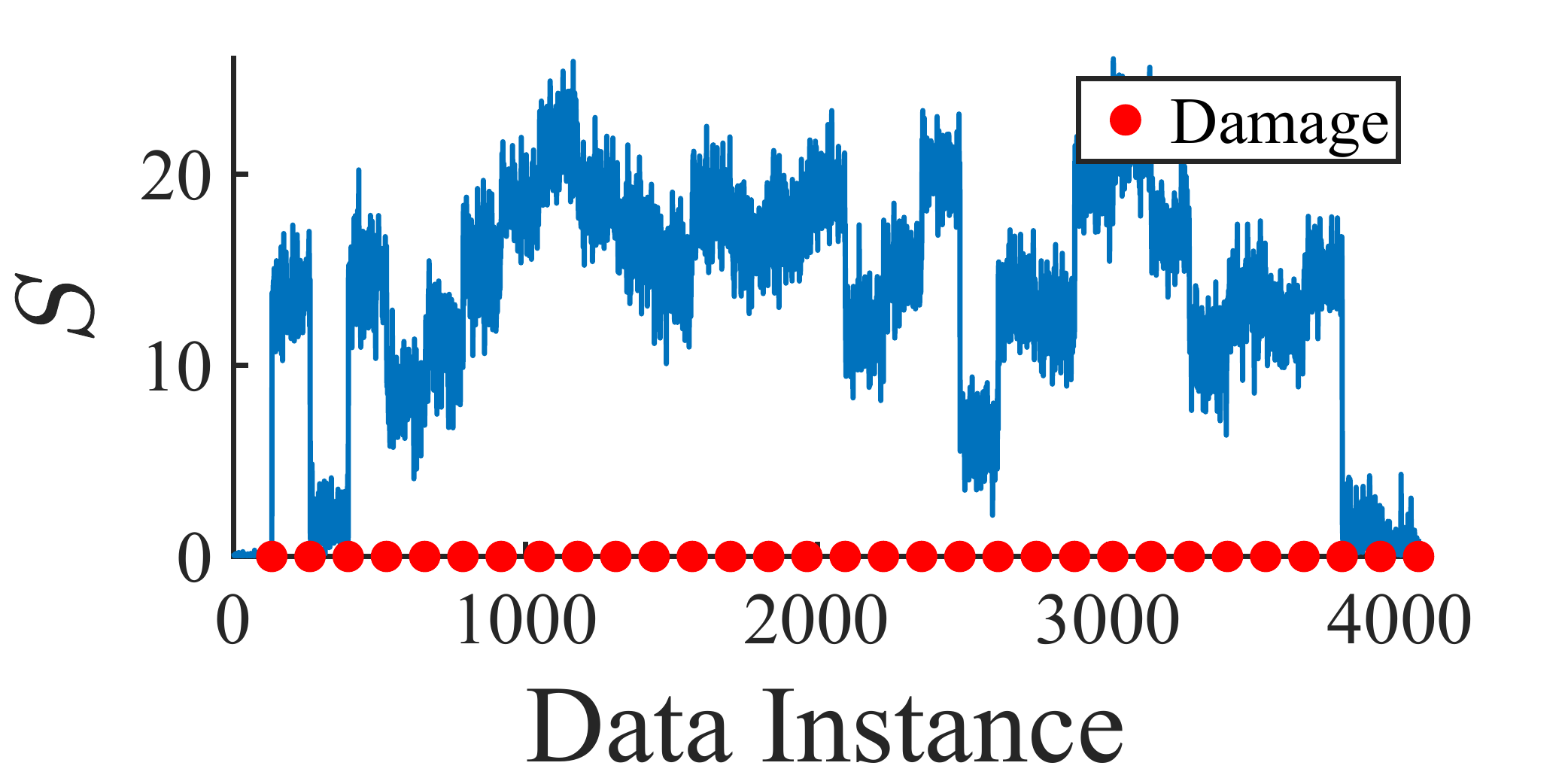}
        \caption {Y2QS15\_1000 performance on QUGS, given the whole no-damage data as DA input.}
        \label{Fig:QUGS_Self_2}  
\end{subfigure}
\caption{Impact of $W$ and $N$ on source-no-damage identifiers; QUGS target.}
\label{QUGS_Self}
\end{figure}

\begin{figure}[h!]
\centering
\begin{subfigure}{1\linewidth} 
\centering
  \includegraphics[width=1\linewidth]{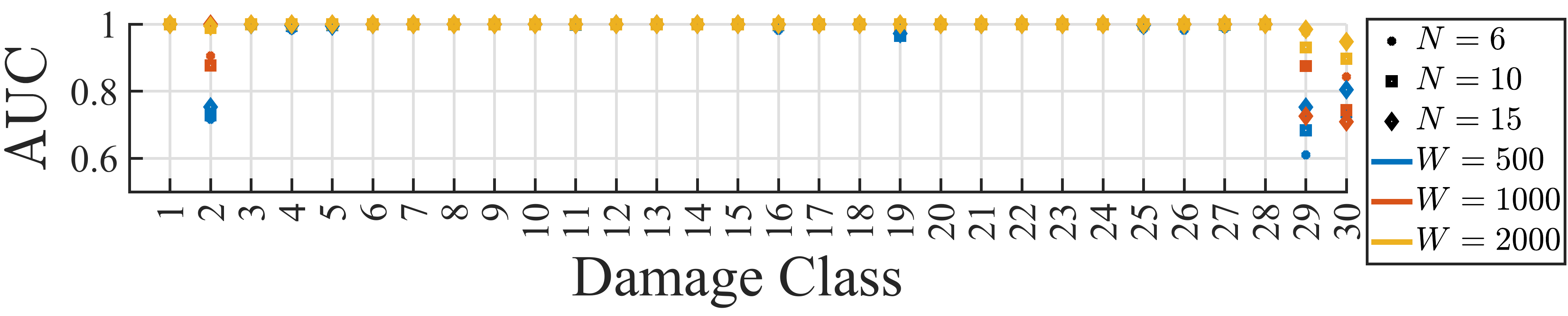}
        \caption {Source Yellow Frame.}
        \label{Fig:Q_R_1}  
\end{subfigure}
\begin{subfigure}{1\linewidth} 
\centering
  \includegraphics[width=1\linewidth]{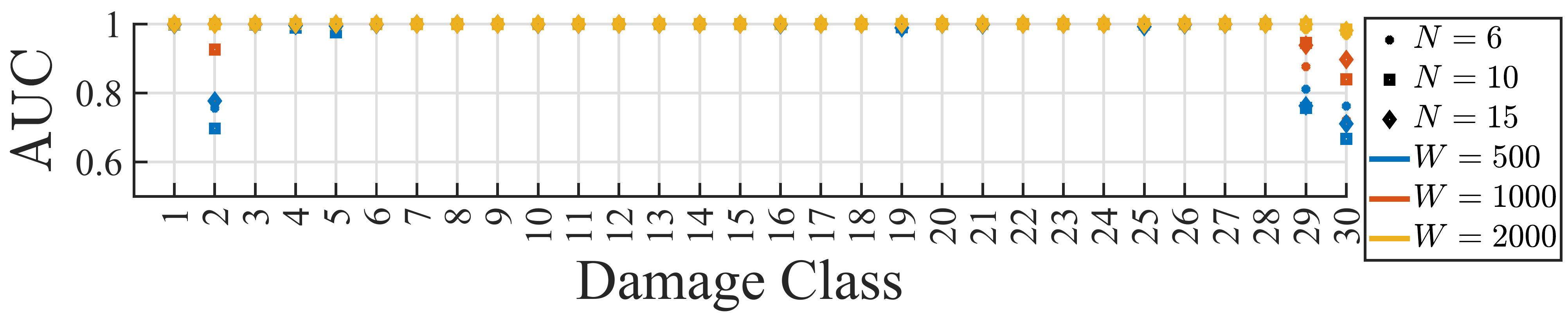}
        \caption {Source Z24.}
        \label{Fig:Q_R_2}  
\end{subfigure}
\caption{TL performance; AUC values for all QUGS TL scenarios.}
\label{Fig:Q_R_A}
\end{figure}

\noindent
TL for the target structure QUGS with 30 damage cases is performed with the source structures Yellow Frame and Z24. ROC curves for 6 of the TL scenarios in Table~\ref{Table:TL_Scenarios} are shown in Fig.~\ref{Fig:Q_ROC_A}. As with the Yellow Frame TL, $W=2000$ and $N=15$ provide a seamless TL performance, with a mean AUC of 0.9975, in which DC29, DC30, and DC2 show the least performance. Further lowering $W$ and $N$, the AUC scores of those cases are reduced to 0.7. Since Yellow Frame has 15 sensors, only the odd sensors (15 sensors) are used in TL for the QUGS (Table~\ref{Table:TL_Scenarios} and Fig.~\ref{Fig:QUGS_Structure}). Thus, no sensors are present near damages DC2 and DC30, a probable reason for the lower AUCs. As with the Yellow Frame, DC3, DC31, and DC30 have patterns similar to the no-damage cases. Lowering $N$ and $W$ values makes them less distinctive from the no-damage case. Inspecting the differentiation performance between QUGS damage and no-damage cases without DA, DC2, DC29, and DC30 have the most similar detection scores with the no-damage case, as shown in Fig.~\ref{Fig:QUGS_Self_1} ($W=20000$ and $N=15$). Further reduction on $W$ and $N$ reduces their differences, and the AUC values eventually drop even more. The same trend can be seen on the TL of QUGS with Yellow Frame $W=2000$ and $N=15$ (Fig.~\ref{Fig:QUGS_Self_2}). It still offers good discrimination between no-damage and all damage cases, including DC2 and DC30. The spectra of all experiments' AUC values are shown in Figs.~\ref{Fig:Q_R_1} and~\ref{Fig:Q_R_1}, with the source Yellow Frame and Z24, respectively. With $N=15$, which are half of QUGS sensors, and with different $W$s, 27 damage cases are seamlessly differentiated from the no-damage case with AUC scores close to 1. Yet, for the cases DC2, DC29, and DC30, lower AUC scores are observed with lower $W$ and $N$. Overall, for the QUGS dataset, the proposed DA method shows satisfactory robustness to $W$, specifically if more sensors are used. 

\subsection{Target : The Z24}
\label{sec:Results_Z24}
\begin{figure}[h!]
\centering
\begin{subfigure}{0.49\linewidth} 
\centering
  \includegraphics[width=0.7\linewidth]{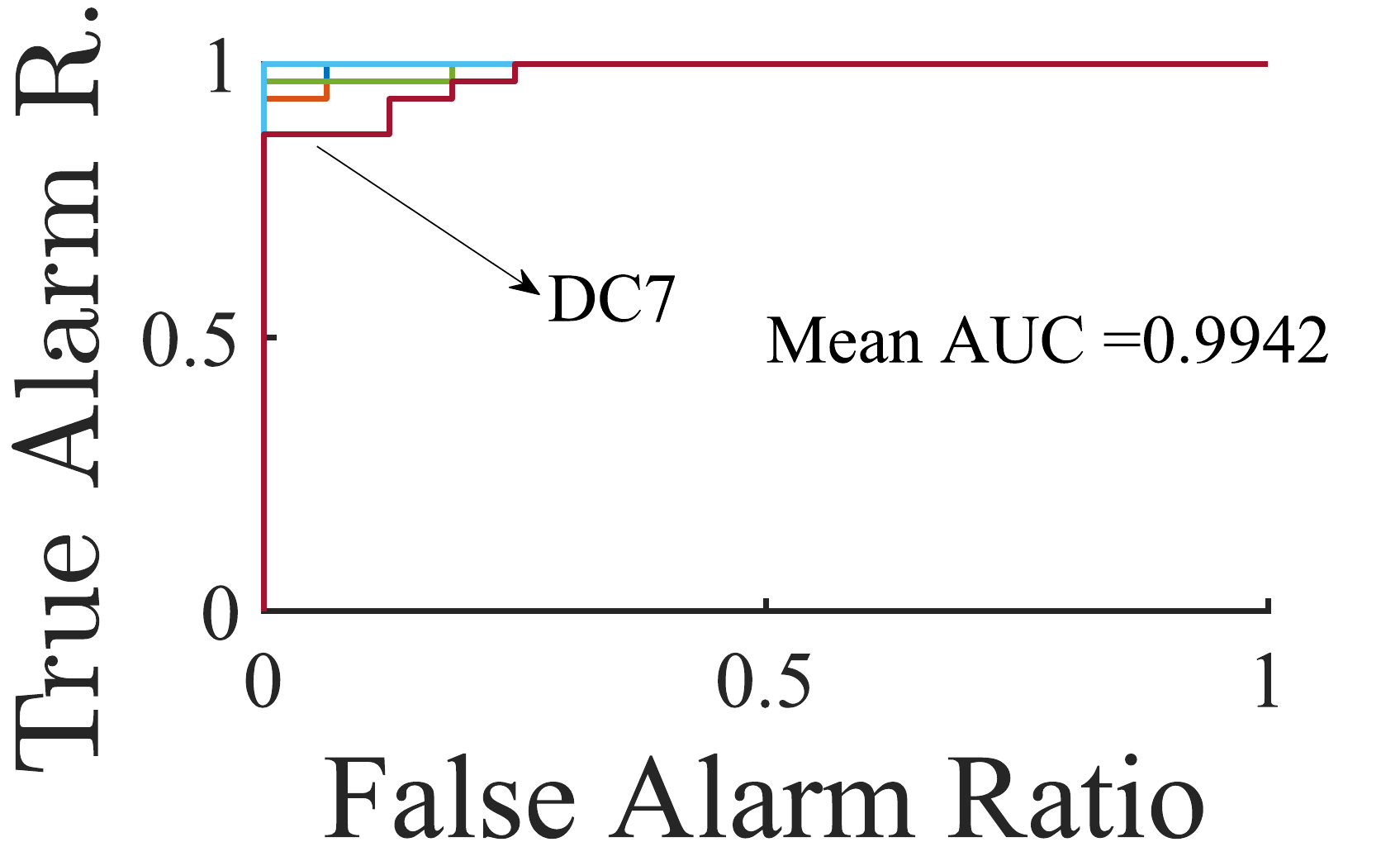}
        \caption {Y2ZS15\_2000; ROC of seven damages.}
        \label{Fig:Z_ROC_1}  
\end{subfigure}
\begin{subfigure}{0.49\linewidth} 
\centering
  \includegraphics[width=0.7\linewidth]{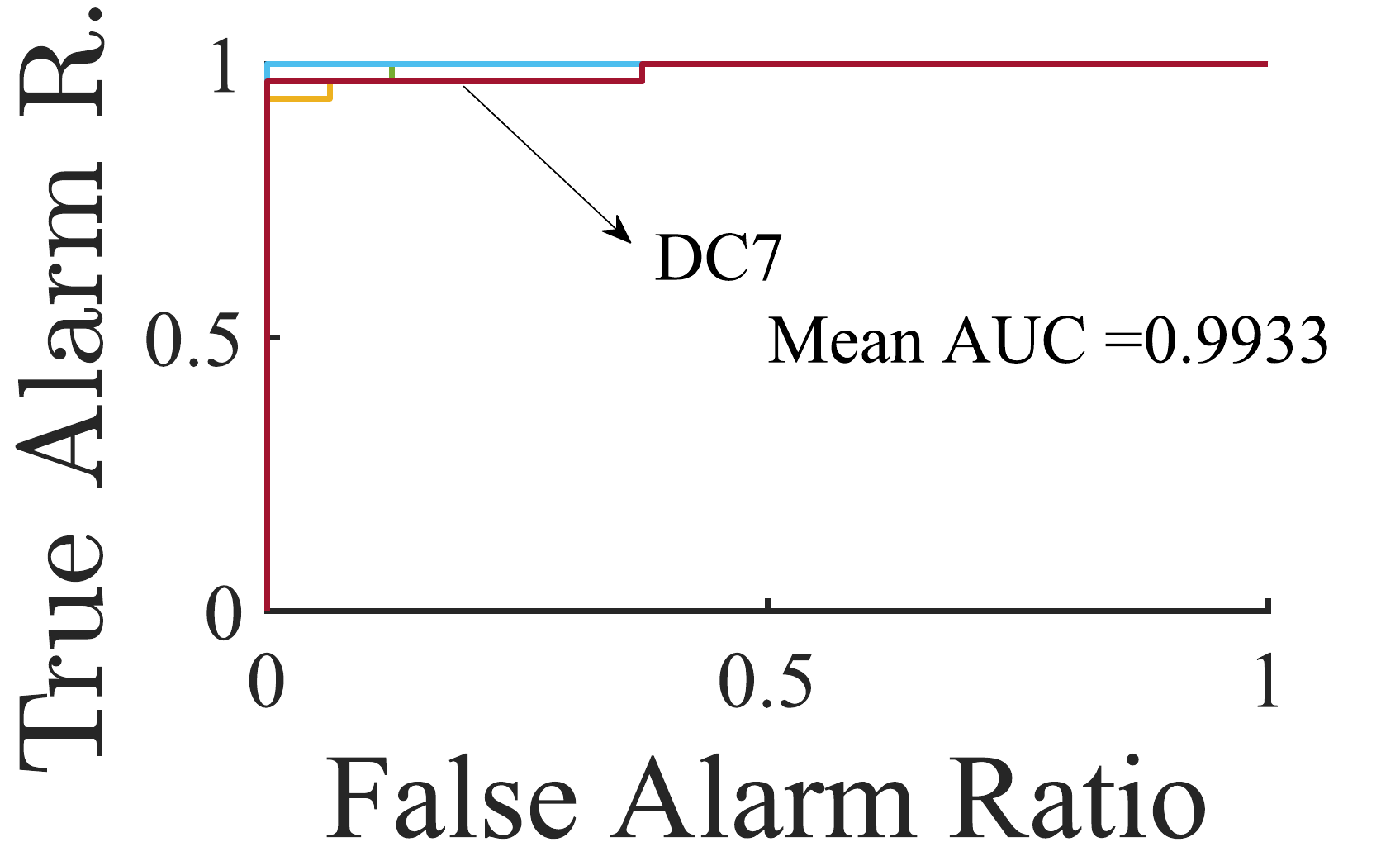}
        \caption {Q2ZS6\_2000; ROC of seven damages.}
        \label{Fig:Z_ROC_2}  
\end{subfigure}

\begin{subfigure}{0.49\linewidth} 
\centering
  \includegraphics[width=0.7\linewidth]{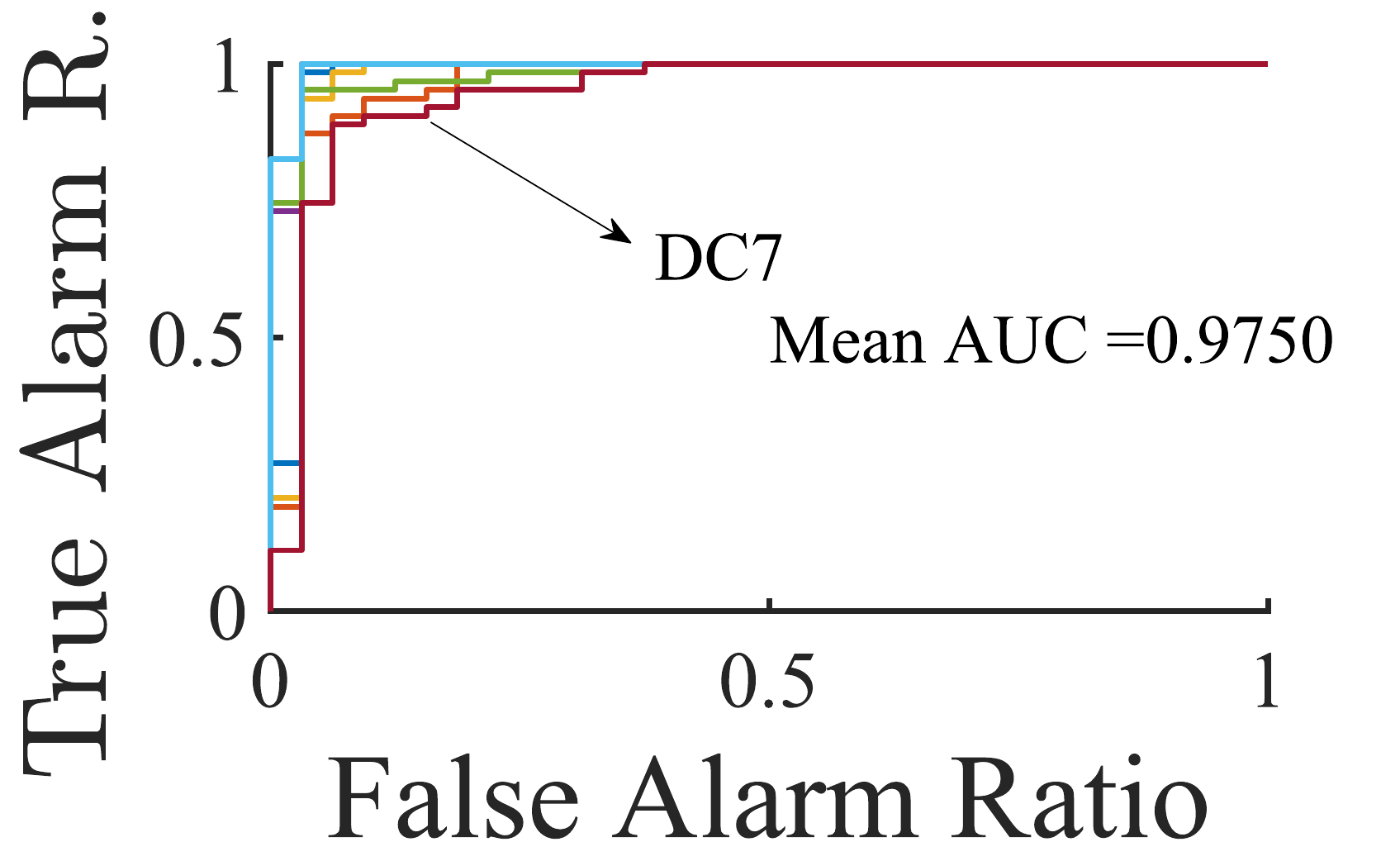}
        \caption {Q2ZS10\_1000; ROC of seven damages.}
        \label{Fig:Z_ROC_3}  
\end{subfigure}
\begin{subfigure}{0.49\linewidth} 
\centering
  \includegraphics[width=0.7\linewidth]{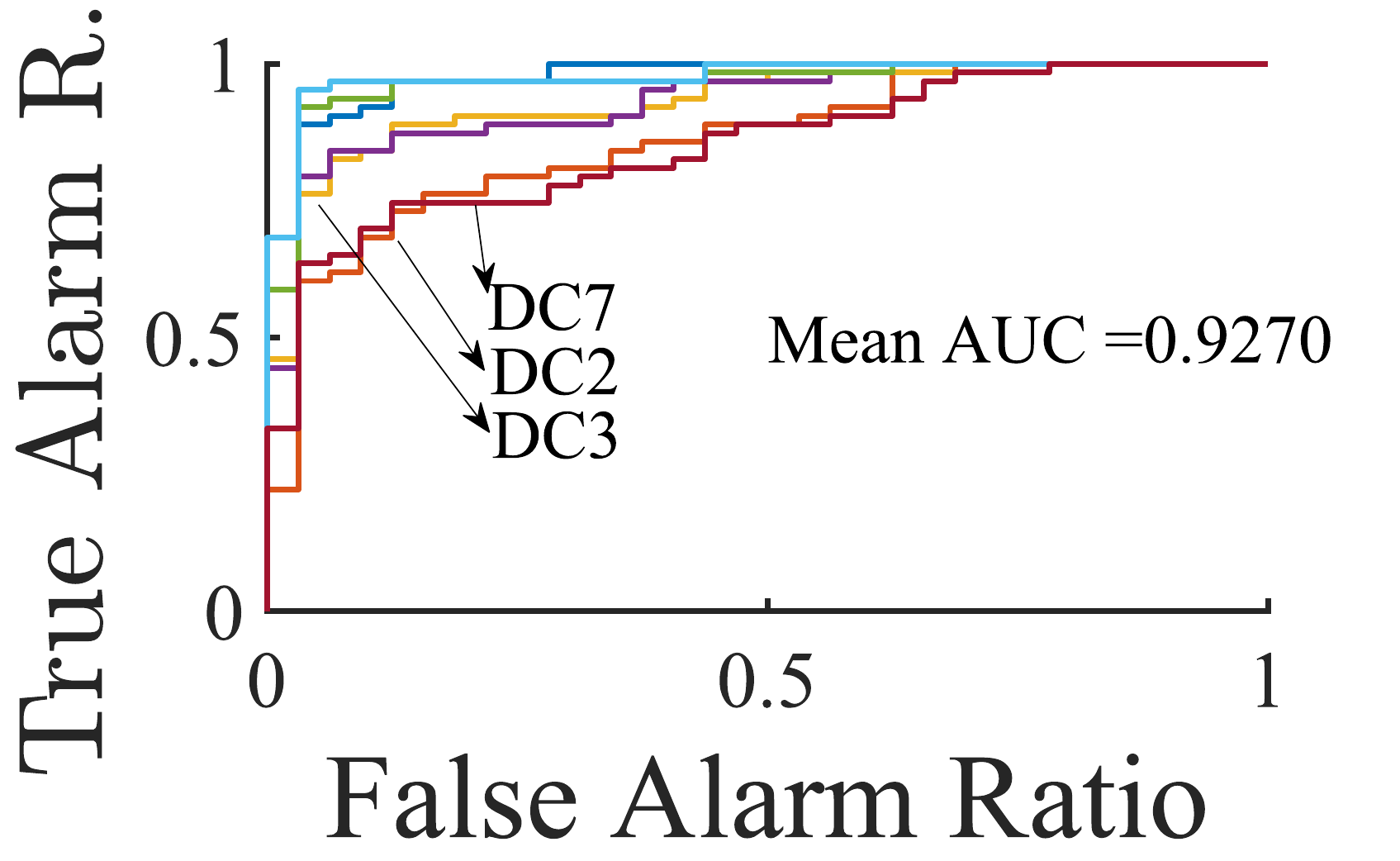}
        \caption {Y2ZS6\_1000; ROC of seven damages.}
        \label{Fig:Z_ROC_4}  
\end{subfigure}
\begin{subfigure}{0.49\linewidth} 
\centering
  \includegraphics[width=0.7\linewidth]{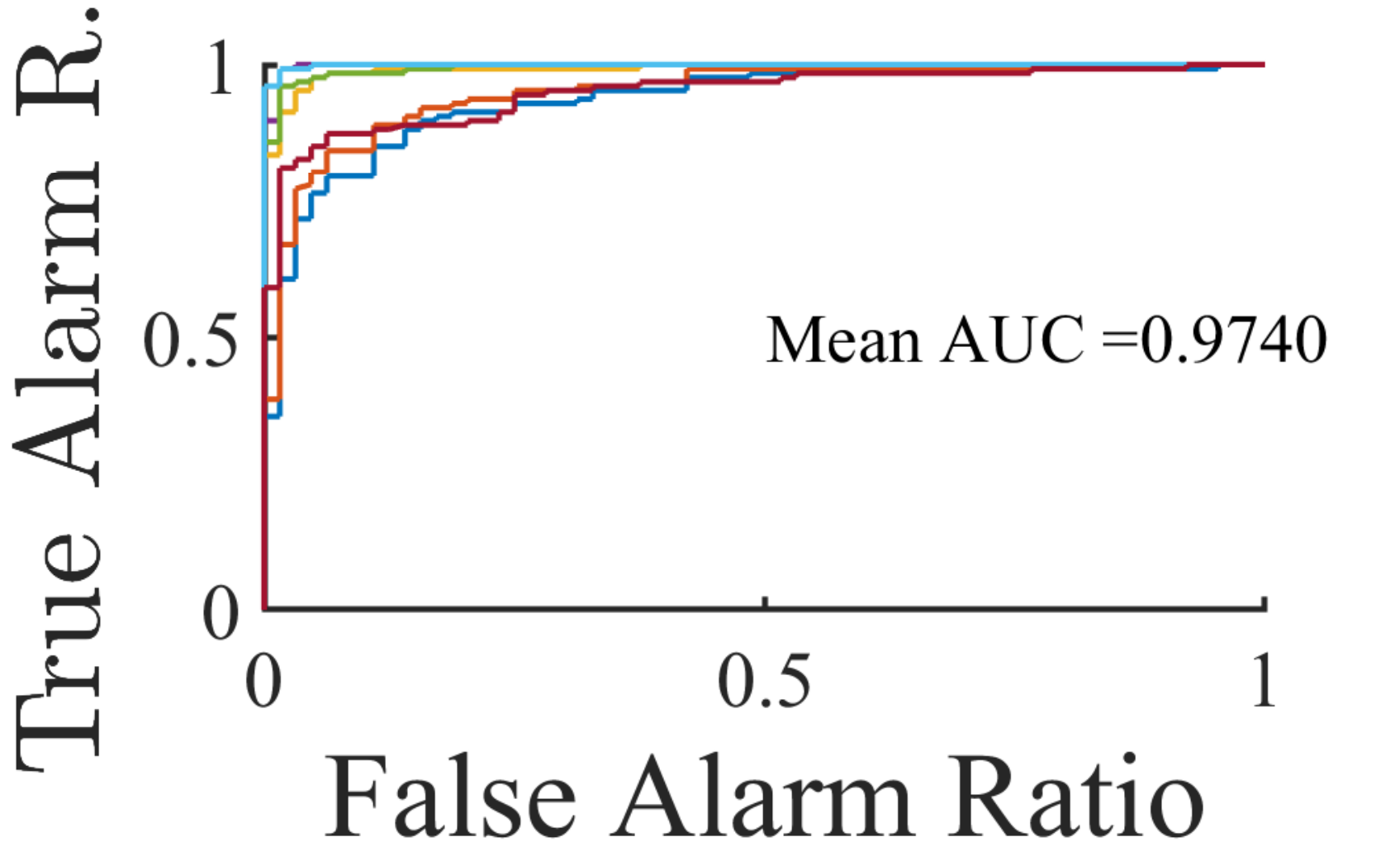}
        \caption {Y2ZS15\_500; ROC of seven damages.}
        \label{Fig:Z_ROC_5}  
\end{subfigure}
\begin{subfigure}{0.49\linewidth} 
\centering
  \includegraphics[width=0.7\linewidth]{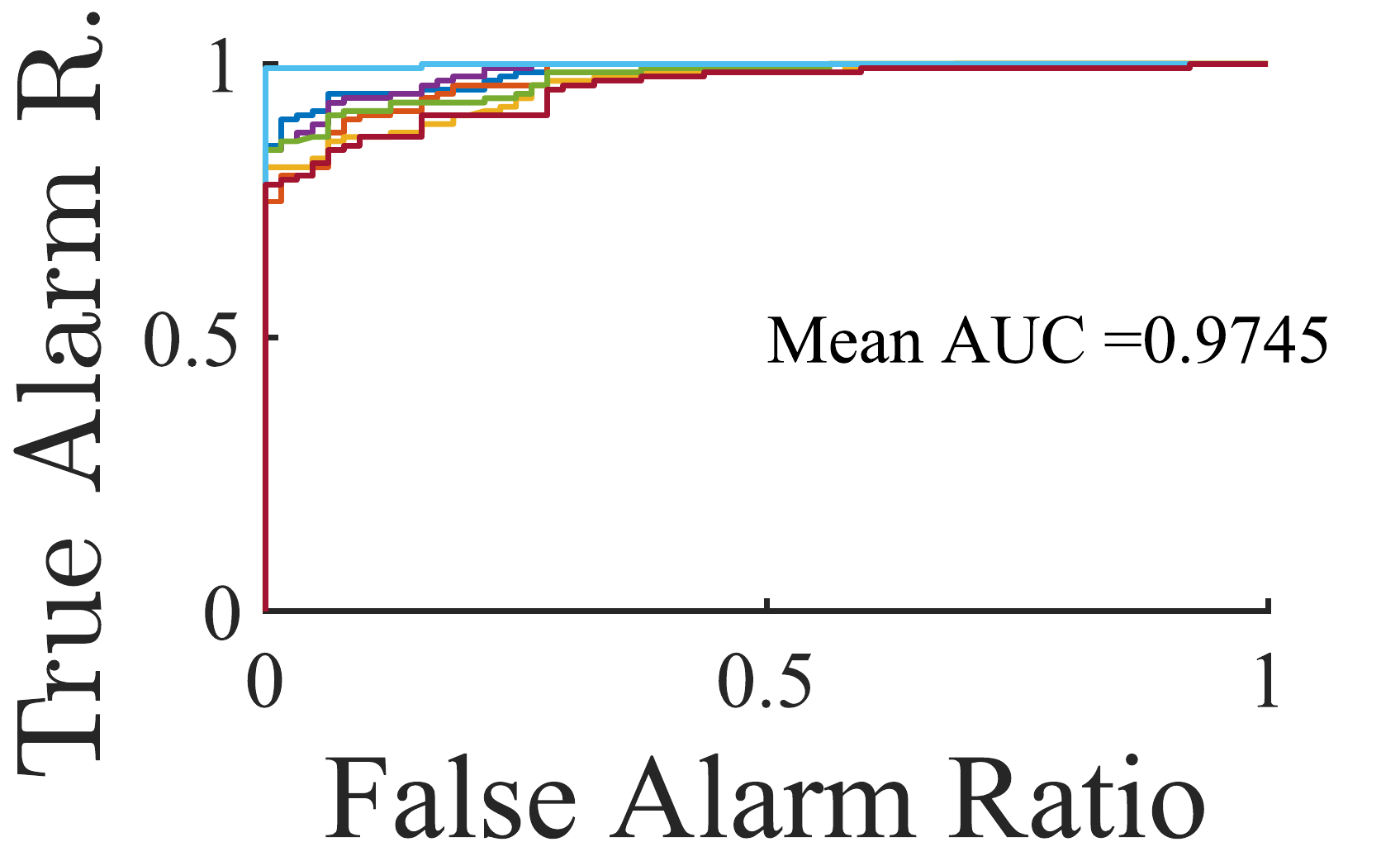}
        \caption {Q2ZS6\_500; ROC of seven damages.}
        \label{Fig:Z_ROC_6}  
\end{subfigure}
\caption{TL performance; Z24 as the target domain.}
\label{Fig:Z_ROC_A}
\end{figure}

\begin{figure}[h!]
\centering
\begin{subfigure}{0.49\linewidth} 
\centering
  \includegraphics[width=1\linewidth]{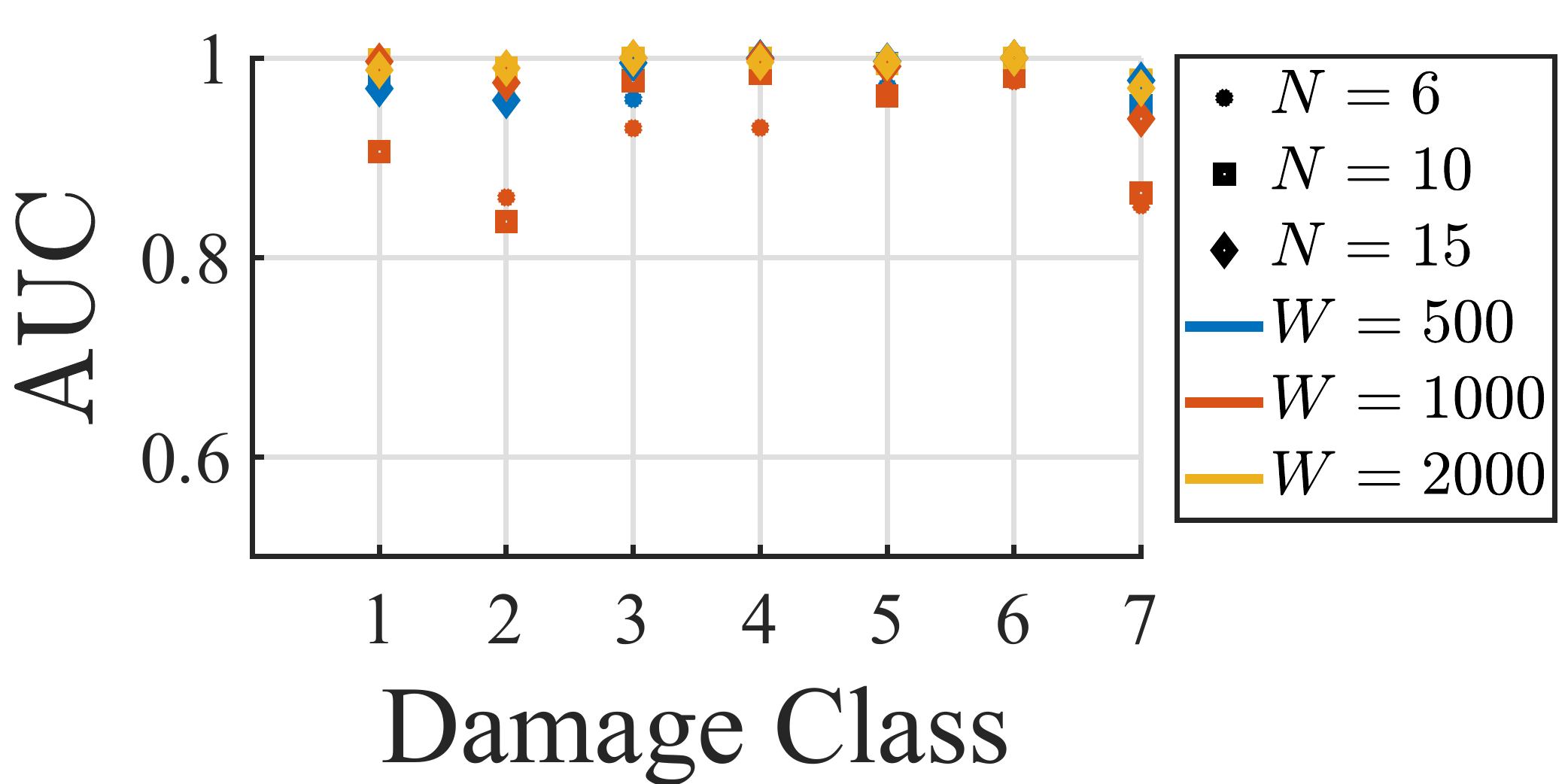}
        \caption {Source Yellow Frame.}
        \label{Fig:Z_R_1}  
\end{subfigure} 
\begin{subfigure}{0.47\linewidth} 
\centering
  \includegraphics[width=1\linewidth]{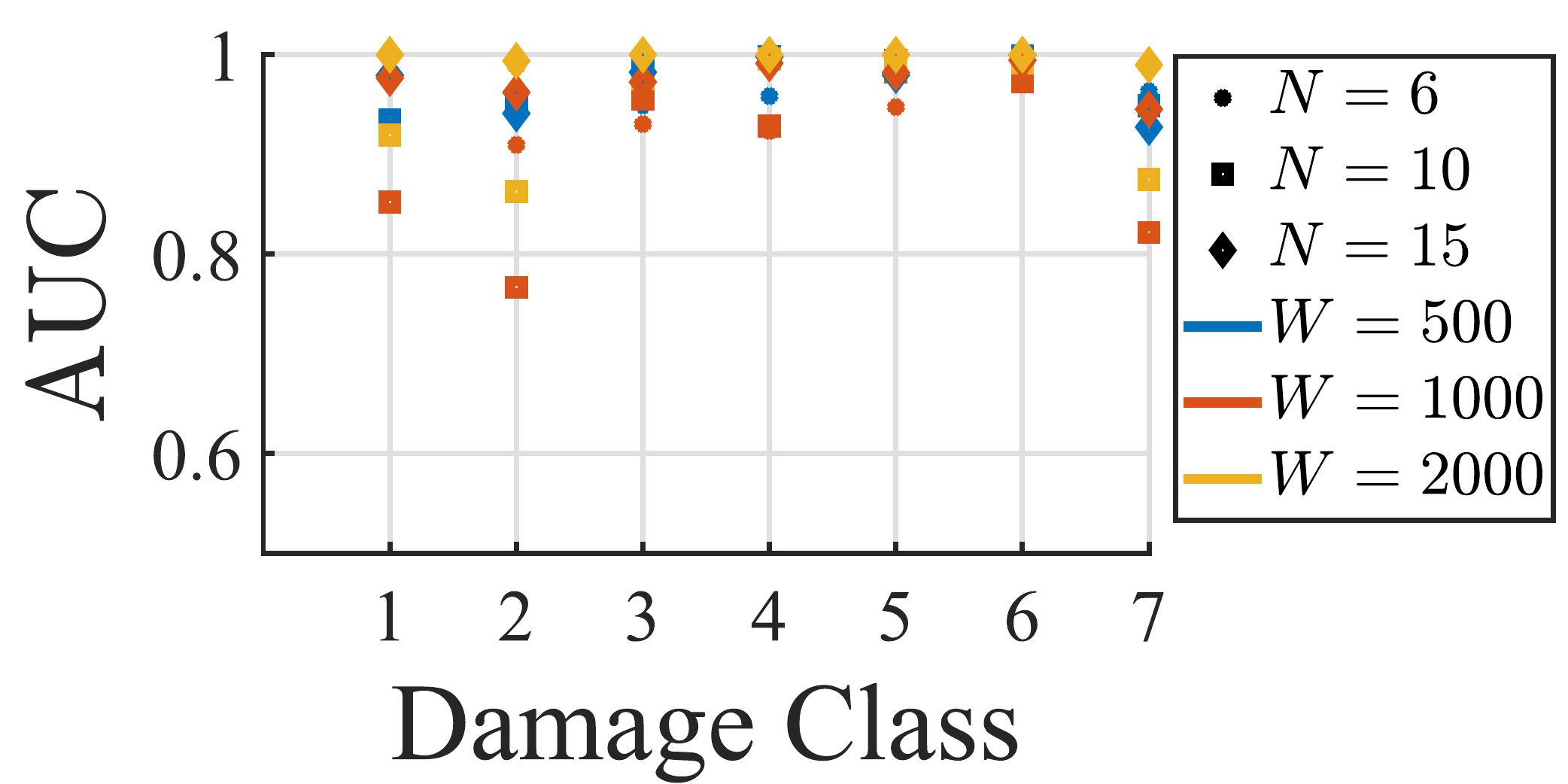}
        \caption {Source QUGS.}
        \label{Fig:Z_R_2}  
\end{subfigure}
\caption{TL performance; AUC values for all Z24 TL scenarios.}
\label{Fig:Z_R_A}
\end{figure}

\noindent
TL for the target structure Z24 with seven damage cases is performed with the source structures Yellow Frame and QUGS. ROC curves for 6 of the TL scenarios in Table~\ref{Table:TL_Scenarios} are shown in Fig.~\ref{Fig:Z_ROC_A}. A notable point in Z24 is that it retains higher AUCs with lower $N$ and $W$. Z24 damages are large-scale (and more realistic) structural damages compared to the QUGS and Yellow Frame, and thus lower $W$, and $N$ values do not alter the AUC as much as in those datasets. Accordingly, DA-fed source-no-damage identifier achieves a high discriminative potential between all no-damage and damage cases, even with $W=500$ and $N=6$. All scenarios' AUCs are presented in Fig.~\ref{Fig:Z_R_A}, showing the DA robustness to $W$.

\subsection{TL-based Zero-shot SDD Results}
\label{sec:T_T}
\begin{figure}[h!]
\centering
\begin{subfigure}{1\linewidth} 
\centering
  \includegraphics[width=1\linewidth]{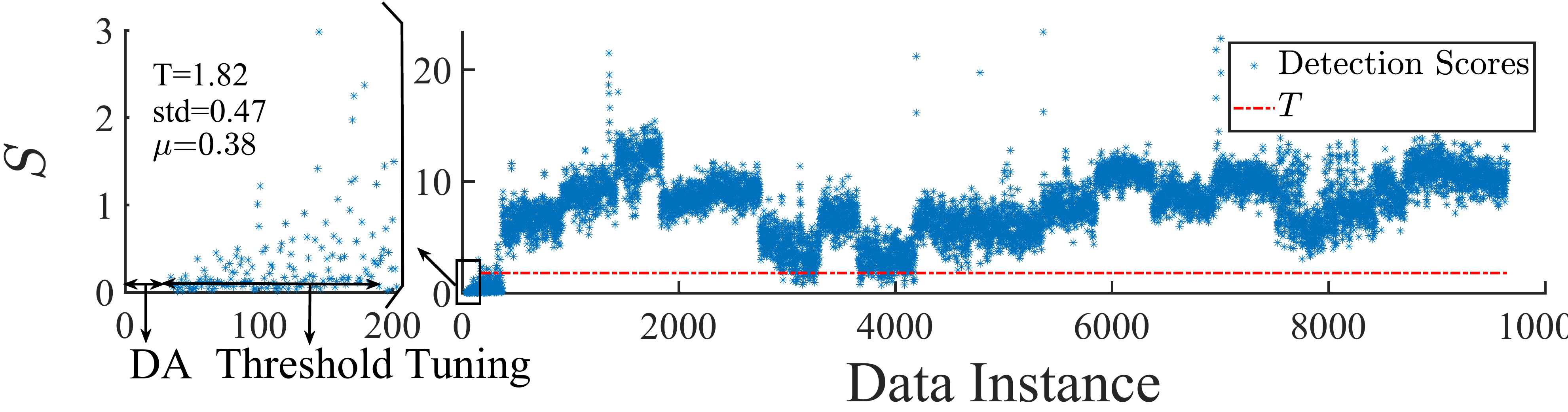}
        \caption {Detection scores.}
        \label{Fig:Z_S_Q_Y_S}  
\end{subfigure}
\begin{subfigure}{1\linewidth} 
\centering
  \includegraphics[width=1\linewidth]{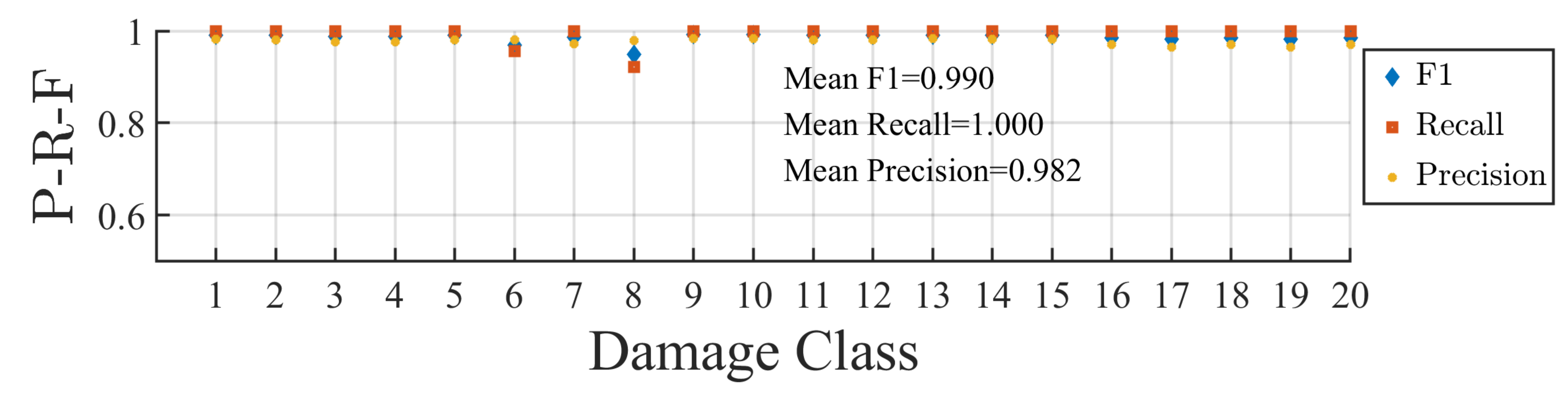}
        \caption {Precision, Recall and F1 values for all cases.}
        \label{Fig:Z_S_Q_Y_P}  
\end{subfigure}
\caption{Zero-shot SDD with source QUGS and target Yellow Frame; $W=2000$, $N=15$.}
\label{Fig:Z_S_Q_Y}
\end{figure}

\begin{figure}[h!]
\centering
\begin{subfigure}{1\linewidth} 
\centering
  \includegraphics[width=1\linewidth]{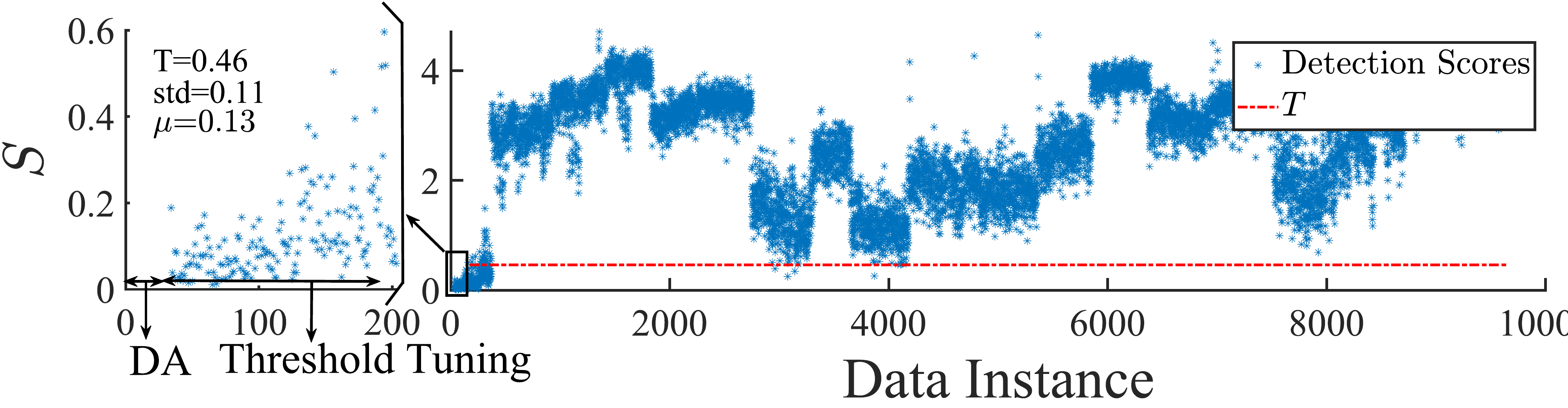}
        \caption {Detection scores.}
        \label{Fig:Z_S_Z_Y_D}  
\end{subfigure}
\begin{subfigure}{1\linewidth} 
\centering
  \includegraphics[width=1\linewidth]{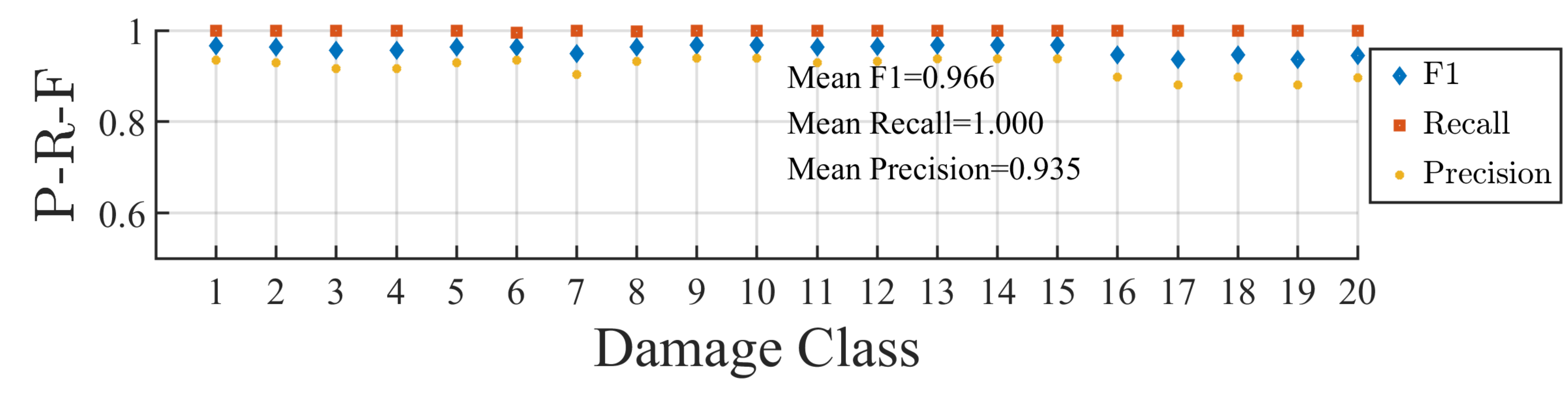}
        \caption {Precision, Recall and F1 values for all cases.}
        \label{Fig:Z_S_Z_Y_P}  
\end{subfigure}
\caption{Zero-shot SDD with source Z24 and target Yellow Frame; $W=2000$, $N=15$.}
\label{Fig:Z_S_Z_Y}
\end{figure}

\begin{figure}[h!]
\centering
\begin{subfigure}{1\linewidth} 
\centering
  \includegraphics[width=1\linewidth]{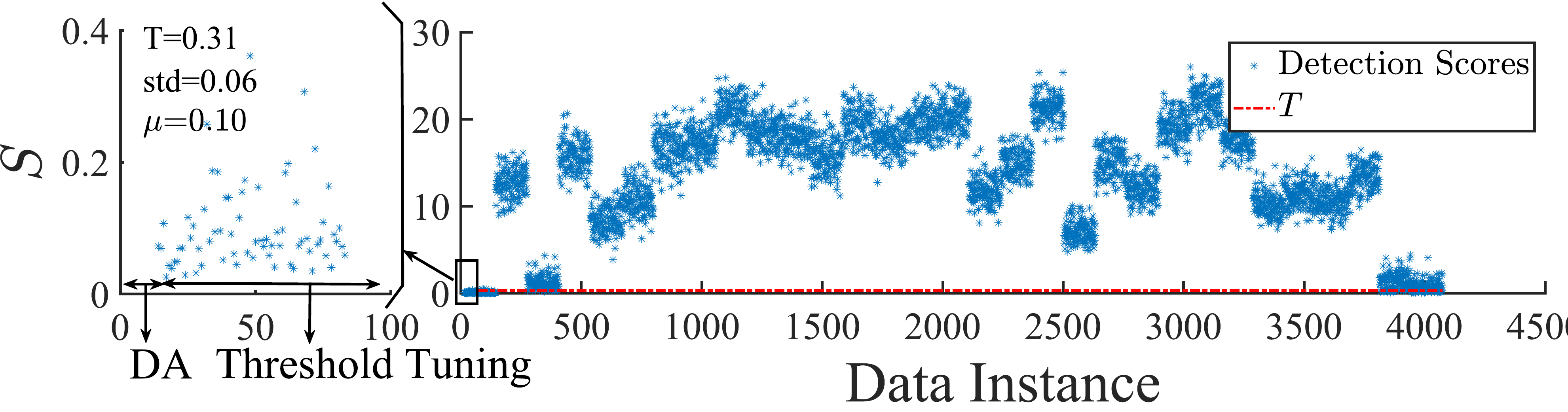}
        \caption {Detection scores.}
        \label{Fig:Z_S_Y_Q_S}  
\end{subfigure}
\begin{subfigure}{1\linewidth} 
\centering
  \includegraphics[width=1\linewidth]{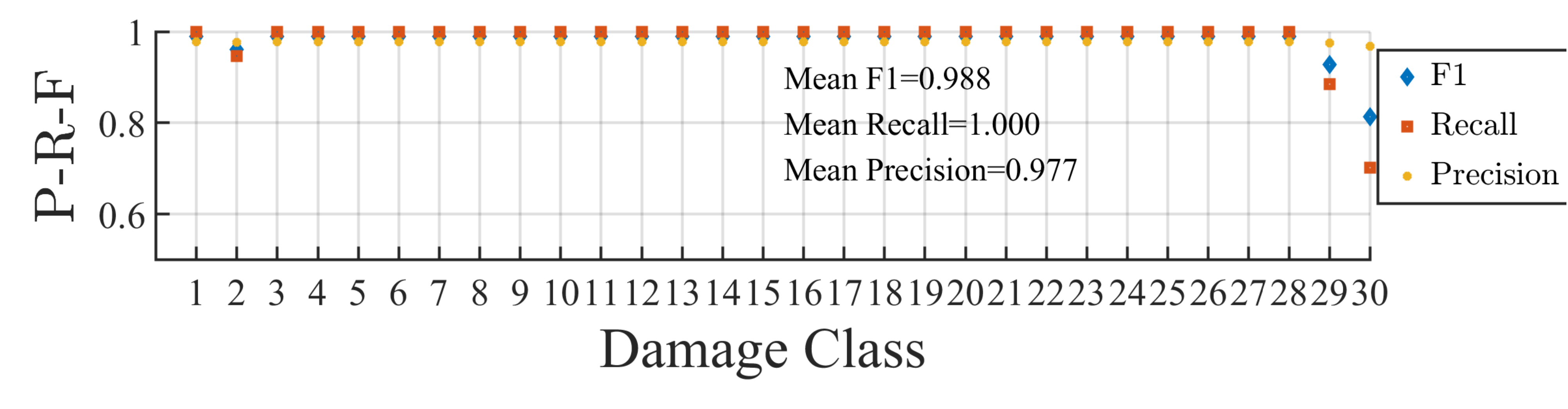}
        \caption {Precision, Recall and F1 values for all cases.}
        \label{Fig:Z_S_Y_Q_P}  
\end{subfigure}
\caption{Zero-shot SDD with source Yellow Frame and target QUGS; $W=2000$, $N=15$.}
\label{Fig:Z_S_Y_Q}
\end{figure}

\begin{figure}[h!]
\centering
\begin{subfigure}{1\linewidth} 
\centering
  \includegraphics[width=1\linewidth]{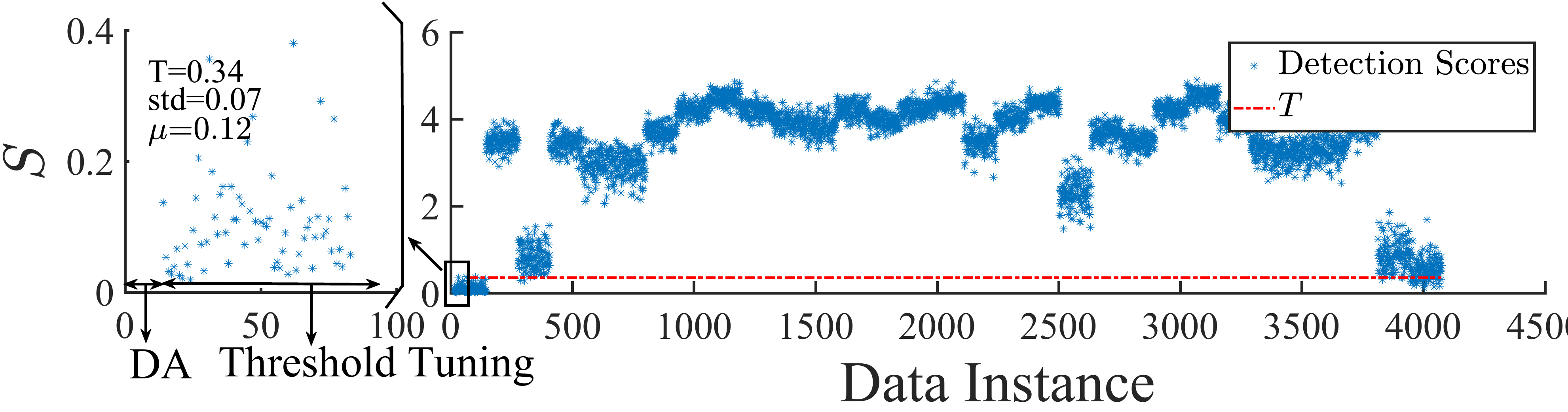}
        \caption {Detection scores.}
        \label{Fig:Z_S_Z_Q_S}  
\end{subfigure}
\begin{subfigure}{1\linewidth} 
\centering
  \includegraphics[width=1\linewidth]{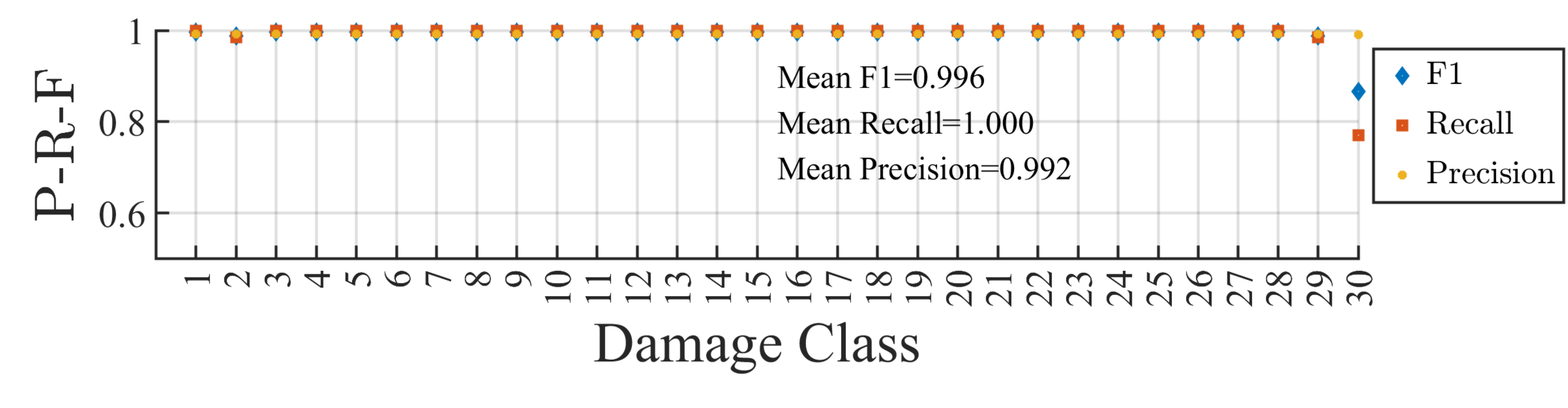}
        \caption {Precision, Recall and F1 values for all cases.}
        \label{Fig:Z_S_Z_Q_P}  
\end{subfigure}
\caption{Zero-shot SDD with source Z24 Frame and target QUGS; $W=2000$, $N=15$.}
\label{Fig:Z_S_Z_Q}
\end{figure}

\begin{figure}[h!]
\centering
\begin{subfigure}{1\linewidth} 
\centering
  \includegraphics[width=1\linewidth]{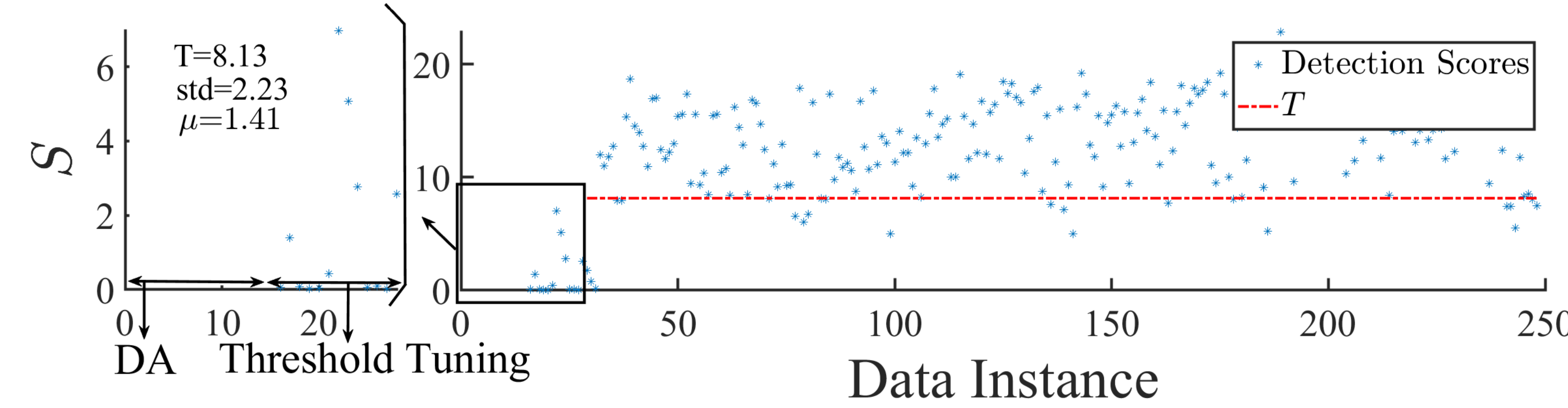}
        \caption {Detection scores.}
        \label{Fig:Z_S_Y_Z_S}  
\end{subfigure}
\begin{subfigure}{1\linewidth} 
\centering
  \includegraphics[width=1\linewidth]{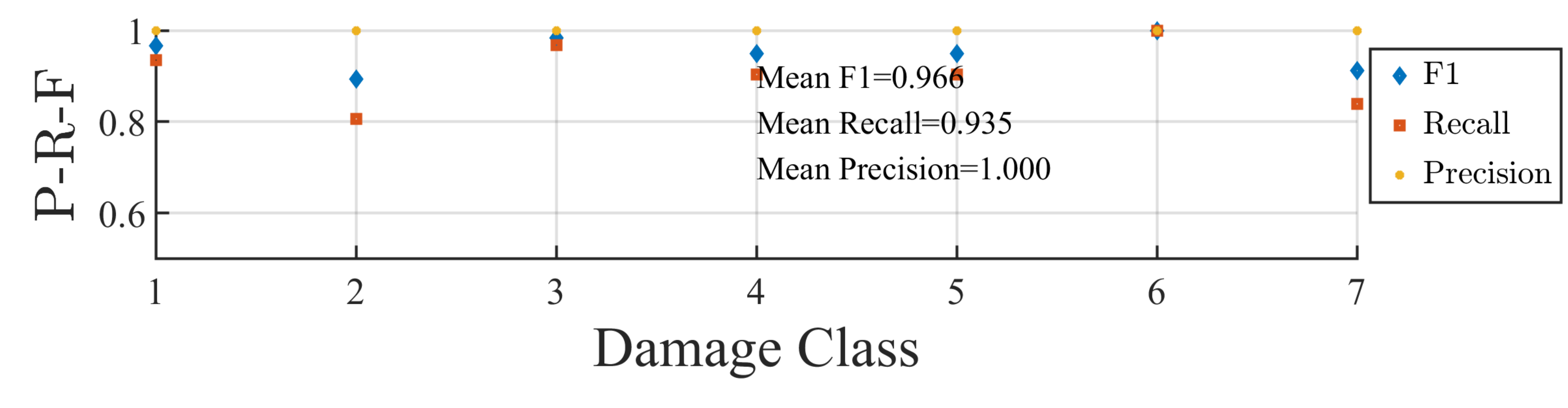}
        \caption {Precision, Recall and F1 values for all cases.}
        \label{Fig:Z_S_Y_Z_P}  
\end{subfigure}
\caption{Zero-shot SDD with source Yellow Frame and target Z24; $W=2000$, $N=15$.}
\label{Fig:Z_S_Y_Z}
\end{figure}

\begin{figure}[h!]
\centering
\begin{subfigure}{1\linewidth} 
\centering
  \includegraphics[width=1\linewidth]{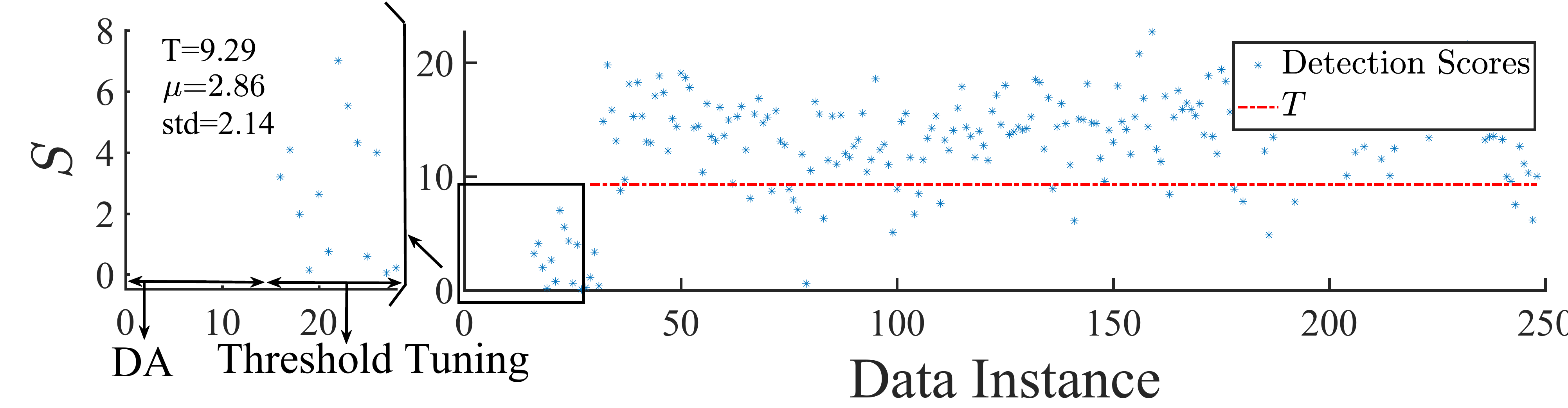}
        \caption {Detection scores.}
        \label{Fig:Z_S_Q_Z_S}  
\end{subfigure}
\begin{subfigure}{1\linewidth} 
\centering
  \includegraphics[width=1\linewidth]{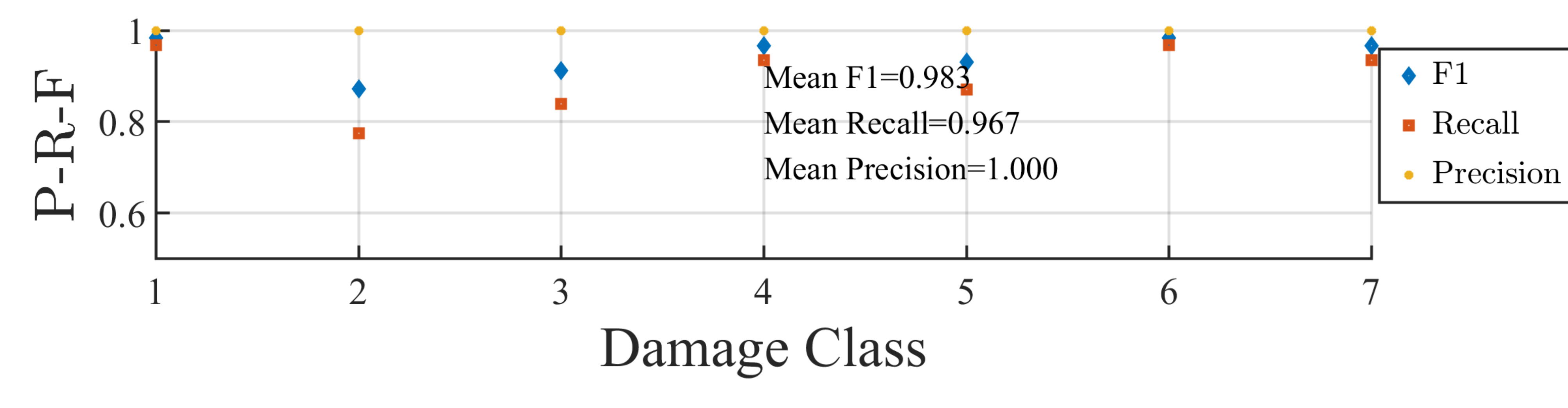}
        \caption {Precision, Recall and F1 values for all cases.}
        \label{Fig:Z_S_Q_Z_P}  
\end{subfigure}
\caption{Zero-shot SDD with source QUGS and target Z24; $W=2000$, $N=15$.}
\label{Fig:Z_S_Q_Z}
\end{figure}

\noindent
This sub-section demonstrates the proposed threshold tuning algorithm's performance in zero-shot SDD. Zero-shot SDD is carried out using each dataset's highest $N$ (Table~\ref{Table:TL_Scenarios}) and $W=2000$ for all damage cases in three target structures: QUGS, Yellow Frame, and Z24. Damage detection results for target structure Yellow Frame are shown in Figs~\ref{Fig:Z_S_Q_Y}  and~\ref{Fig:Z_S_Z_Y} when the QUGS and Z24 are the source structures. For the QUGS structure as the target, SDD results are shown in Figs~\ref{Fig:Z_S_Y_Q}  and~\ref{Fig:Z_S_Z_Q} when the Yellow Frame and Z24 are the source structures. Finally, for the Z24 as the target and Yellow Frame and QUGS as source structures, SDD results are shown in Figs~\ref{Fig:Z_S_Y_Z}  and~\ref{Fig:Z_S_Q_Z}, respectively.

The results indicate that with only two possible source structures and dozens of distinct and complicated damage cases, almost all damages in the target domains are detected with a high $F1$ score. Another important observation is that $S$ values have a wide range of distributions, indicating that the proposed DA and the trained $\mathcal{D}$ models not only distinguish no-damage and damage cases but also discern between damage cases. With the Yellow frame as the target, Z24 contrasts between damage cases better (\ie different ranges of $S$) than QUGS, while with the QUGS as the target, the Yellow frame offers a better contrast compared to Z24. The differences between the contributions of each source structure to each target structure can be the basis for future multi-view ensemble studies in more advanced DA applications for SHM and SDD. Such multi-view approaches tend to lower false alarms in cases such as Yellow Frame SDD with Z24 as the source, where a non-optimal SDD threshold provokes false alarms while damage and no-damage cases are distinguished from each other.

\section{Conclusions}
\noindent
Deep-Learning (DL)-aided Structural Health Monitoring (SHM) has already reached high-performance levels on individual structures, ranging from laboratory-based setups to real-world infrastructure under varying environmental and operational effects. However, these DL-based SHM methods offer only case-dependent solutions, hindering SHM applications on populations of diverse infrastructure (large-scale SHM) where no damage cases are yet present. Difficulties include the need for labeled data, case-dependent dimensionality reduction, and method re-training for each new structure. Case dependency is, thus, the major obstacle to establishing large-scale SHM tools/methods operating over diverse infrastructure with no comprehensive prior data. Transfer Learning (TL)---which accepts full-spectrum FFT amplitudes as features and is well-aligned with the zero-shot learning paradigm---is a viable solution to address the aforementioned challenges. This study presented a novel and highly effective TL approach that addresses the aforementioned limitations and shortcomings. A dedicated detector $\mathcal{D}$ model of GAN was trained only in the source structure(s), which accomplished highly accurate damage detection on target domains using a domain adaptation (DA) technique that was also introduced in the present study.

The proposed approach was rigorously evaluated on three benchmark datasets: Z24 (seven damage and one no-damage cases), Yellow Frame (20 damage and one no-damage case), and Qatar University Grandstand Simulator - QUGS (30 damage and one no-damage cases). Results revealed that the accumulated knowledge on distinguishing no-damage cases from damage cases is transferred seamlessly across those datasets. The area under the Receiver Operating Characteristics curves (Area Under the Curve - AUC) is used for assessing the quality of differentiation between the no-damage and damage case while avoiding any threshold. The achieved high AUC values (mean AUC of 0.9986 across all datasets using all data channels) allow for selecting a simple Structural Damage Detection (SDD) threshold tuning. For the datasets studied in this paper, 10\% of the no-damage case data is used for spectral mapping and the other 40\% for threshold tuning. A Gaussian distribution is fitted to the observed data with the mean plus three times the standard deviation set as the threshold. The proposed zero-shot SDD approach achieved mean precision, recall, and $F1$ scores that ranged between 0.96 to 0.99 for all target structures.

 \section*{Acknowledgments}
 
The authors would like to thank Dr. Carlos Ventura and Dr. Alexander Mendler for providing the Yellow Frame dataset and Dr. Giacomo Bernagozzi for providing its modal information. Resources and codes to reproduce results in Sections~\ref{Sec:Source_Tuning} and~\ref{sec:res} will be made publicly available on GitHub upon publication at \url{https://github.com/Hesam-92-19/Transfer_Learning_SHM}, which will enable users to interact with the repository using their datasets. 

 \section*{Declaration of Conflicting Interests}
The authors declared no potential conflicts of interest with respect to this article's research, authorship, and/or publication.

\bibliography{Bib.bib}
\end{document}